\documentclass[twocolumn, switch]{article} 

\usepackage{preprint}

\usepackage{amsmath, amsthm, amssymb, amsfonts}
\usepackage{bm}
\usepackage{mathtools}

\usepackage[numbers,square]{natbib}
\usepackage[utf8]{inputenc}	
\usepackage[T1]{fontenc}	
\usepackage{xcolor}		
\usepackage[colorlinks = true,
            linkcolor = blue,
            urlcolor  = blue,
            citecolor = blue,
            anchorcolor = black]{hyperref}	
\usepackage{booktabs} 		
\usepackage{nicefrac}		
\usepackage{microtype}		
\usepackage{lineno}		
\usepackage{float}			

\usepackage{newfloat}
\DeclareFloatingEnvironment[name={Supplementary Figure}]{suppfigure}
\usepackage{sidecap}
\sidecaptionvpos{figure}{c}

\usepackage{titlesec}
\titlespacing\section{0pt}{12pt plus 3pt minus 3pt}{1pt plus 1pt minus 1pt}
\titlespacing\subsection{0pt}{10pt plus 3pt minus 3pt}{1pt plus 1pt minus 1pt}
\titlespacing\subsubsection{0pt}{8pt plus 3pt minus 3pt}{1pt plus 1pt minus 1pt}

\usepackage{graphics}
\usepackage{graphicx}
\usepackage{epstopdf}
\usepackage{epsfig}
\usepackage{hyperref}
\usepackage{color}
\usepackage{multirow}
\usepackage{hhline}
\usepackage[flushleft]{threeparttable}
\usepackage{subfigure}
\usepackage{amsthm,amsmath,amssymb}
\usepackage{mathrsfs}
\usepackage{subfigure}
\usepackage{graphicx}
\usepackage{amsmath}
\usepackage{amssymb}
\usepackage{multirow}
\usepackage{booktabs}
\usepackage{bm}
\usepackage{rotating}
\usepackage{threeparttable}
\usepackage{array}
\graphicspath{{figures/}}

\title{Multiple Video Frame Interpolation via Enhanced Deformable Separable Convolution}

\usepackage{authblk}

\author{Xianhang Cheng}
\author{Zhenzhong Chen\thanks{\tt{zzchen@ieee.org}}}
\affil{School of Remote Sensing and Information Engineering, Wuhan University}

\begin{document}

\twocolumn[ 
  \begin{@twocolumnfalse} 

\maketitle

\begin{abstract}
Generating non-existing frames from a consecutive video sequence has been an interesting and challenging problem in the video processing field. 
Typical kernel-based interpolation methods predict pixels with a single convolution process that convolves source frames with spatially adaptive local kernels, which circumvents the time-consuming, explicit motion estimation in the form of optical flow.
However, when scene motion is larger than the pre-defined kernel size, these methods are prone to yield less plausible results. In addition, they cannot directly generate a frame at an arbitrary temporal position because the learned kernels are tied to the midpoint in time between the input frames.
In this paper, we try to solve these problems and propose a novel non-flow kernel-based approach that we refer to as enhanced deformable separable convolution (EDSC) to estimate not only adaptive kernels, but also offsets, masks and biases to make the network obtain information from non-local neighborhood. During the learning process, different intermediate time step can be involved as a control variable by means of an extension of coord-conv trick, allowing the estimated components to vary with different input temporal information. This makes our method capable to produce multiple in-between frames.
Furthermore, we investigate the relationships between our method and other typical kernel- and flow-based methods.
Experimental results show that our method performs favorably against the state-of-the-art methods across a broad range of datasets. Code will be publicly available on URL: \url{https://github.com/Xianhang/EDSC-pytorch}.
\end{abstract}
\vspace{0.35cm}

  \end{@twocolumnfalse} 
] 



\section{Introduction}
{\let\thefootnote\relax\footnote{{This work was supported in part by grants from the National Natural Science Foundation of China under Grant 62036005 and the Fundamental Research Funds for the Central Universities. (Corresponding author: Zhenzhong Chen, E-mail:  \texttt{zzchen@ieee.org})}}}Video frame interpolation aims to synthesize middle non-existent frames between the original input video frames, which is a long-studied problem in computer vision.
The technology is beneficial to various applications in the field of video processing, ranging from frame rate up-conversion \cite{538926, 8334253}, frame recovery and intra prediction in video coding \cite{7358172, dnn}, slow motion generation \cite{superslomo, MEMCNet, dain} to novel view synthesis \cite{DeepStereo}.

Early proposed methods exploit the motion from time-varying images with 2D flow fields, in which pixel movements are represented by coordinate shifts \cite{161350, mdpflow}. Based on the estimated optical flow, frame interpolation algorithms typically warp and blend original frames to produce interpolation results \cite{Broxflow, middleburry}.
As the optical flow from the existent frames to target frame can be approximately estimated from the bi-directional flows, intermediate frames with multiple time steps can be generated.
However, directly synthesizing the intermediate frames guided by optical flow may produce visual artifacts. In some challenging conditions such as occlusion, large motion, illumination or nonlinear structural changes, the optical flow accuracy decreases, resulting in distortion or artifacts. Recent deep learning approaches towards optical flow estimation have found remarkable success \cite{flownet, flownet2, spynet, liteflownet, PWCNet, PWCNet2, flowfields}.
While the progress has been made to some extent, they aim at flow estimation rather than frame interpolation, producing less convincing results \cite{Ctxsyn, MSPFT, MEMCNet}.

Some recent deep learning methods adopt advanced flow estimation model or its variations as sub-networks to directly synthesizing the interpolation frames in an end-to-end manner \cite{Ctxsyn, MSPFT, MEMCNet, dain, Toflow,PoSNet,softsplat}, where the intermediate frames act as supervision signals for training. Typically, occlusion masks or visibility maps are learned to smoothly transition across images as the synthesis happens in both the ``forward" and ``backward" direction, simultaneously. However, these approaches heavily depend on the quality of bi-directional optical flows, whose estimation process is sophisticated and time-consuming.

Another major trend in this research is to leverage adaptive convolution for interpolation \cite{adaconv, sepconv}.
For each output pixel, a pair of 2D kernels or four 1D kernels (two for horizontal and the other two for vertical direction) are learned with a neural network.
Notably, to handle large motion, large kernel size is required for these kernel-based interpolation methods.
Though these methods are able to generate reasonable results, there are some drawbacks:
1) These methods can be problematic since the pre-defined kernel size is certain, which impedes the interpolation results when scene motion is larger than kernel size.
2) It is expensive to consider thousands of pixels to synthesize only one output pixel.
3) These methods cannot produce a frame at an arbitrary time because the kernel parameters are tied to the time step of the intermediate frame.
Some methods try to integrate optical flow into kernel-based methods with adaptive warping layers to deal with the limitations \cite{MEMCNet, dain}.
They inevitably inherit some corresponding drawbacks from both sides.
On one hand, flow estimation is computationally expensive.
On the other hand, these methods only consider pixels in a small square area, which makes it rather challenging when handling inaccurate motion estimates.

In this paper, we address the drawbacks mentioned above by presenting a more powerful and effective approach coined Enhanced Deformable Separable Convolution (EDSC). We argue that the limitation of the previous kernel-based interpolation methods \cite{adaconv, sepconv} is because they process the pixels only in the local neighborhood, which takes no effect on pixels outside the regular grid. Drawing inspiration from the success of deformable convolution networks \cite{dcn, dcn2}, we propose to learn adaptive kernels, offsets, masks and biases for interpolation, allowing us to use far fewer but more effective pixels to deal with large motion. We further propose to involve different intermediate time steps, making it possible for non-flow interpolation methods to generate a frame at any time instant between two frames.
Moreover, we show in detail that conventional flow-based interpolation methods can be regarded as specific instances of our method in terms of pixel reference. Our experiments show that the proposed method achieves the best performance of any existing kernel-based methods and performs favorably against representative state-of-the-art interpolation methods without relying on any other pre-trained components.

Hence, our contributions are:

(1) A novel kernel-based method is proposed, which learns not only spatially-adaptive separable convolution kernels, but also deformable offsets, masks and biases to obtain information in a non-local neighborhood. This model is able to handle different degrees of motion, which is not constrained by the pre-defined kernel size.

(2) In our network, different estimators are designed, in which temporal information can be involved as a control variable by means of an extension of coord-conv trick. Such a design enables our network to directly produce a frame at an arbitrary time, without using a recursive manner.

(3) From the perspective of convolution, both some flow-based and kernel-based methods are theoretically demonstrated as special cases of our proposed EDSC.

Based on the above contributions, our model performs favorably against the state-of-the-art methods, even though any extra, complex and pre-calculated information (like context, depth, flow and edge information) is not involved in our network.

Please note that, this paper is the extension of our earlier publication \cite{dsepconv} in the $34^{\mathrm{th}}$ AAAI Conference on Artificial Intelligence. The changes and improvements are summarized here.
First, in the encoder-decoder architecture, heterogeneous convolution (HetConv) \cite{hetconv} is utilized to reduce computation and parameters of the model. In contrast to DSepConv \cite{dsepconv}, we save about 79.6\% FLOPs in computation and 59.6\% parameters with no loss in accuracy.
Second, an additional bias estimator is introduced to learn residual values to account for pixel synthesis that cannot be well performed by the adaptive convolution. Such a design allows us to shift the pixel values up and down to fit the prediction with the data better, which counterparts the bias term in the convolution operation.
Third, based on the observation that convolutions with extra coordinate channels are particularly beneficial to spatially-conditioned generation tasks \cite{coordconv}, we propose to input temporal index as a new control variable.
This trick enables our model to output different kernels, offsets and masks at different time steps.
Alternatively, more comprehensive analysis and evaluations are provided in this paper.

\section{Related Work}
In this section, we discuss and provide an overview of recent interpolation methods in the following parts.

\subsection{Single Frame Interpolation}
Most recently existing interpolation methods are designed specifically for single frame interpolation, which mainly consider the midpoint (in time) between two reference frames.
Typically, substantial effort is made to first estimate bi-directional optical flow or its variations and then to synthesize the in-between frame guided by motion. Considering the input frames are not equally informative due to occlusion, mask maps are often estimated together with optical flow for adaptively blending the warped frames. Specifically, Liu \emph{et al.} \cite{DVF} proposed a fully-convolutional network DVF to predict 3D flow across space and time. The in-between frame was then generated by trilinear sampling. Liu \emph{et al.} \cite{cyclicgen} further improved the performance of DVF by leveraging edge information \cite{edge} and a novel cycle consistency loss. Jiang \emph{et al.} \cite{superslomo} proposed SuperSloMo, which utilized two U-Net architectures to compute bi-directional optical flows and soft visibility maps, respectively. Furthermore, based on SuperSloMo \cite{superslomo}, Reda \emph{et al.} \cite{unslomo} proposed unsupervised techniques to synthesize intermediate frames using cycle consistency. Yuan \emph{et al.} \cite{zoomintocheck} proposed a model which warped not only input frames, but also their corresponding features extract from ResNet \cite{resnet}.

In order to get more accurate optical flow, some methods utilized off-the-shelf flow estimation architectures with pre-trained parameters as sub-modules in their networks. For instance, Xue \emph{et al.} \cite{Toflow} proposed ToFlow which utilized SpyNet \cite{spynet} to estimate optical flow. Niklaus \emph{et al.} \cite{Ctxsyn, softsplat} utilized PWC-Net \cite{PWCNet} and a modified GridNet \cite{gridnet} to warp and generate interpolated frames. Xu \emph{et al.} \cite{qvi} utilized PWC-Net \cite{PWCNet} to compute optical flows from four input frames. Haris \emph{et al.} \cite{STAR} adopted flow images computed by \cite{liu2009beyond} and refined them for both video frame interpolation and super resolution.
Cheng \emph{et al.} \cite{MSPFT} proposed a position feature transform layer, transforming optical flow calculated from PWC-Net \cite{PWCNet} into scaling factors to adjust frame interpolation process.

Some methods borrow operations from other image or video processing tasks (\emph{e.g.}, video super resolution) and generate intermediate frames without a component of optical flow computation. For instance, Choi \emph{et al.} proposed CAIN \cite{cain}, which employed PixelShuffle \cite{pixelshuffle} and operation with channel attention mechanism \cite{attention}. Shen \emph{et al.} \cite{bin} proposed a blurry video frame interpolation (BIN) method for jointly frame interpolation and deblurring. Xiang \emph{et al.} \cite{zoomingslowmo} proposed a one-stage space-time video super-resolution for jointly frame interpolation and super-resolution. Choi \emph{et al.} \cite{meta} proposed to improve the performance of an interpolation algorithm by incorporating meta-learning.

\begin{table*}[th]
\caption{A list of notations mainly used in this paper.}
\centering
{
\resizebox{\linewidth}{!}{
\begin{tabular}{|l|l|}
\hline
Symbol & Definition \\
\hline
\hline
$\bm{\mathrm{I}}_1$,$\bm{\mathrm{I}}_2,\hat{\bm{\mathrm{I}}}$ & Previous frame, current frame, estimated intermediate frame \\
\hline
$t$ & Arbitrary intermediate time step, $t\in(0, 1)$\\
\hline
$x,y$ & Pixel coordinates in a frame \\
\hline
$n$ & Specific kernel size\\
\hline
$\bm{\mathrm{P}}_i(x,y)$ & A \emph{local} patch centered at $(x,y)$ in the input frame $\bm{\mathrm{I}}_i$, for $i=1,2$ \\
\hline
$\bm{\mathrm{P}}_i^{'}(x,y)$ & A resampled \emph{non-local} patch centered at $(x,y)$ guided by \emph{learnable offsets} in the input frame $\bm{\mathrm{I}}_i$, for $i=1,2$ \\
\hline
$\bm{\mathrm{P}}_i^{''}(x,y)$ & A resampled \emph{non-local} patch centered at $(x,y)$ guided by \emph{optical flow} in the input frame $\bm{\mathrm{I}}_i$, for $i=1,2$ \\
\hline
$\bm{\mathrm{B}}_i(x,y)$ & \emph{Fixed} convolutional kernels (bilinear interpolation coefficients) for a patch centered at $(x, y)$ in the input frame $\bm{\mathrm{I}}_i$, for $i=1,2$ \\
\hline
$\bm{\mathrm{K}}_i(x,y)$ & \emph{Learned} convolutional kernels for a patch centered at $(x, y)$ in the input frame $\bm{\mathrm{I}}_i$, for $i=1,2$ \\
\hline
$\bm{\mathrm{k}}_{i,v},\bm{\mathrm{k}}_{i,h}$ & Learned separable convolutional kernels in vertical and horizontal direction\\
\hline
${\mathrm{k}}_{i}$ & Occlusion masks for each pixel used in flow-based interpolation methods, for $i=1,2$\\
\hline
$\bm{\mathrm{p}}_{i,j}$ & Each pre-specified offset for the $j$-th ($j\in [1, n^2]$) location in patch $\bm{\mathrm{P}}_i$ or $\bm{\mathrm{P}}_i^{'}$, for $i=1,2$ \\
\hline
$\Delta\bm{\mathrm{p}}_{i,j}$ & Learned offset for the $j$-th ($j\in [1, n^2]$) location in patch $\bm{\mathrm{P}}_i$, for $i=1,2$ \\
\hline
$\Delta\bm{\mathrm{m}}_{i,j}$ & Learned mask (modulation scalar) for the $j$-th ($j\in [1, n^2]$) location in patch $\bm{\mathrm{P}}_i$, for $i=1,2$ \\
\hline
$\Delta\mathrm{b}(x, y)$ & Learned bias for each output pixel centered at $(x,y)$\\
\hline

\end{tabular}}
}
\label{tab:1}
\end{table*}

There are some other studies that regard flow estimation as an intermediate step, which can be circumvented with a single convolution process. As a prior of kernel based interpolation methods, AdaConv \cite{adaconv} was proposed to estimate a pair of spatially-adaptive convolution kernels for each output pixel with a neural network. To reduce large memory demand, Niklaus \emph{et al.} \cite{sepconv} proposed SepConv that separated each 2D convolution kernel into two 1D kernels. Choi \emph{et al.} \cite{dnn} further improved the structure of SepConv \cite{sepconv} that both uni-directional and bi-directional prediction were available in video coding. Peleg \emph{et al.} \cite{imnet} modified SepConv \cite{sepconv} into a multi-scale architecture and formulated interpolated motion estimation as classification by calculating the center-of-mass of the convolution kernels. Concurrently to our work, Lee \emph{et al.} \cite{adacof} proposed a new warping module AdaCoF with a similar motivation to ours. They further introduce a dual-frame adversarial loss to improve their performance.
Moreover, Bao \emph{et al.} \cite{MEMCNet, dain} combined the advantages of flow based and kernel based methods, proposed an adaptive warping layer that warps images or features based on the given optical flow and learned local convolution kernels.

\subsection{Multiple Frame Interpolation}
A straight-forward way to generate multiple intermediate frames is to recursively apply a single frame video interpolation method. However, this manner is not flexible enough and error would accumulate during the recursive process. Some flow based interpolation methods \cite{Ctxsyn, superslomo, dain, unslomo, qvi, softsplat, PoSNet} are also well-suited for multi-frame interpolation while the other are not. The difference among these methods is whether the occlusion reasoning is tied to an arbitrary time step \cite{superslomo, unslomo} and whether motion compensation is performed before synthesizing the output frame \cite{Ctxsyn, dain, softsplat, PoSNet, qvi}.

Several methods utilize phase information to learn the motion relationship for multiple video frame interpolation. Meyer \emph{et al.} \cite{phase} proposed the phase-based method which utilized phase information across the levels of a multi-scale pyramid. Furthermore, combined with CNNs, PhaseNet \cite{phasenet} was proposed with a better performance. Another related problem is video frame inpainting, which focus on the intersection of general video inpainting, frame interpolation and video prediction. Szeto \emph{et al.} \cite{bitai} devised a method bi-TAI that was composed of a bidirectional video prediction module and a temporally-aware frame interpolation module, achieving impressive inpainting results.

In relation to non-flow kernel-based interpolation methods \cite{adaconv, sepconv, dnn, dsepconv, adacof}, to our best knowledge, none of these methods can directly generate frames at an arbitrary temporal position. In light of this limitation, we suggest to solve the under-explored problem.

\section{Proposed Method}
In this section, we introduce our proposed algorithm for video frame interpolation, including the details of our network architecture and our training details.
The notations are provided in Table \ref{tab:1} for clarity.

\begin{figure*}[th]
\centering
\subfigure[Kernel based.]{\label{fig1:subfig:a}
\includegraphics[width=0.48\linewidth]{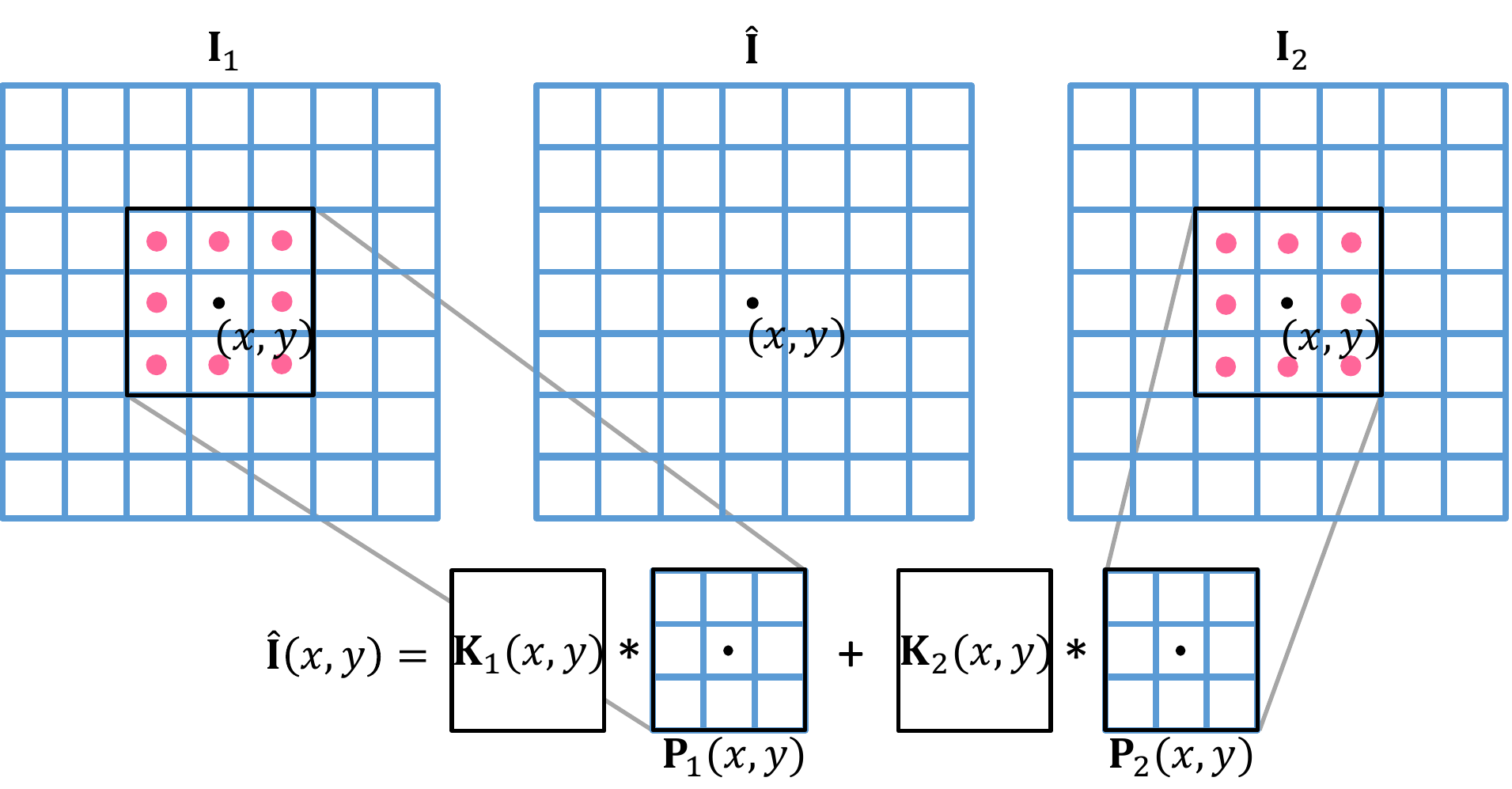}}
\hfill
\subfigure[Our proposed.]{\label{fig1:subfig:b}
\includegraphics[width=0.48\linewidth]{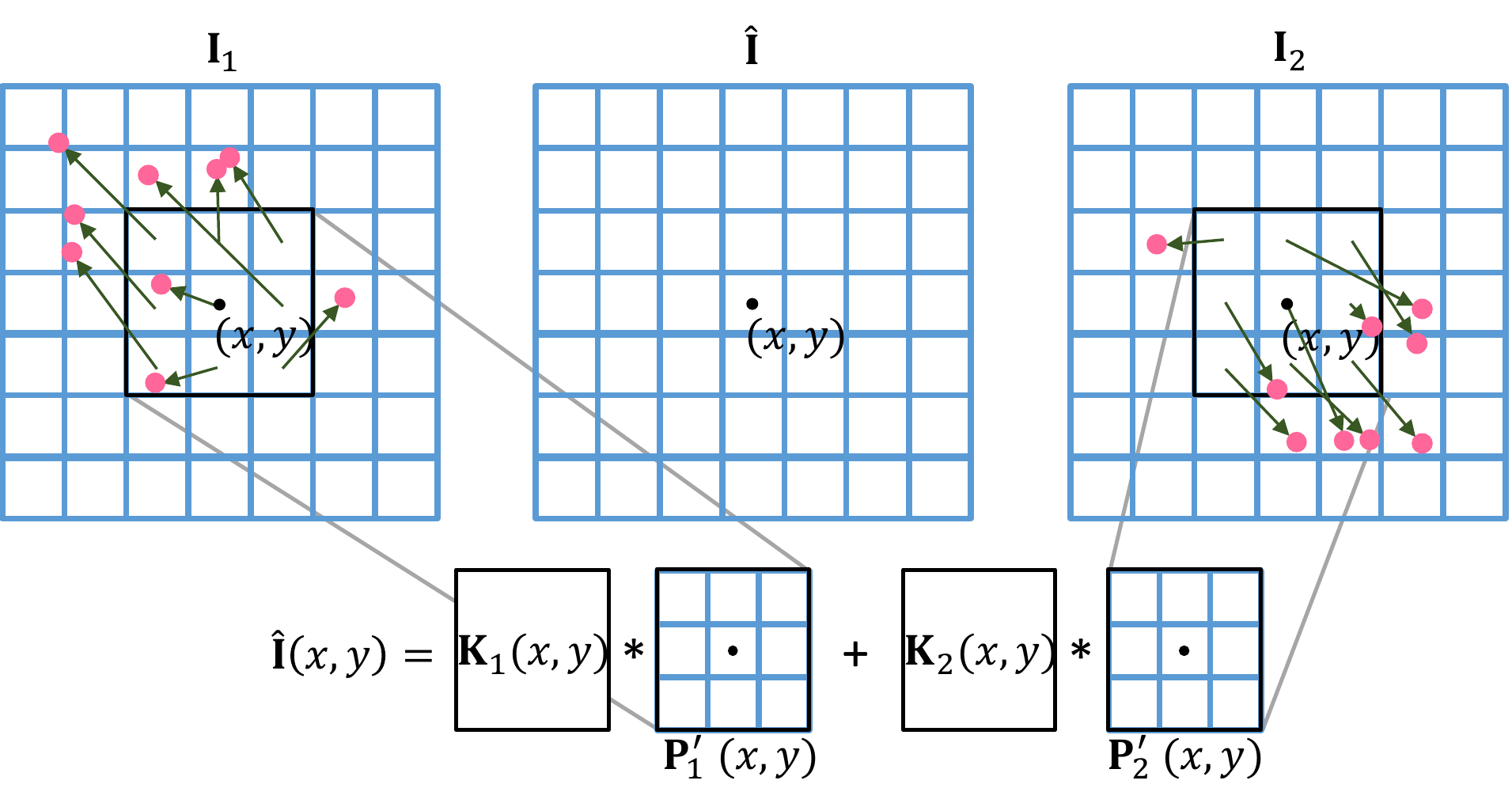}}
\subfigure[Flow based.]{\label{fig1:subfig:c}
\includegraphics[width=0.48\linewidth]{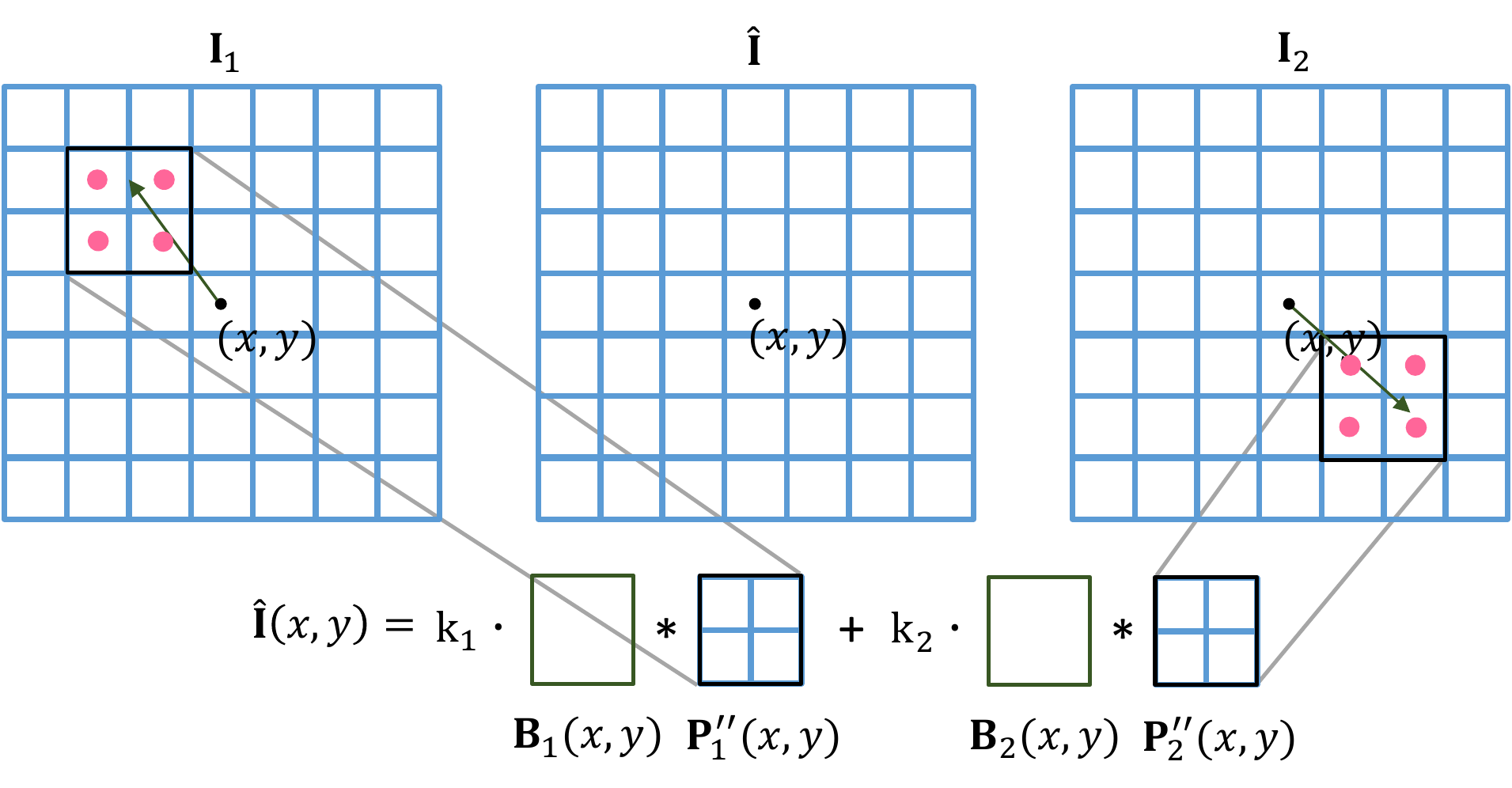}}
\hfill
\subfigure[Adaptive warping based.]{\label{fig1:subfig:d}
\includegraphics[width=0.48\linewidth]{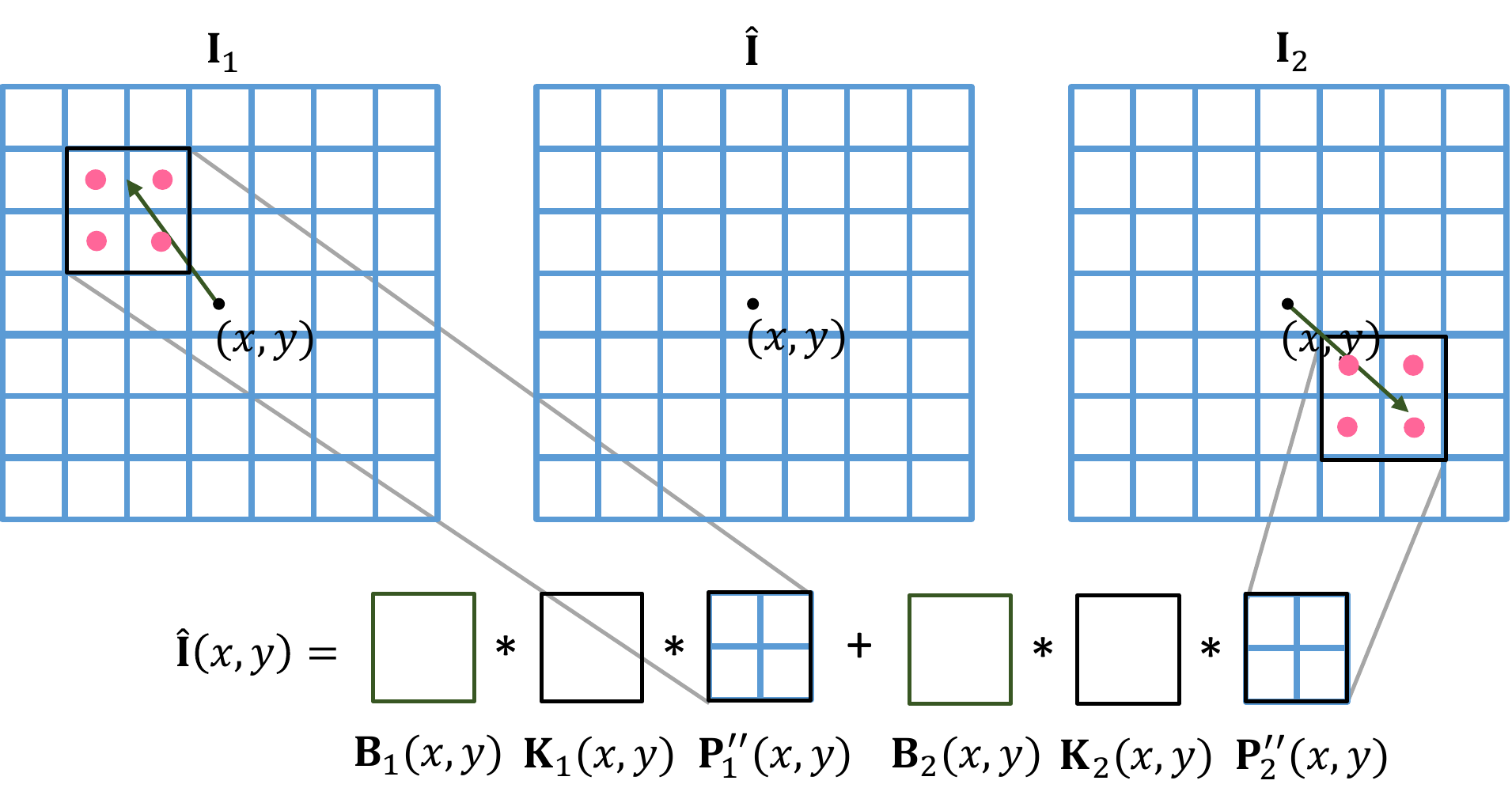}}
\caption{Illustration of the sampling locations (pink point) in a $7\times7$ checkerboard, in which the black square represents a convolution patch. In the center of the blue rectangle lattice, the sampling location is integer. (a) Baseline kernel-based methods with a $3\times3$ convolution patch. (b) Our method with a $3\times3$ convolution patch. (c) Conventional flow based methods. (d) Adaptive warping based methods with a $2\times2$ convolution patch.}
\label{fig1}
\end{figure*}
\subsection{Problem Statement}
To explore the relationships among kernel and flow based methods, we introduce our frame interpolation algorithm for single or multiple time steps individually.
\subsubsection{Single Frame Interpolation}

Assume that there are two temporally neighboring frames $\bm{\mathrm{I}}_1$ and $\bm{\mathrm{I}}_2$, our purpose is to interpolate frame $\hat{\bm{\mathrm{I}}}$ that in the midpoint of the them.
For each pixel $\hat{\bm{\mathrm{I}}}(x,y)$ to be synthesized, the widely used kernel-based interpolation model \cite{dnn, sepconv, adaconv} learns a pair of convolution kernels and uses them to convolve the local patches $\bm{\mathrm{P}}_1(x,y)$ and $\bm{\mathrm{P}}_2(x,y)$.
This process can be formulated as
\begin{equation}
\bm{\mathrm{\hat{I}}}(x,y)=\bm{\mathrm{K}}_1(x,y)*\bm{\mathrm{P}}_1(x,y)+\bm{\mathrm{K}}_2(x,y)*\bm{\mathrm{P}}_2(x,y),
\label{eq1}
\end{equation}
where $*$ means convolution operation and $\bm{\mathrm{K}}_1, \bm{\mathrm{K}}_2 \in \mathbb{R}^{n\times n}$ represent $n\times n$ 2D convolution kernels. Figure \ref{fig1:subfig:a} illustrates this kind of method. For standard local convolution, $n$ has to be big enough to capture large motion. For instance, in AdaConv \cite{adaconv}, the kernel size $n$ equals to 41. However, estimating such an amazing number of kernels ($41\times41$) simultaneously entails heavy computational load. In \cite{sepconv}, each 2D kernel is approximated with two 1D kernels $\langle\bm{\mathrm{k}}_{1,v},\bm{\mathrm{k}}_{1,h}\rangle$ or $\langle\bm{\mathrm{k}}_{2,v},\bm{\mathrm{k}}_{2,h}\rangle$ with formulation:
\begin{equation}
\left\{
\begin{aligned}
\bm{\mathrm{K}}_1(x,y) = \bm{\mathrm{k}}_{1,v}(x,y)\cdot \bm{\mathrm{k}}_{1,h}^{\top}(x,y) , \\
\bm{\mathrm{K}}_2(x,y) = \bm{\mathrm{k}}_{2,v}(x,y)\cdot \bm{\mathrm{k}}_{2,h}^{\top}(x,y) ,
\end{aligned}
\right.
\label{eq2}
\end{equation}
which helps to reduce the memory consumption from $O(n^2)$ to $O(2n)$. Nonetheless, despite thousands of pixels have been considered, these methods are limited to
motions up to $n$ pixels between two input frames.

To solve this problem, we propose to make convolution deformable by using much smaller convolution kernel size and learning additional offsets and masks. This allows us to focus on fewer but more relevant pixels rather than all the pixels in a large neighborhood.
Towards this end, the patches that filled with local pixels should be resampled by those pixels which mostly contribute to the final value.

Let $\bm{\mathrm{p}}_{i,j}$ denote the pre-specified offset for the $j$-th ($j\in[1, n^2]$) location in a specific patch and $i$ represents either of the two input frames.
Particularly, for an $n\times n$ convolution, the pre-specified offset are specified with a regular grid $\mathcal{R}$,

$$
\begin{aligned}
\mathcal{R} = \{&(-\frac{n-1}{2}, -\frac{n-1}{2}), (-\frac{n-1}{2}, -\frac{n-1}{2}+1), ..., \\
&(\frac{n-1}{2}-1, \frac{n-1}{2}), (\frac{n-1}{2}, \frac{n-1}{2})\}.
\end{aligned}
$$
In other words, $\bm{\mathrm{p}}_{\_,j}$ enumerates the locations in $\mathcal{R}$.
Moreover, with learned offset $\Delta\bm{\mathrm{p}}_{i,j}$ and modulation scalar $\Delta\bm{\mathrm{m}}_{i,j}$, the pixels in a resampled patch $\bm{\mathrm{P^{'}}}$ can be expressed as
\begin{equation}
\left\{
\begin{aligned}
\bm{\mathrm{P}}_1^{'}(x,y;\bm{\mathrm{p}}_{1,j}) = \bm{\mathrm{P}}_1(x,y;\bm{\mathrm{p}}_{1,j}+\Delta\bm{\mathrm{p}}_{1,j})\cdot \Delta\bm{\mathrm{m}}_{1,j} , \\
\bm{\mathrm{P}}_2^{'}(x,y;\bm{\mathrm{p}}_{2,j}) = \bm{\mathrm{P}}_2(x,y;\bm{\mathrm{p}}_{2,j}+\Delta\bm{\mathrm{p}}_{2,j})\cdot \Delta\bm{\mathrm{m}}_{2,j} .
\end{aligned}
\right.
\label{eq3}
\end{equation}

As the learned offsets are typically fractional, pixels located at non-integral coordinates are bilinearly sampled. Moreover, 1D separable kernels are used to calculate 2D convolution kernels in Eq. (\ref{eq2}) and we further introduce to learn pixel-wise residual values $\Delta\mathrm{b}(x,y)$ in case that the convolution kernels are less accurate.
Therefore, our final interpolation process is expressed as
\begin{equation}
\begin{aligned}
\bm{\mathrm{\hat{I}}}(x,y) =
&\bm{\mathrm{K}}_1(x,y)*\bm{\mathrm{P}}_1^{'}(x,y)+ \\
&\bm{\mathrm{K}}_2(x,y)*\bm{\mathrm{P}}_2^{'}(x,y)+ \Delta\mathrm{b}(x,y)\\
=
& \bm{\mathrm{k}}_{1,v}(x,y)\cdot \bm{\mathrm{k}}_{1,h}^{\top}(x,y)*\bm{\mathrm{P}}_1^{'}(x,y) + \\
& \bm{\mathrm{k}}_{2,v}(x,y)\cdot \bm{\mathrm{k}}_{2,h}^{\top}(x,y)*\bm{\mathrm{P}}_2^{'}(x,y) + \Delta\mathrm{b}(x,y).
\end{aligned}
\label{eq4}
\end{equation}

\subsubsection{Relationships with Kernel and Flow Based Methods}
\label{relationship}
In our method, both previous kernel-based methods \cite{sepconv, dnn} and conventional flow-based methods can be seen as specific instances of our approach. In Eqs. (\ref{eq3}) and (\ref{eq4}), it is easy to make out that when $\Delta\bm{\mathrm{p}}=\bm{0},\Delta\bm{\mathrm{m}}=\bm{1}$ and $\Delta{\mathrm{b}}=0$, the interpolation process is the same as those proposed in \cite{dnn, sepconv}.

\newcommand{\tabincell}[2]{\begin{tabular}{@{}#1@{}}#2\end{tabular}}
\begin{table}[t]
\centering
\caption{A list of conditions in which our method can be equivalent or similar to the other kinds of algorithms in terms of pixel reference.}
{
\begin{tabular}{lll}
\toprule
Type & Condition & Relationship\\
\midrule
Kernel based \cite{dnn, sepconv} & \tabincell{l}{$n=51, \Delta\bm{\mathrm{p}}=\bm{0},$\\ $\Delta\bm{\mathrm{m}}=\bm{1}, \Delta{\mathrm{b}}=0$} & Equivalence \\
\midrule
Flow based \cite{Toflow, DVF} & \tabincell{l}{$n=1, \Delta\bm{\mathrm{m}}=\bm{1},$\\ $\Delta{\mathrm{b}}=0$} & Equivalence\\
\midrule
Our previous work \cite{dsepconv} & $\Delta{\mathrm{b}}=0$ & Equivalence\\
\midrule
Adaptive warping \cite{MEMCNet, dain} & $n=4, \Delta\bm{\mathrm{m}}=\bm{\mathrm{B}}$ & Resemblance\\
\bottomrule
\end{tabular}}
\label{tab:2}
\end{table}
\begin{figure}[t]
\centering
\subfigure{
\begin{minipage}[t]{0.3\linewidth}
\centerline{\includegraphics[width=1.05\linewidth]{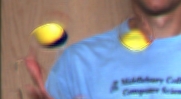}}
\centerline{t=0.1}
\end{minipage}
}
\subfigure{
\begin{minipage}[t]{0.3\linewidth}
\centerline{\includegraphics[width=1.05\linewidth]{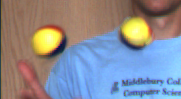}}
\centerline{t=0.5}
\end{minipage}
}
\subfigure{
\begin{minipage}[t]{0.3\linewidth}
\centerline{\includegraphics[width=1.05\linewidth]{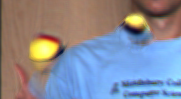}}
\centerline{t=0.9}
\end{minipage}
}
\centering
\caption{Modeling arbitrary time interpolation from networks trained for $t=0.5$. Despite correct pixels are chosen for synthesis, the occlusion is mistakenly solved, making the balls incomplete.}
\label{fig12}
\end{figure}
\begin{figure*}[t]
\centering
\includegraphics[width=\linewidth]{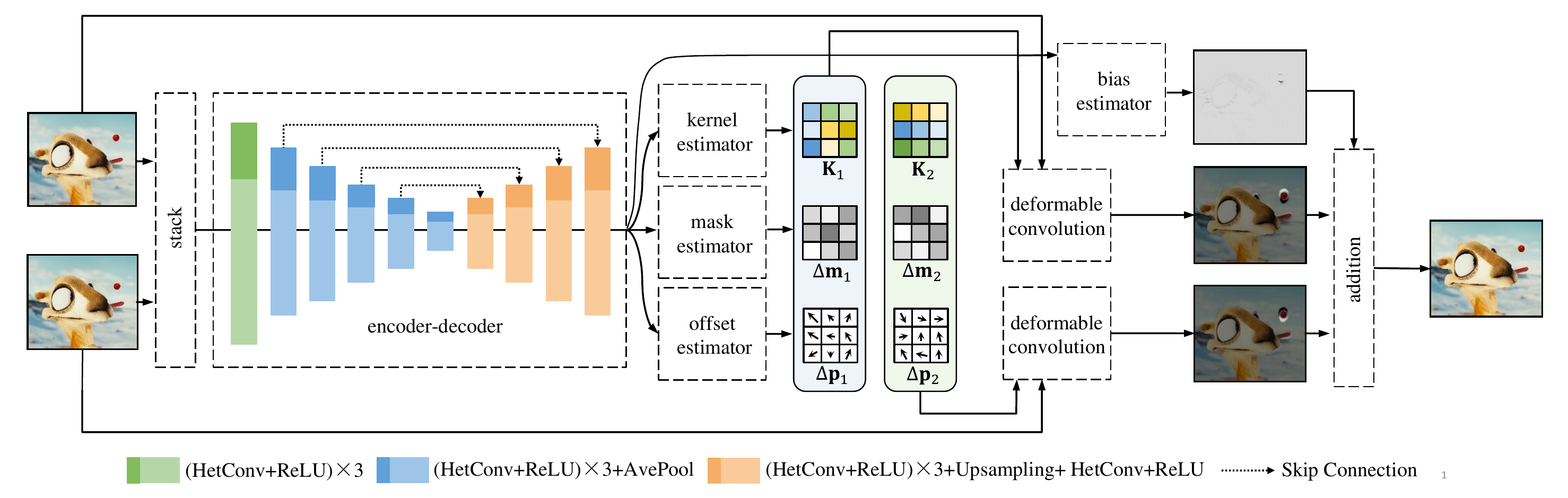}
\caption{Illustration of the architecture of our proposed EDSC network, which includes five major components: an encoder-decoder architecture and a set of kernel, mask, offset and bias estimators.
}
\label{fig2}
\end{figure*}
As for flow-based method, typically, the warping function can be formulated as
\begin{equation}
\begin{aligned}
\bm{\mathrm{\hat{I}}}(x,y)={\mathrm{k}}_1\cdot\bm{\mathrm{I}}_1(x + u_1,y + v_1)+{\mathrm{k}}_2\cdot\bm{\mathrm{I}}_2(x + u_2,y + v_2),
\label{eq5}
\end{aligned}
\end{equation}
where $\langle u_1, v_1\rangle$ and $\langle u_2, v_2\rangle$ denote the bidirectional optical flow values and $\mathrm{k_1}, \mathrm{k_1}$ represent occlusion masks.

In relation to the bilinear interpolation process in those flow based methods, as shown in Figure \ref{fig1:subfig:c}, we redefine the warping operation in Eq.(\ref{eq5}) as a $2\times 2$ pixel-wise convolution process with the formulation:

\begin{equation}
\bm{\mathrm{\hat{I}}}(x,y)=\mathrm{k_1}\cdot\bm{\mathrm{B}}_1(x,y)*\bm{\mathrm{P}}_1^{''}(x,y)+\mathrm{k_2}\cdot\bm{\mathrm{B}}_2(x,y)*\bm{\mathrm{P}}_2^{''}(x,y),
\label{eq6}
\end{equation}
where $\bm{\mathrm{B}}$ denote fixed bilinear interpolation coefficients and
$\bm{\mathrm{P}}_i^{''}(x,y)$ is calculated by:

\begin{equation}
\begin{aligned}
&\bm{\mathrm{P}}_i^{''}(x,y) = \bm{\mathrm{P}}_i(x,y;\bm{\mathrm{p}}_{i, j} + \Delta\bm{\mathrm{p}}_{i, j}), i\in[1, 2], j\in[1, 4], \\
&\left\{
\begin{aligned}
&\Delta\bm{\mathrm{p}}_{i,1} = (\lfloor u_i\rfloor, \lfloor v_i\rfloor)\\
&\Delta\bm{\mathrm{p}}_{i,2} = (\lfloor u_i\rfloor, \lfloor v_i\rfloor + 1)\\
&\Delta\bm{\mathrm{p}}_{i,3} = (\lfloor u_i\rfloor + 1, \lfloor v_i\rfloor)\\
&\Delta\bm{\mathrm{p}}_{i,4} = (\lfloor u_i\rfloor + 1, \lfloor v_i\rfloor + 1),\\
\end{aligned}
\right.
\end{aligned}
\label{eq7}
\end{equation}
where $\lfloor \cdot\rfloor$ represents floor operation.

In Eqs. (\ref{eq3}) and (\ref{eq4}), if we set $n=1, \Delta\bm{\mathrm{m}}=\bm{1}$ and $\Delta{\mathrm{b}}=0$, our interpolation process is the same as the one in Eq. (\ref{eq5}), indicating that the flow based method can be a specific case of our method.

We further show the adaptive warping method proposed in \cite{MEMCNet, dain} in Figure \ref{fig1:subfig:d}. When $\Delta\bm{\mathrm{m}}$ in Eq. (\ref{eq3}) equals to the bilinear interpolation coefficients $\bm{\mathrm{B}}$, our method bears some resemblance to the operation of adaptive warping. The difference is that the locations of pixels used to resample the convolutional patches $\bm{\mathrm{P}}^{'}$ can be dispersed, while those used in $\bm{\mathrm{P}}^{''}$ are restricted in a small square area.

In Table \ref{tab:2}, we summarize the main relationships and the conditions between our method and previous kernel-based \cite{dnn, sepconv}, flow-based \cite{Toflow, DVF}, adaptive warping based \cite{MEMCNet, dain} algorithms as well as our prior work \cite{dsepconv}.

\subsubsection{Multiple Frame Interpolation}
So far, none of the kernel-based interpolation methods can directly generate in-between frames at an arbitrary temporal position. This is because the pixels chosen for the final adaptive convolution are tied to a specific time step $t=0.5$. A possible solution is to resample the pixels based on $t$, which is easy for methods with learned offsets. For instance, given the model trained for $t=0.5$, we can respectively multiply the learned offsets $\Delta\bm{\mathrm{p}}_{1,j}$ and $\Delta\bm{\mathrm{p}}_{2,j}$ by $t/0.5$ and $(1-t)/(1-0.5)$ in Eq. (\ref{eq3}) to shift the locations of the reference pixels, producing an intermediate frame at arbitrary time $t$. However, as shown in Figure \ref{fig12}, this solution is problematic since the occlusion is still handled for $t=0.5$, indicating that the learned masks and kernels should also be controlled by $t$ for multiple video frame interpolation.

Followed a similar route in Eq. (\ref{eq4}), kernels, masks, offsets and biases are needed for multiple frame interpolation. The only difference is that intermediate time step $t$ is a crucial control variable in pixel synthesis. Thus, the formulation with respect to arbitrary time frame interpolation is:
\begin{equation}
\begin{aligned}
\bm{\mathrm{\hat{I}}}(x,y,t) =
& \bm{\mathrm{k}}_{1,v}(x,y,t)\cdot \bm{\mathrm{k}}_{1,h}^{\top}(x,y,t)*\bm{\mathrm{P}}_1^{'}(x,y,t) + \\
& \bm{\mathrm{k}}_{2,v}(x,y,1-t)\cdot \bm{\mathrm{k}}_{2,h}^{\top}(x,y,1-t)*\bm{\mathrm{P}}_2^{'}(x,y,1-t) \\
&+ \Delta\mathrm{b}(x,y).
\end{aligned}
\label{eq8}
\end{equation}

\subsection{Network Architecture}

We use a fully convolutional neural network modified from our baseline SepConv \cite{sepconv}. The whole network can be divided into the following submodules: the encoder-decoder architecture, kernel estimator, offset estimator, mask estimator and bias estimator as illustrated in Figure \ref{fig2}.
\subsubsection{Encoder-decoder Architecture}
Given two input frames, the encoder-decoder architecture aims to extract deep features for estimating kernels, masks, offsets and a bias value for each output pixel.

We use a U-Net structure as the backbone of our encoder-decoder module, where skip connections are employed to facilitate the feature mixture across encoder and decoder. We found that in SepConv \cite{sepconv} and DSepConv \cite{dsepconv}, the parameters in the encoder-decoder module occupy a large proportion (97\%) of the whole network, which can be reduced by leveraging HetConv \cite{hetconv} to replace the original standard convolution operation. A convolutional layer is said to be a HetConv layer if it contains different sizes of filters and more details about HetConv can be found in \cite{hetconv}. In our encoder-decoder architecture, all the standard convolution layers are modified with HetConv, in which 25\% of the filters are $3\times3$ and the others are $1\times1$. This modification helps to save about 79.6\% FLOPs in computation and 59.6\% parameters compared to our previous work \cite{dsepconv} without sacrificing the accuracy.

\subsubsection{Estimators}
\label{estimators}

\begin{figure}[t]
\centering
\includegraphics[width=0.7\linewidth]{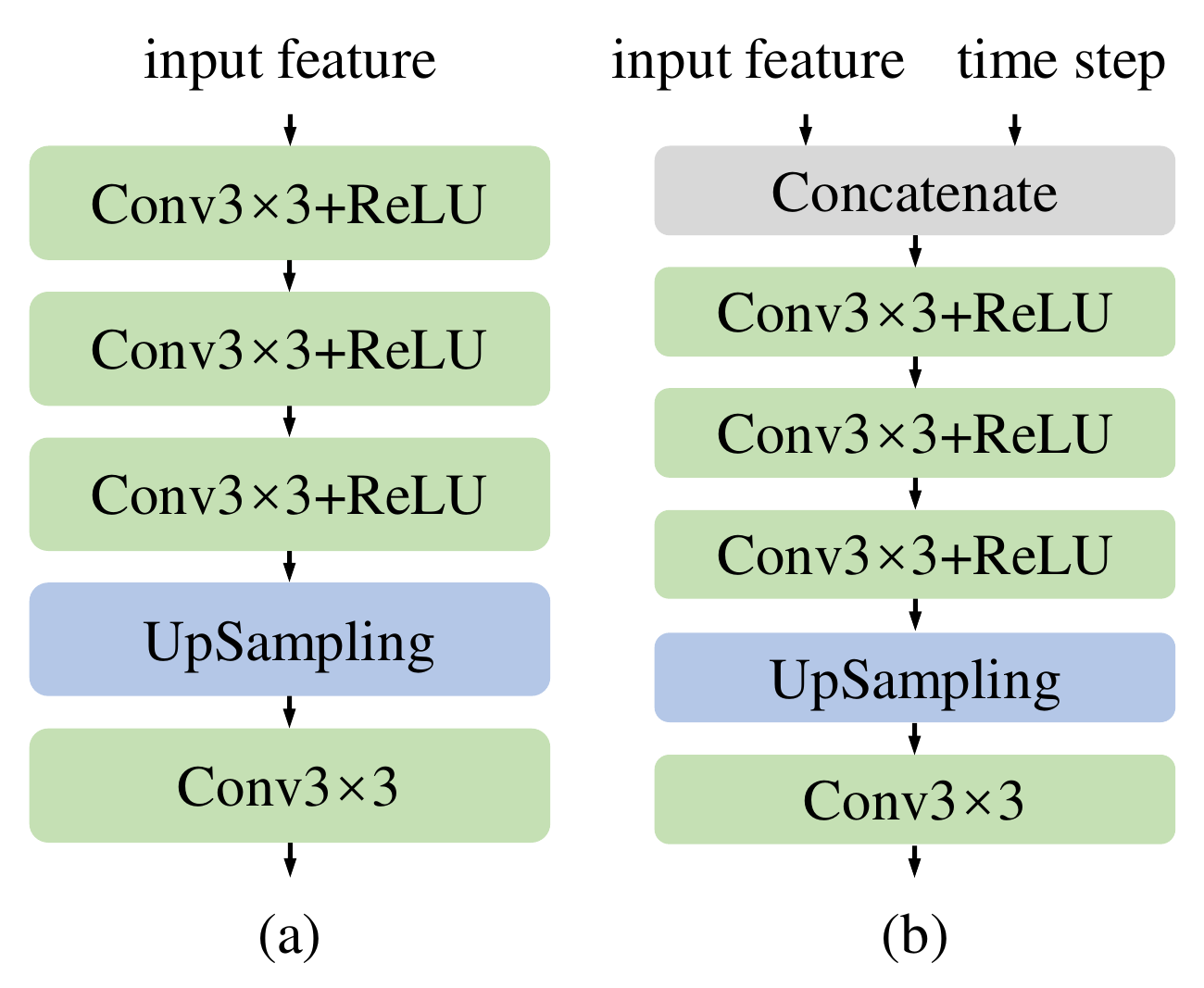}
\caption{Architecture of the sub-network of kernel estimator: (a) for single in-between frame generation and (b) multiple frame generation.
}
\label{fig3}
\end{figure}

\textbf{Kernel estimator.} The kernel estimator consists of four parallel sub-networks with analogous structure, which estimates adaptive vertical and horizontal 1D kernels for each pixel of the two frames. For each sub-network, shown in Figure. \ref{fig3} (a), three $3\times3$ convolution layers with Rectified Linear Units (ReLU) \cite{relu}, a bilinear upsampling layer and another $3\times3$ convolution layer are stacked, yielding a 3D tensor whose height and width match the frame resolution and whose depth equals the specific kernel size $n$ ($n=5$ in our case). The numbers of channels at different layers of the kernel estimator are \{64, 32, 32, 32, 5\} from top to bottom. Subsequently, the estimated four 1D kernels are used to calculate two 2D kernels described in Eq. (\ref{eq2}).

Noticeably, together with the information flow directed from encoder-decoder architecture, intermediate time step $t$ is fed into the kernel estimator as an extra channel as shown in Figure \ref{fig3} (b). We expand $t$ into a 3D tensor with one channel whose height and width are the same as the information flow.
We found that applying such an analogous coord-conv trick \cite{coordconv} for multiple video interpolation is effective. By concatenating an extra channel filled with (constant, untrained) time information, the learned kernel parameters can be tied to different time steps, making it possible for our method to generate arbitrary intermediate interpolation frames. Additionally, as described in Eq. (\ref{eq8}), we use $t$ and $1-t$ respectively to estimate kernels for the two input frames because they are not equally informative with different time steps.

\textbf{Offset estimator.} The offset estimator, sharing the same structure and inputs as the kernel one described above, contains four parallel sub-networks to learn two directional (vertical and horizontal) offsets for each location of the two frame patches. With a specific kernel size $n$, there are $n^2$ pixels in each regular grid patch. Hence, the depth of the output 3D tensors equals $n^2$. The channels at different layers of offset estimator are \{64, 32, 32, 32, 25\} from top to bottom.

\textbf{Mask estimator.} Inspired by \cite{dcn2}, learnable masks $\Delta\bm{\mathrm{m}}$ are introduced as a modulation mechanism that expands the scope of modeling and gives a significant improvement in performance. The design of mask estimator is similar, whose only difference is that the output channels are fed to a sigmoid layer. There are two parallel sub-networks, each of which produces tensors with $n^2$ channels.

\textbf{Bias estimator.} Though the estimators mentioned above could produce compelling interpolation results, there may be some blur or artifacts around the occlusion areas. We design an extra bias estimator to learn residual values for better pixel adaption. The bias estimator only takes the features from the encoder-decoder architecture as input and outputs a 3D tensor with 3 channels.


\subsubsection{Deformable Convolution}
The deformable convolution utilizes the estimated kernels, offsets and masks to adaptively convolve input frames, yielding an intermediate interpolation result.
Specifically, the deformation part for each convolutional patch is defined in Eq. (\ref{eq3}), with the convolution part in Eq. (\ref{eq4}). As depicted in Figure \ref{fig1:subfig:b}, deformable convolution is able to utilize information outside the local neighborhood.
Note that our operation is different from the process described in \cite{dcn2}, whose offsets and modulation scalars are obtained by applying a convolutional layer over the same input feature map and whose kernels share the same weights.
Instead, we individually learn these components for each pixel, making the synthesis process adaptive from pixel to pixel.
In the right part of Figure \ref{fig2}, the frames generated from deformable convolution look dimmer than the final interpolation result in brightness except area with occlusion (\emph{e.g.}, area around the red ball), suggesting the effectiveness of our method to handle motion and occlusion.

\subsection{Training}
\subsubsection{Loss functions}
We consider two kinds of loss functions to penalize the interpolated frame $\bm{\mathrm{\hat{I}}}$ that is not similar to the ground truth $\bm{\mathrm{I}}^{\mathrm{GT}}$.

The first loss measures the difference between the interpolated pixel color and the ground-truth color with the function:
\begin{equation}
\mathcal{L}_{C} = \rho(\bm{\mathrm{\hat{I}}} - \bm{\mathrm{I}}^{\mathrm{GT}}),
\label{eq9}
\end{equation}
\begin{equation}
\rho(x)=\sqrt{x^2+\epsilon^2},
\label{eq10}
\end{equation}
where $\rho(\cdot)$ represents the Charbonnier penalty function \cite{Charbonnier} and the constant $\epsilon$ is set to be 1$e$-6.

The second type of loss functions aims to penalize results that are not perceptually similar to ground truth by additionally defining a distance measure between high-level features extracted from a pre-trained network. The definition is as follows.
\begin{equation}
\mathcal{L}_{F} = \big{\Arrowvert}\phi(\bm{\mathrm{\hat{I}}})-\phi(\bm{\mathrm{I}}^{\mathrm{GT}})\big{\Arrowvert}_2^2,
\label{eq11}
\end{equation}
where $\phi$ represents the feature extractor based on the \texttt{relu4\_4} layer of the pre-trained VGG-19 network \cite{vgg19}. When training with this loss function, we use the model pre-trained on $\mathcal{L}_{C}$ loss function and then fine tune it with both $\mathcal{L}_{C}$ loss and $\mathcal{L}_{F}$ loss for 2 more epochs.

\begin{table*}[t]
\centering
\caption{Network setting comparison and analysis on different frame interpolation algorithms.}
\resizebox{\linewidth}{!}{
\begin{tabular}{lccccccccc}
\toprule
\multirow{2}*{Methods}
\multirow{3}{*}{ } &
\multirow{2}*{Venue}
\multirow{3}{*}{ } &
\multirow{2}*{\shortstack{Training\\dataset} }
\multirow{3}{*}{ } &
\multicolumn{6}{c}{ Sub-networks }
\multirow{3}{*}{ } &
\multirow{2}*{\shortstack{Parameters\\(million)} } \\

\cmidrule(lr){4-9}

& & & Flow  & Kernel(size) & Mask & Context & Other & Post-proc.  \\
\midrule
AdaConv \cite{adaconv} & CVPR'17 & proprietary & $\times$ & learned(41) & $\times$ &$\times$ &$\times$ &$\times$ & ---\\
SepConv \cite{sepconv} & ICCV'17 & proprietary & $\times$ & learned(51) & $\times$ &$\times$ &$\times$ &$\times$ & 21.6\\
IM-Net \cite{imnet} & CVPR'19 & proprietary & $\times$ & learned(25) & $\surd$ & $\times$ & $\times$ & $\surd$ & ---\\
DSepConv \cite{dsepconv} & AAAI'20 & Vimeo90K & $\times$ & learned(5) & $\surd$ & $\times$ &$\times$ &$\times$ & 21.8\\
AdaCoF \cite{adacof} & CVPR'20 & Vimeo90K & $\times$ & learned(5) & $\surd$ & $\times$ &$\times$ &$\times$ & 21.8\\
AdaCoF+ \cite{adacof} & CVPR'20 & Vimeo90K & $\times$ & learned(11) & $\surd$ & $\times$ &$\times$ &$\times$ & 22.9\\
\midrule
MEMC-Net$^*$ \cite{MEMCNet} & TPAMI'19 & Vimeo90K & FlowNetS & learned(4) & $\surd$ & ResNet & $\times$ & $\surd$ & 70.3\\
DAIN \cite{dain} & CVPR'19 & Vimeo90K & PWC-Net & learned(4) & $\times$ & Enc-Dec & Megadepth & $\surd$ & 24.0 \\
\midrule
DVF \cite{DVF} & ICCV'17 & UCF101 & Enc-Dec & bilinear(2) & $\surd$ & $\times$ & $\times$ & $\times$ & 1.6 \\
SuperSlomo \cite{superslomo} & CVPR'18 & Adobe240 & Enc-Dec & bilinear(2) & $\surd$ & $\times$ & $\times$ & $\times$ & 19.8 \\
CtxSyn \cite{Ctxsyn} & CVPR'18 & proprietary & PWC-Net &bilinear(2) & $\times$ & ResNet & $\times$ & $\surd$ & ---\\
ToFlow \cite{Toflow} & IJCV'19 & Vimeo90K & SpyNet & bilinear(2) & $\surd$ &  $\times$ & $\times$ & $\surd$ & 1.1\\
CyclicGen \cite{cyclicgen} & AAAI'19 & UCF101 & Enc-Dec & bilinear(2) & $\surd$ & $\times$ & HED & $\times$ & 3.0\\
CyclicGen+ \cite{cyclicgen} & AAAI'19 & UCF101, M.B. & Enc-Dec & bilinear(2) & $\surd$ & $\times$ & HED & $\times$ & 19.8\\
MS-PFT \cite{MSPFT} & TCSVT'20 & Vimeo90K & PWC-Net & $\times$ & $\times$ & $\times$ & $\times$ & $\times$ & 10.6\\
STAR-T$_{\mathrm{HR}}$ \cite{STAR} & CVPR'20 & Vimeo90K & Liu's & $\times$ & $\times$ & $\times$ & RBPN & $\times$ & 111.6\\
SoftSplat \cite{softsplat} \cite{softsplat} & CVPR'20 & Vimeo90K & PWC-Net & bilinear(2) & $\times$ & Pyramid & $\times$ & $\surd$ & ---\\
\midrule
CAIN \cite{cain} & AAAI'20 & Vimeo90K &$\times$ & $\times$ & $\times$ & $\times$ &$\times$ &$\times$ & 42.8\\
\midrule
EDSC(ours) & --- & Vimeo90K & $\times$ & learned(5) & $\surd$ & $\times$ &$\times$ &$\surd$ & 8.9\\
\bottomrule
\end{tabular}}
\label{tab:3}
\end{table*}
\subsubsection{Training Details.}
We trained two versions of our model: one produces only \emph{single} midpoint in time of the frames (EDSC\_s) and another generates \emph{multiple} intermediate frames (EDSC\_m) at arbitrary in-between time. The only difference between them is whether time information is involved in the estimators as detailed in section \ref{estimators}.
In addition, two kinds of loss functions were used for both models.

For EDSC\_s, we use Vimeo90K-Interp dataset \cite{Toflow}, which contains 51,312 triplets with a resolution of $256\times 448$ pixels. The triplets were randomly flipped horizontally or vertically for data augmentation. In the context of EDSC\_m, Vimeo90K-Septuplet dataset \cite{Toflow} is used instead because more consecutive frames are desired for multiple time step frame generation. The Vimeo90K-Septuplet dataset consists of 91,701 sequences with a resolution of $256\times 448$ pixels, each of which contains 7 consecutive frames.
 When training EDSC\_m, five target frames $\bm{\mathrm{\hat{I}}}_t, (t=0.167, 0.333, 0.5, 0.667, 0.833)$ are randomly generated.

The models were trained using Adam optimizer \cite{kingma2014adam}.
We first trained our network for 120 epochs using a learning rate schedule of 1$e$-4, dropping by half every 40 epochs. The training patch size was randomly cropped into $256\times 256$ pixels and the batch size was 4. Notice that some previous works trained their networks with large patches \cite{Ctxsyn,MEMCNet,dain}, we fine-tuned our network using the entire frames with learning rates of 1.25$e$-5 for another 10 epochs. This makes us use smaller batch size (which equals to 2) to deal with the increasing memory footprint.

\section{Experiments}
\label{experiments}
In this section, we first introduce the evaluation datasets and metrics. We then compare the proposed method with state-of-the-art algorithms. Finally, we perform comprehensive ablation studies to analyze the contribution of some important components.

\subsection{Experimental Setup}

\subsubsection{Datasets}
A wide variety of datasets are involved to evaluate our method.

\textbf{UCF101.} We use 379 triplets from UCF101 dataset \cite{UCF101} which were chosen by \cite{DVF}. The image resolution is $256\times256$ of pixels.

\textbf{Vimeo90K.} The Vimeo90K dataset \cite{Toflow} has been widely used for evaluation in video processing tasks. There are 3,782 triplets with a resolution of $448\times256$ pixels for video frame interpolation.

\textbf{Middlebury.} The Middlebury dataset \cite{middleburry} contains an \texttt{Evaluation} set (8 sequences, hidden ground truth) and an \texttt{Other} set (12 sequences, with ground truth), with maximum resolution of $640\times480$ pixels.


\textbf{SNU-FILM.} The SNU-FILM dataset \cite{cain} is based on high frame rate videos including videos from GOPRO test set \cite{GOPRO} and YouTube. The evaluation set contains four subsets: \texttt{Easy}, \texttt{Medium}, \texttt{Hard} and \texttt{Extreme} with different degrees of motions, each of which consists 310 triplets. The maximum resolution of this dataset is $1280\times720$ pixels.

\subsubsection{Metrics}
For quantitative evaluation, we use Peak Signal-to-Noise Ratio (PSNR), Structural Similarity Index (SSIM \footnote{Please note that we use the $\mathrm{MATLAB}$ function $\texttt{ssim()}$ for computing the SSIM metric.}) \cite{SSIM}
and Learned Perceptual Image Patch Similarity (LPIPS) \cite{LPIPS} metrics. In addition, we report the average Interpolation Error (IE) on the Middlebury dataset.
Bigger PSNR and SSIM indicates better performance, while for LPIPS and IE, the smaller, the better.

\subsubsection{Baselines}
\label{baselines}
We compare and analyze our method with most of the recent state-of-the-art interpolation methods since 2017. We divide these methods into four categories:
1) those with a component of learned convolutional kernel estimation (kernel based);
2) those with components of both optical flow and learned convolutional kernel estimation (adaptive warping based);
3) those with a component of optical flow or its variations’ estimation (flow based);
4) those without any components of optical flow or adaptive convolutional kernel estimation.

For the first category, we typically choose kernel-based interpolation methods, including AdaConv \cite{adaconv}, SepConv \cite{sepconv}, IM-Net \cite{imnet}, DSepConv \cite{dsepconv} and AdaCoF \cite{adacof}. Our method belongs to this category as well.
The second category contains MEMC-Net$^*$ \cite{MEMCNet} and DAIN \cite{dain}.
The third category includes DVF \cite{DVF}, SuperSlomo \cite{superslomo}, CtxSyn \cite{Ctxsyn}, ToFlow \cite{Toflow}, CyclicGen \cite{cyclicgen}, MS-PFT \cite{MSPFT}, STAR-T$_{\mathrm{HR}}$ \cite{STAR} as well as SoftSplat \cite{softsplat}.
Additionally, we include CAIN \cite{cain} which makes use of PixelShuffle and attention mechanism in the fourth category.

Notably, considering that some methods provide more than one version of the same model, we report all their performances and treat them differently (\emph{e.g.}, CtxSyn, SepConv and SoftSplat are trained with two kinds of loss functions. CyclicGen and AdaCoF provides two models).

\subsection{Comparisons with state-of-the-arts}
Since most of the baselines focus on single-frame interpolation, we here first discuss our EDSC\_s model in section \ref{single} and the EDSC\_m model will be discussed in section \ref{arbitrary}.
\subsubsection{Network setting comparisons}
We analyse and report different network settings contributed to interpolation algorithms with following components: training dataset, sub-networks and model parameters shown in Table \ref{tab:3}. The Middlebury dataset \cite{middleburry} is abbreviated by M.B. for the sake of simplicity.
We further divide the sub-networks into different parts: flow, kernel, mask, context estimation networks as well as post-processing networks (abbreviated by post-proc.). Specially, networks for learning information that falls outside the mentioned five modules will be categorized as ``Other" class.

In Table \ref{tab:3}, the column ``Flow" specifies which methods are based on a pre-inferred displacement fields such as SpyNet \cite{spynet}, PWC-Net \cite{PWCNet}, FlowNetS \cite{flownet} and Liu's method \cite{liu2009beyond}. A self-defined encoder-decoder structure is abbreviated by Enc-Dec.
``Kernel(size)" refers to whether an algorithm utilizes adaptive kernels with a specific size. Noticeably, the bilinear interpolation for backward or forward warping guided by optical flow can be viewed as using fixed bilinear convolutional kernels with size of 2. ``Mask" determines whether some occlusion or visibility maps are performed.
``Context"
specifies whether contextual features are involved together with input frames. ``Other" indicates that whether other information is leveraged for video frame interpolation. For instance, CyclicGen \cite{cyclicgen} makes use of edge information extracted by HED \cite{HED}; STAR-T$_{\mathrm{HR}}$ \cite{STAR} utilizes pre-trained RBPN \cite{RBPN} as sub-network; DAIN \cite{dain} employs and fine-tunes pre-trained depth estimation network MegaDepth \cite{megadepth}.
In addition, ``Post-proc." refers to whether any post-processing networks applied on generated frames are performed. As the biases learned from bias estimator are added to the convolved intermediate frame pixels, we categorize it into ``Post-proc." based on this functionality. It is noteworthy that our bias estimator is specifically designed for non-flow kernel-based method and is a counterpart to the bias term in a convolutional layer. Compared with other post-processing networks such as the one used in MEMC-Net$^*$ \cite{MEMCNet} and the GridNet \cite{gridnet} used in CtxSyn \cite{Ctxsyn}, our bias estimator does not learn from the warped frames, contexture features or optical flows. Thus it is quite simpler, with little computational cost.
\begin{table*}[th]
\centering
\caption{Quantitative comparisons against methods using adaptive convolutional kernels.The numbers in \textbf{bold} and with an \underline{underline} represent the first and the second best performance, respectively.}
\resizebox{\linewidth}{!}{
\begin{threeparttable}
\begin{tabular}{lcccccccccc}
\toprule
\multirow{2}*{Methods}
\multirow{3}{*}{ } &
\multirow{2}*{\shortstack{Training\\dataset} }
\multirow{3}{*}{ } &
\multicolumn{3}{c}{ UCF101 \cite{UCF101, DVF} } &
\multicolumn{3}{c}{ Vimeo90K \cite{Toflow} } &
\multicolumn{2}{c}{ M.B.-\texttt{Other} \cite{middleburry} } &
\multirow{2}*{\shortstack{Parameters\\(million)} } \\

\cmidrule(lr){3-5}
\cmidrule(lr){6-8}
\cmidrule(lr){9-10}

& & PSNR $\uparrow$ & SSIM $\uparrow$& LPIPS $\downarrow$& PSNR $\uparrow$& SSIM $\uparrow$& LPIPS $\downarrow$& IE $\downarrow$& LPIPS $\downarrow$ \\
\midrule
AdaConv \cite{adaconv} & proprietary & --- & --- & --- & $^\dagger$32.33 & $^\dagger$0.957 & --- & --- & --- & ---\\
SepConv-$\mathcal{L}_{1}$ \cite{sepconv} & proprietary & 34.79 & 0.967 & 0.029 & 33.80 & 0.970 & 0.027 & 2.27 & 0.017 & 21.6 \\
SepConv-$\mathcal{L}_{F}$ \cite{sepconv} & proprietary & 34.69 & 0.966 & \underline{0.024} & 33.45 & 0.967 & \underline{0.019} & 2.44 & \underline{0.013} & 21.6 \\
IM-Net \cite{imnet} & proprietary & --- & --- & --- & $^\ddagger$33.50 & --- & --- & --- & --- & ---\\
DSepConv \cite{dsepconv} & Vimeo90K & \underline{35.08} & \textbf{0.969} & 0.030 & \underline{34.73} & 0.974 & 0.028 & 2.06 & 0.022 & 21.8\\
AdaCoF \cite{adacof} & Vimeo90K & 34.91 & \underline{0.968} & 0.029 & 34.27 & 0.971 & 0.031 & 2.31 & 0.029 & 21.8\\
AdaCoF+ \cite{adacof} & Vimeo90K & 34.90 & \underline{0.968}  & 0.030 & 34.47 & 0.973 & 0.029 & 2.23 & 0.026& 22.9\\
MEMC-Net$^*$ \cite{MEMCNet} & Vimeo90K & 35.01 & \underline{0.968} & 0.030 & 34.40 & 0.974 & 0.027 & 2.10 & 0.020 & 70.3 \\
DAIN \cite{dain} & Vimeo90K & 35.00 & \underline{0.968} & 0.028 & 34.72 & \textbf{0.976} & 0.022 & \underline{2.04} & 0.017 & 24.0\\
\midrule
EDSC\_s-$\mathcal{L}_{C}$ & Vimeo90K & \textbf{35.13} & \underline{0.968} & 0.029 & \textbf{34.84} & \underline{0.975} & 0.026 & \textbf{2.02} & 0.020 & 8.9\\
EDSC\_s-$\mathcal{L}_{F}$ & Vimeo90K & 34.78 & 0.967 & \textbf{0.023} & 34.49 & 0.972 & \textbf{0.016} & 2.15 & \textbf{0.010} & 8.9\\
\bottomrule
\end{tabular}
\begin{tablenotes}
\footnotesize
\raggedleft
\item[$\dagger$]: Results copied from \cite{Toflow}. $^\ddagger$:  Results copied from \cite{imnet}.
\end{tablenotes}
\end{threeparttable}}
\label{tab:5}
\end{table*}

\subsubsection{Single intermediate frame interpolation}
\label{single}

We first perform quantitative comparisons on the three common datasets against the state-of-the-art frame interpolation methods. Additionally, we divide these methods into two types according to whether they make use of adaptive convolutional kernels in Tables \ref{tab:5} and \ref{tab:4}, respectively.
The first type corresponds to kernel based and adaptive warping based methods and the second type corresponds to the last two categories mentioned in section \ref{baselines}.
We make substantial effort to guarantee that all the performances are tested under the same metrics, including their implementation details. For methods whose open source implementations from the respective authors are not completely publicly available, we copy the performance results from corresponding papers under the confirmation of the baseline metrics reported the same as our calculation.

We compare approaches which incorporate adaptive kernel estimations and the results are shown in Table \ref{tab:5}. Among all the methods, our $\mathcal{L}_{C}$-trained model achieves the best performance in terms of PSNR and IE and our $\mathcal{L}_{F}$-trained model performs the best in terms of LPIPS. In particular, we achieve 0.12 dB and 0.13 dB gain in terms of PSNR on the UCF101 and Vimeo90K datasets compared to DAIN \cite{dain}, without relying on pre-trained sub-models like PWC-Net \cite{PWCNet} and MegaDepth \cite{megadepth}. Additionally, we can see that our $\mathcal{L}_{C}$-trained model outperforms AdaCoF+ \cite{adacof} by 0.23 dB on UCF101 and 0.37 dB on Vimeo90K in terms of PSNR, whilst requiring 61\% fewer parameters.

\begin{figure}[t]
\centering
\vspace{-2mm}

\includegraphics[width=1\linewidth]{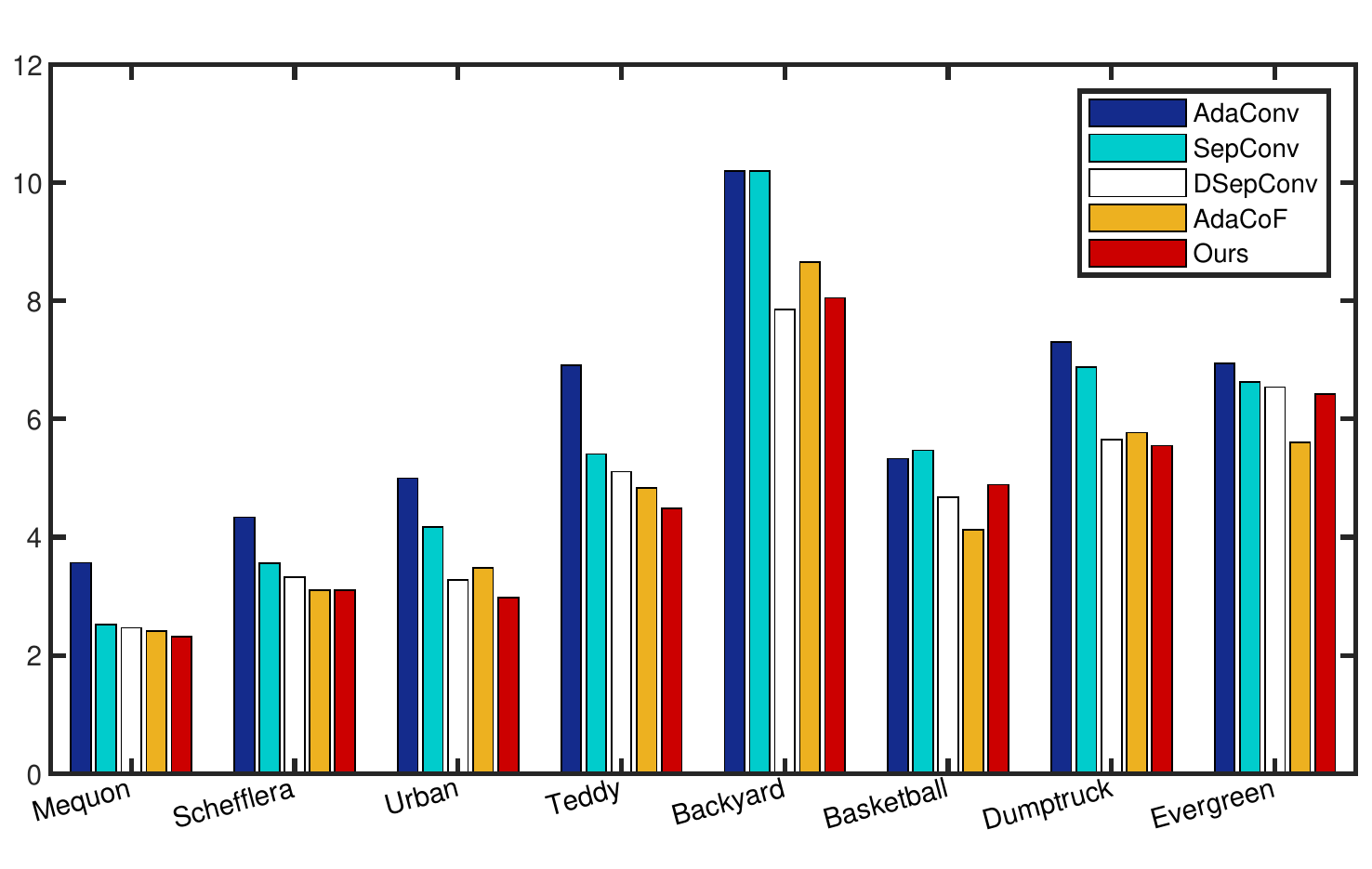}
\caption{Interpolation error comparisons among kernel based methods on the Middlebury \texttt{Evaluation} set \cite{middleburry}. Lower bars represent better performances.
}
\label{fig4_5}
\end{figure}

We also submit the interpolation results of our $\mathcal{L}_{C}$-trained model on \texttt{Evaluation} set to the Middlebury benchmark\footnote{\url{http://vision.middlebury.edu/flow/eval/results/results-i1.php}}. According to the feedback from the benchmark organizer, our approach ranks $3^{\mathrm{rd}}$ in terms of IE and $2^{\mathrm{nd}}$ in terms of NIE among all published algorithms at the time of submission. We specifically show the comparisons among kernel based methods which do not rely on any other information in Figure \ref{fig4_5}.
Among these methods, our model performs the best on 5 out of 8 sequences and achieves the best performance on average
, which demonstrates the good generalization ability of our method.

In what follows we compare methods that make no use of adaptive convolutional kernels. As shown in Table \ref{tab:4}, we can learn the fact that the usage of off-the-shelf and pre-trained model provides good performance. It is true that our method is inferior to STAR-T$_{\mathrm{HR}}$ \cite{STAR} and SoftSplat \cite{softsplat}. Combined the summary reported in Table \ref{tab:3}, STAR-T$_{\mathrm{HR}}$ \cite{STAR} additionally utilizes pre-inferred displacement fields \cite{liu2009beyond} and pre-trained RBPN \cite{RBPN}, leading to a reasonable performance with enormous parameters (which is more than $11\times$ bigger than our method). For SoftSplat \cite{softsplat}, they have reached the best performance so far due to their effectively handling cases where multiple source pixels map to the same target location, conditioned on pre-calculated optical flow \cite{PWCNet}. Nonetheless, we can see that a good kernel learner competitive without relying on any other extra information (like context, flow or edge information). In addition to STAR-T$_{\mathrm{HR}}$ \cite{STAR} and SoftSplat \cite{softsplat}, our $\mathcal{L}_{C}$-trained model convincingly outperforms the other methods in terms of most of the PSNR, SSIM and IE whereas our $\mathcal{L}_{F}$-trained model performs the best in terms of LPIPS.

\begin{table*}[t]
\centering
\caption{Quantitative comparisons against methods without using adaptive convolutional kernels. The numbers in \textbf{bold} and with an \underline{underline} represent the first and the second best performance, respectively.}
\resizebox{\linewidth}{!}{
\begin{threeparttable}
\begin{tabular}{lcccccccccc}
\toprule
\multirow{2}*{Methods}
\multirow{3}{*}{ } &
\multirow{2}*{\shortstack{Training\\dataset} }
\multirow{3}{*}{ } &
\multicolumn{3}{c}{ UCF101 \cite{UCF101, DVF} } &
\multicolumn{3}{c}{ Vimeo90K \cite{Toflow} } &
\multicolumn{2}{c}{ M.B.-\texttt{Other} \cite{middleburry} } &
\multirow{2}*{\shortstack{Parameters\\(million)} } \\

\cmidrule(lr){3-5}
\cmidrule(lr){6-8}
\cmidrule(lr){9-10}

& & PSNR $\uparrow$  & SSIM $\uparrow$& LPIPS $\downarrow$& PSNR $\uparrow$& SSIM $\uparrow$& LPIPS $\downarrow$& IE $\downarrow$& LPIPS $\downarrow$ \\
\midrule
DVF \cite{DVF} & UCF101 & $^\dagger$34.12 & $^\dagger$0.963 & --- & $^\dagger$31.54 & $^\dagger$0.946 & --- & $^\dagger$4.04 & --- & 1.6\\
SuperSlomo \cite{superslomo} & Adobe240 & $^\dagger$34.75 & $^\dagger$\underline{0.968} & --- & $^\dagger$33.15 & $^\dagger$0.966 & --- & $^\dagger$2.28 &---& 19.8\\
CtxSyn-$\mathcal{L}_{Lap}$ \cite{Ctxsyn} & proprietary & $^\ddagger$34.62 & --- & $^\ddagger$0.031 & $^\ddagger$34.39 & --- & $^\ddagger$0.024 & --- & $^\ddagger$0.016 & ---\\
CtxSyn-$\mathcal{L}_{F}$ \cite{Ctxsyn} & proprietary & $^\ddagger$34.01 & --- & $^\ddagger$0.024 & $^\ddagger$33.76 & --- & $^\ddagger$0.017 & --- & $^\ddagger$0.013 & ---\\
ToFlow \cite{Toflow} & Vimeo90K & 34.58 & 0.967 & 0.027 & 33.73 & 0.968 & 0.027 & 2.51 & 0.024 & 1.1 \\
CyclicGen \cite{cyclicgen} & UCF101 & 35.11 & \underline{0.968} & 0.030 & 32.10 & 0.949 & 0.058 & 2.86 & 0.046 & 3.0\\
CyclicGen+ \cite{cyclicgen} & UCF101, M.B. & 34.69 & 0.966 & 0.034 & 31.46 & 0.940 & 0.060 & 3.04 & 0.053 & 19.8 \\
MS-PFT \cite{MSPFT} & Vimeo90K & 34.70 & 0.967 &\underline{0.023} & 34.26 & 0.971 & 0.020 & 2.28 & 0.014 & 10.6 \\
STAR-T$_{\mathrm{HR}}$ \cite{STAR} & Vimeo90K & \underline{35.17} & \textbf{0.969} & 0.030 & 35.14 & \textbf{0.976} & 0.026 & \textbf{1.95} & --- & 111.6 \\
SoftSplat-$\mathcal{L}_{Lap}$ \cite{softsplat} & Vimeo90K & $^\ddagger$\textbf{35.39} & --- & $^\ddagger$0.033 & $^\ddagger$\textbf{36.10} & --- & $^\ddagger$0.021 & --- & $^\ddagger$0.016 & ---\\
SoftSplat-$\mathcal{L}_{F}$ \cite{softsplat} & Vimeo90K & $^\ddagger$35.10 & --- & $^\ddagger$\textbf{0.022} & $^\ddagger$\underline{35.58} & --- & $^\ddagger$\textbf{0.013} & --- & $^\ddagger$\textbf{0.008} & ---\\
CAIN \cite{cain} & Vimeo90K & $^\dagger$34.91 & $^\dagger$\textbf{0.969} & 0.032 & $^\dagger$34.65 & $^\dagger$0.973 & 0.031 & $^\dagger$2.28 & 0.025 & 42.8 \\
\midrule
EDSC\_s-$\mathcal{L}_{C}$ & Vimeo90K & 35.13 & \underline{0.968} & 0.029 & 34.84 & \underline{0.975} & 0.026 & \underline{2.02} & 0.020 & 8.9\\
EDSC\_s-$\mathcal{L}_{F}$ & Vimeo90K & 34.78 & 0.967 & \underline{0.023} & 34.49 & 0.972 & \underline{0.016} & 2.15 & \underline{0.010} & 8.9\\
\bottomrule
\end{tabular}
\begin{tablenotes}
\footnotesize
\raggedleft
\item[$\dagger$]: Results copied from \cite{cain}. $^\ddagger$:  Results copied from \cite{softsplat}.
\end{tablenotes}
\end{threeparttable}}
\label{tab:4}
\end{table*}
For qualitative comparisons, we compare our method against interpolation methods published since the year of 2019, including MEMC-Net$^*$ \cite{MEMCNet}, CyclicGen \cite{cyclicgen}, ToFlow \cite{Toflow}, DAIN \cite{dain}, STAR-T$_{\mathrm{HR}}$ \cite{STAR} ,CAIN \cite{cain}, DSepConv \cite{dsepconv} as well as AdaCoF \cite{adacof}.

\begin{figure*}[th]
\centering
\vspace{-3mm}
\subfigure[Overlayed.]{\label{fig4:subfig:a}
\includegraphics[width=0.1575\linewidth]{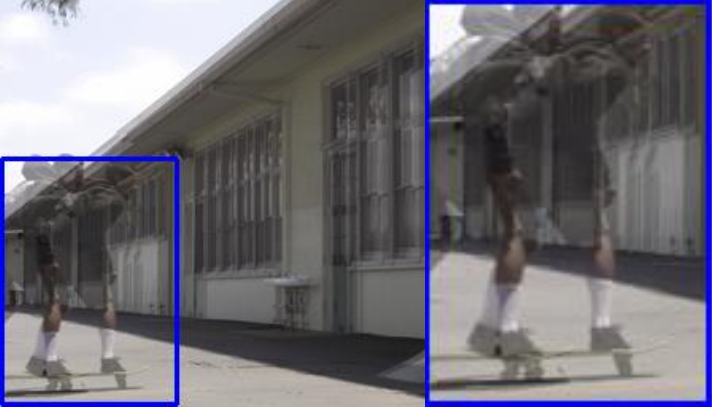}}
\subfigure[MEMC-Net$^*$.]{\label{fig4:subfig:b}
\includegraphics[width=0.1575\linewidth]{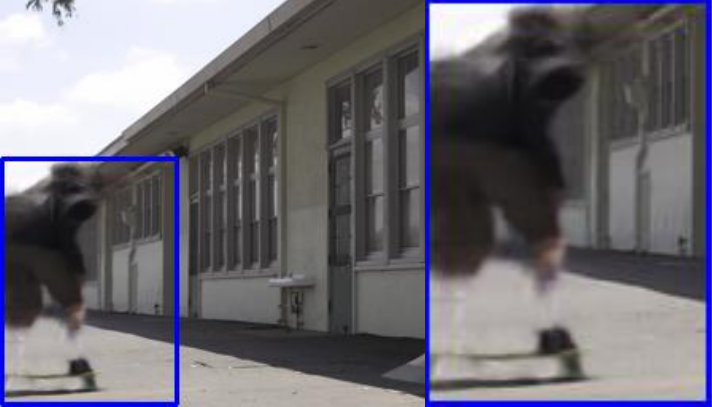}}
\subfigure[CyclicGen.]{\label{fig4:subfig:c}
\includegraphics[width=0.1575\linewidth]{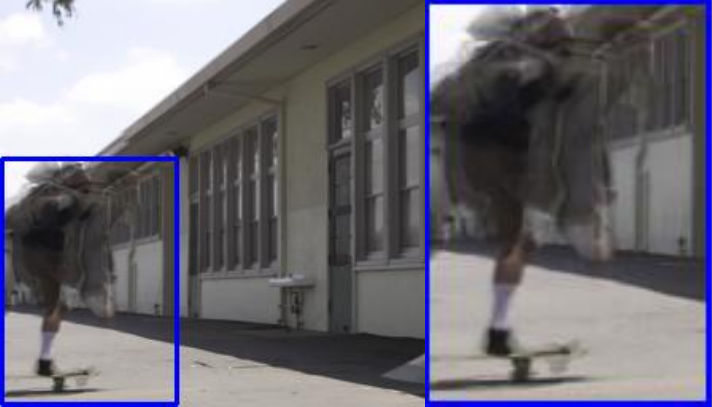}}
\subfigure[ToFlow.]{\label{fig4:subfig:d}
\includegraphics[width=0.1575\linewidth]{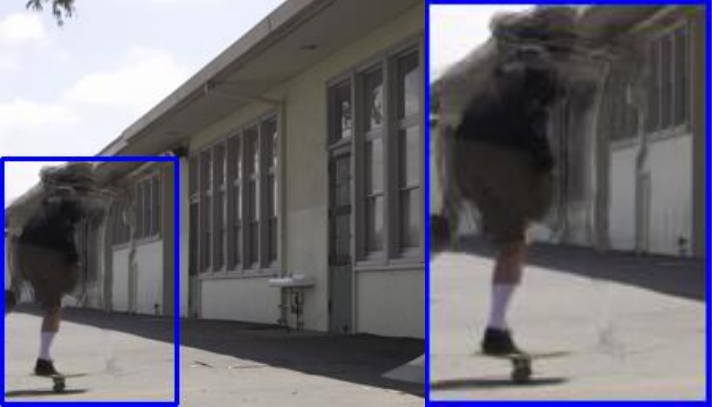}}
\subfigure[DAIN.]{\label{fig4:subfig:e}
\includegraphics[width=0.1575\linewidth]{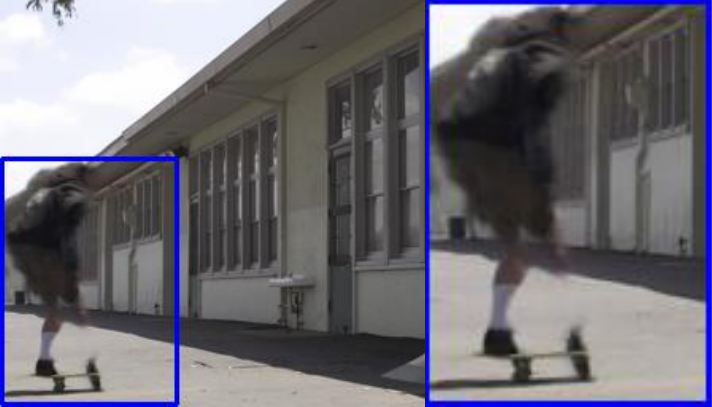}}
\subfigure[STAR-T$_{\mathrm{HR}}$.]{\label{fig4:subfig:f}
\includegraphics[width=0.1575\linewidth]{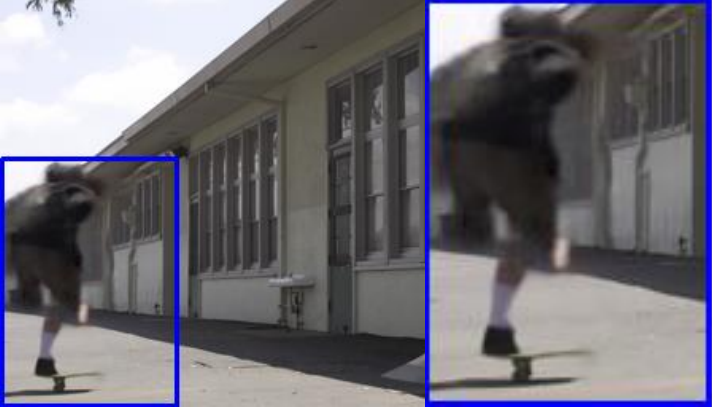}}

\subfigure[CAIN.]{\label{fig4:subfig:g}
\includegraphics[width=0.1575\linewidth]{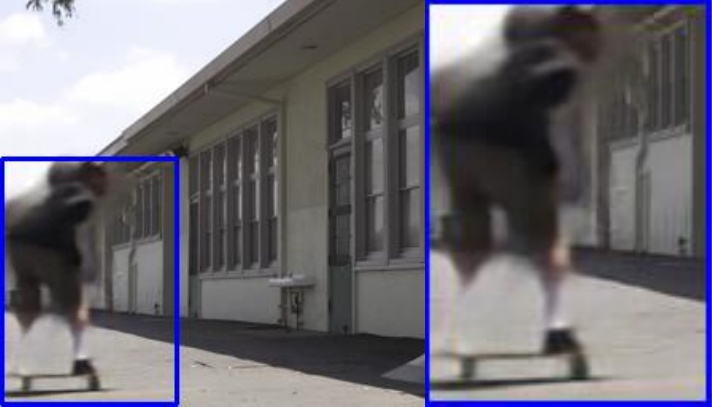}}
\subfigure[DSepConv.]{\label{fig4:subfig:h}
\includegraphics[width=0.1575\linewidth]{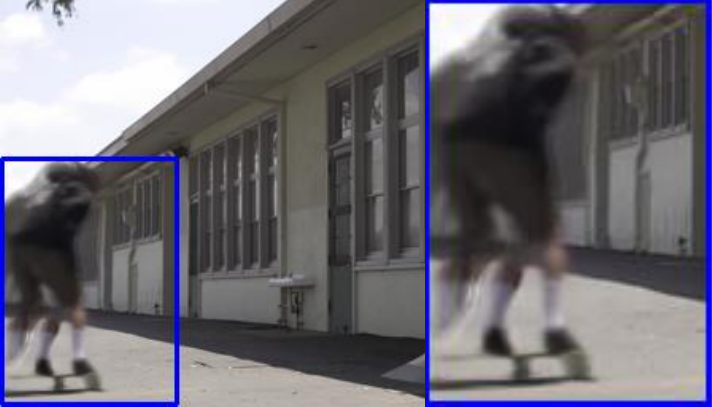}}
\subfigure[AdaCoF.]{\label{fig4:subfig:i}
\includegraphics[width=0.1575\linewidth]{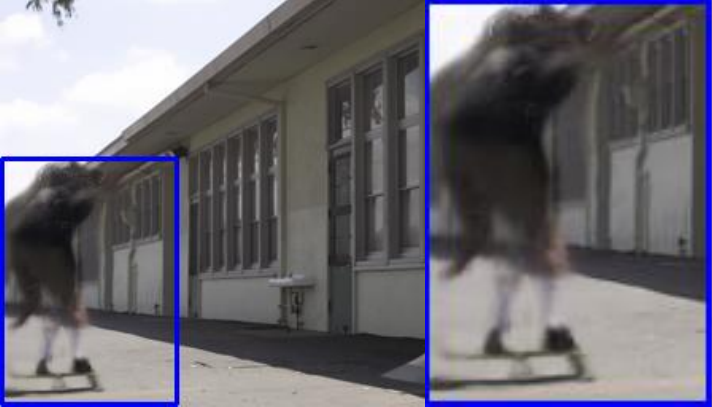}}
\subfigure[Ours-$\mathcal{L}_{C}$.]{\label{fig4:subfig:j}
\includegraphics[width=0.1575\linewidth]{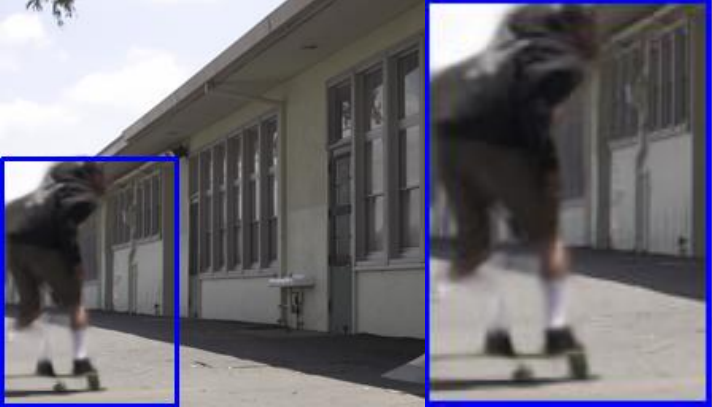}}
\subfigure[Ours-$\mathcal{L}_{F}$.]{\label{fig4:subfig:l}
\includegraphics[width=0.1575\linewidth]{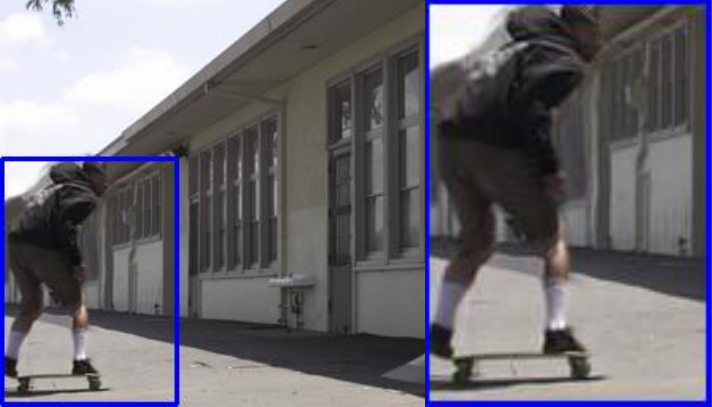}}
\subfigure[Ground Truth.]{\label{fig4:subfig:l}
\includegraphics[width=0.1575\linewidth]{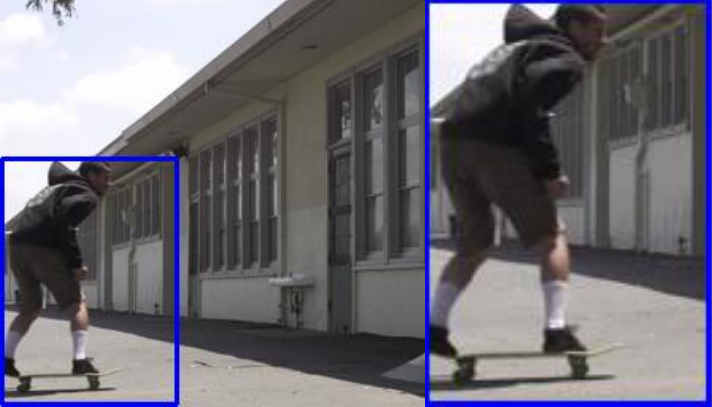}}
\vspace{-3mm}
\caption{Qualitative evaluation on a video with significant object motion.
}
\label{fig4}
\end{figure*}

In Figure \ref{fig4}, we show an example of a skateboarder playing in front of a building. From the overlayed frame in Figure \ref{fig4:subfig:a}, we can see that only one leg is shown in one frame while both two legs can be seen in the other, making it difficult to estimate optical flow accurately. Therefore, methods that make use of optical flow (like MEMC-Net$^*$, ToFlow, DAIN, STAR-T$_{\mathrm{HR}}$) generate visible blur or artifacts. Since the scene is also complex, the attention equipped method CAIN cannot blend the content of source images well and loses some information around the left shin. The results from DSepConv and AdaCoF contain exhibit blurriness as a result of inaccurate kernel learning. Our $\mathcal{L}_{C}$-trained model suffers from some information lost, whereas our result from $\mathcal{L}_{F}$-trained model appears clear with fewer visual distortions.

\begin{figure*}[!ht]
\centering
\vspace{-3mm}
\subfigure[Overlayed.]{\label{fig5:subfig:a}
\includegraphics[width=0.1575\linewidth]{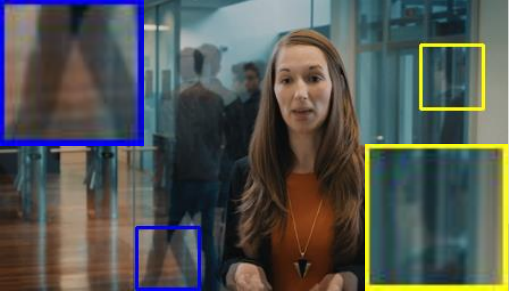}}
\subfigure[MEMC-Net$^*$.]{\label{fig5:subfig:b}
\includegraphics[width=0.1575\linewidth]{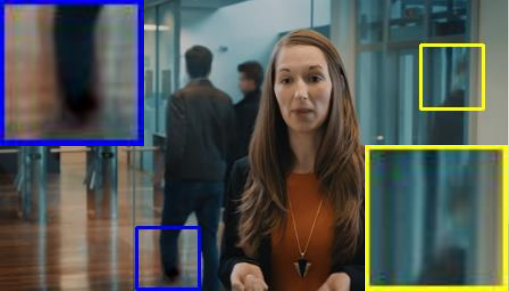}}
\subfigure[CyclicGen.]{\label{fig5:subfig:c}
\includegraphics[width=0.1575\linewidth]{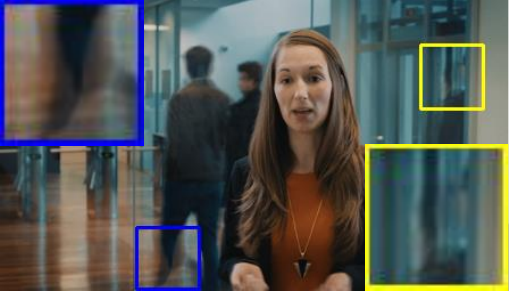}}
\subfigure[ToFlow.]{\label{fig5:subfig:d}
\includegraphics[width=0.1575\linewidth]{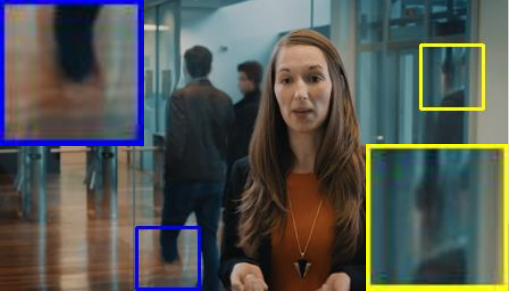}}
\subfigure[DAIN.]{\label{fig5:subfig:e}
\includegraphics[width=0.1575\linewidth]{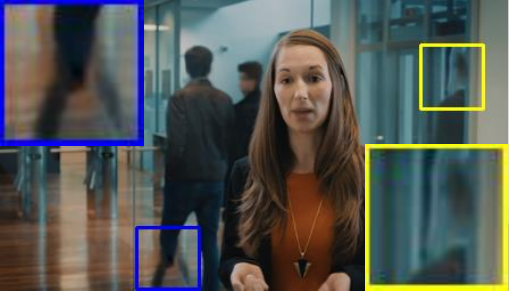}}
\subfigure[STAR-T$_{\mathrm{HR}}$.]{\label{fig5:subfig:f}
\includegraphics[width=0.1575\linewidth]{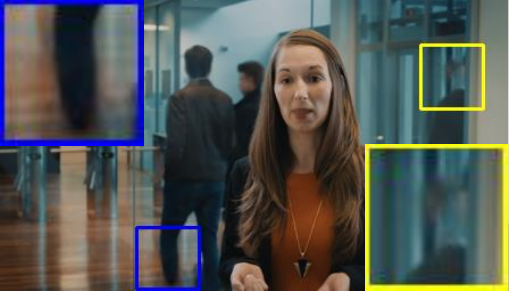}}

\subfigure[CAIN.]{\label{fig5:subfig:g}
\includegraphics[width=0.1575\linewidth]{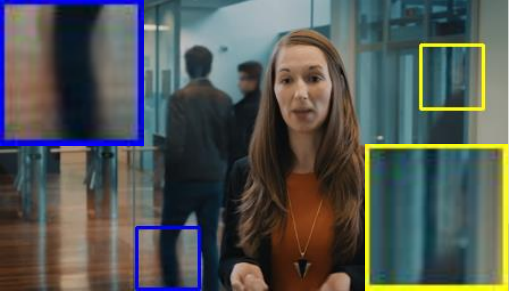}}
\subfigure[DSepConv.]{\label{fig5:subfig:h}
\includegraphics[width=0.1575\linewidth]{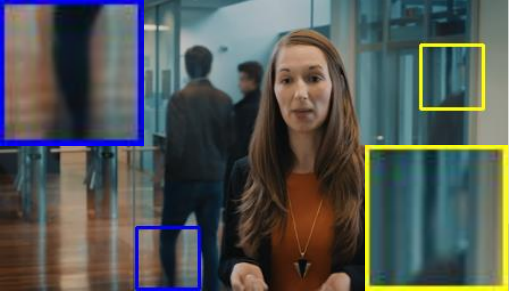}}
\subfigure[AdaCoF.]{\label{fig5:subfig:i}
\includegraphics[width=0.1575\linewidth]{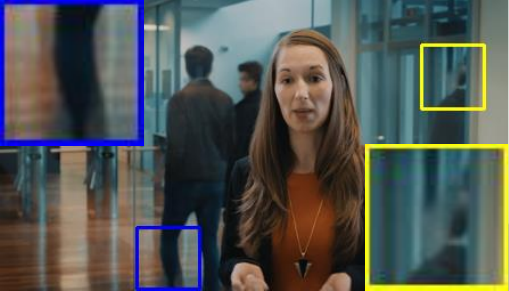}}
\subfigure[Ours-$\mathcal{L}_{C}$.]{\label{fig5:subfig:j}
\includegraphics[width=0.1575\linewidth]{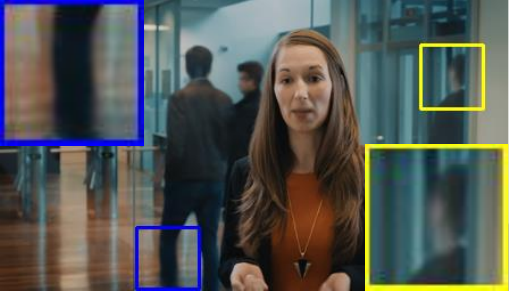}}
\subfigure[Ours-$\mathcal{L}_{F}$.]{\label{fig5:subfig:k}
\includegraphics[width=0.1575\linewidth]{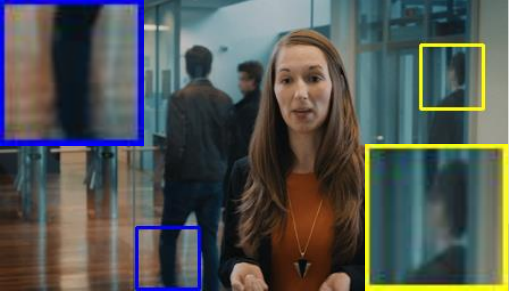}}
\subfigure[Ground Truth.]{\label{fig5:subfig:l}
\includegraphics[width=0.1575\linewidth]{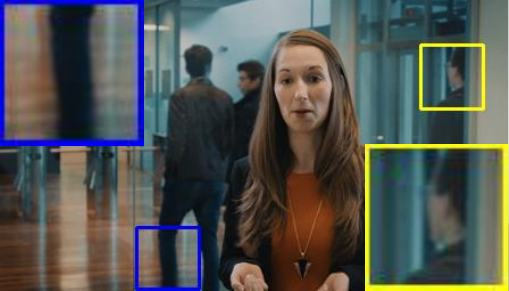}}
\vspace{-3mm}
\caption{Qualitative evaluation with respect to large motion (blue rectangle) and occlusion (yellow rectangle).
}
\label{fig5}
\end{figure*}

Figure \ref{fig5} shows an example of rotation motion around the knee joint (the blue rectangle) and occlusion (the yellow rectangle).

\begin{figure*}[!ht]
\centering
\vspace{-3mm}
\subfigure[Overlayed.]{\label{fig6:subfig:a}
\includegraphics[width=0.1575\linewidth]{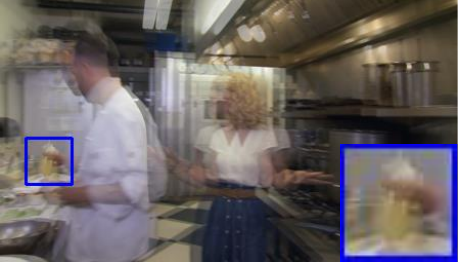}}
\subfigure[MEMC-Net$^*$.]{\label{fig6:subfig:b}
\includegraphics[width=0.1575\linewidth]{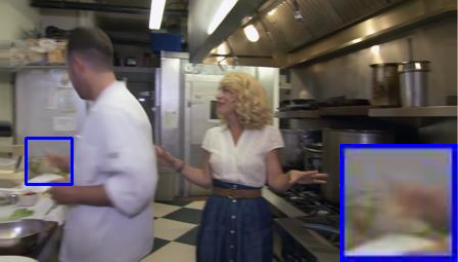}}
\subfigure[CyclicGen.]{\label{fig6:subfig:c}
\includegraphics[width=0.1575\linewidth]{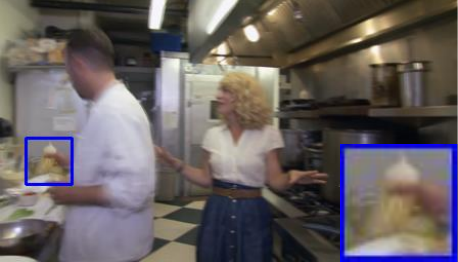}}
\subfigure[ToFlow.]{\label{fig6:subfig:d}
\includegraphics[width=0.1575\linewidth]{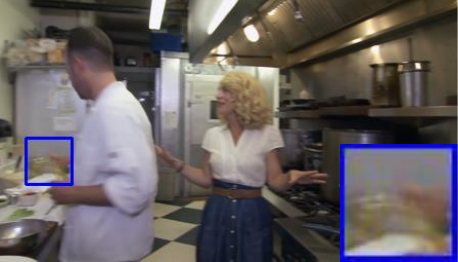}}
\subfigure[DAIN.]{\label{fig6:subfig:e}
\includegraphics[width=0.1575\linewidth]{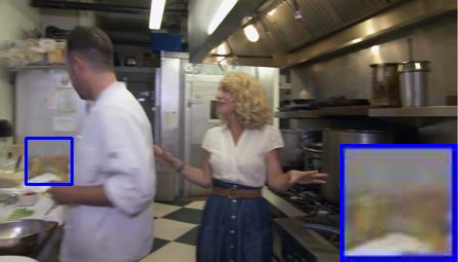}}
\subfigure[STAR-T$_{\mathrm{HR}}$.]{\label{fig6:subfig:f}
\includegraphics[width=0.1575\linewidth]{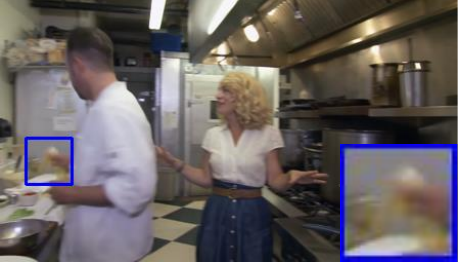}}

\subfigure[CAIN.]{\label{fig6:subfig:g}
\includegraphics[width=0.1575\linewidth]{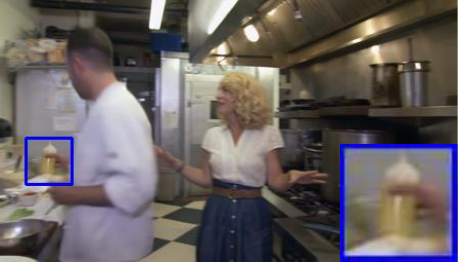}}
\subfigure[DSepConv.]{\label{fig6:subfig:h}
\includegraphics[width=0.1575\linewidth]{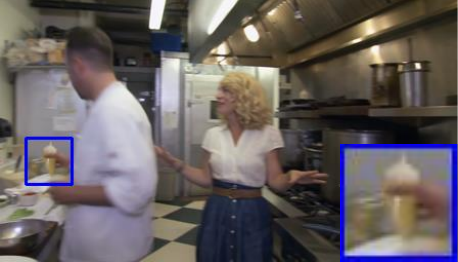}}
\subfigure[AdaCoF.]{\label{fig6:subfig:i}
\includegraphics[width=0.1575\linewidth]{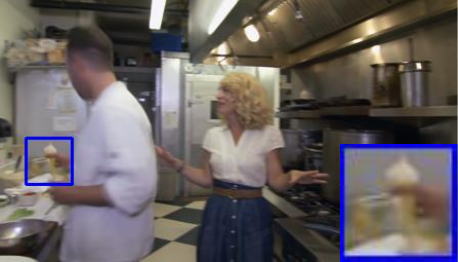}}
\subfigure[Ours-$\mathcal{L}_{C}$.]{\label{fig6:subfig:j}
\includegraphics[width=0.1575\linewidth]{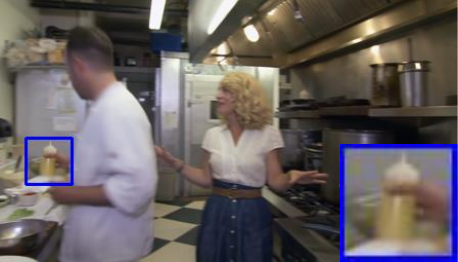}}
\subfigure[Ours-$\mathcal{L}_{F}$.]{\label{fig6:subfig:k}
\includegraphics[width=0.1575\linewidth]{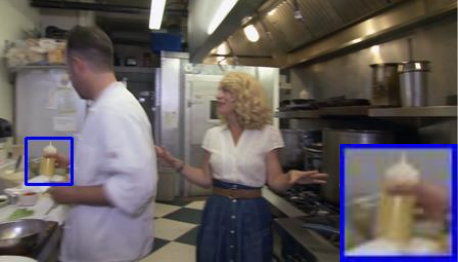}}
\subfigure[Ground Truth.]{\label{fig6:subfig:l}
\includegraphics[width=0.1575\linewidth]{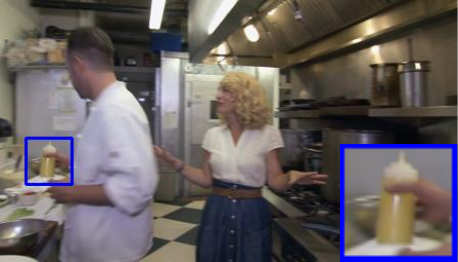}}
\vspace{-3mm}
\caption{Qualitative evaluation on a video with explicit camera motion.
}
\label{fig6}
\end{figure*}

\begin{figure*}[!ht]
\centering
\vspace{-3mm}
\subfigure[Overlayed.]{\label{fig7:subfig:a}
\includegraphics[width=0.1575\linewidth]{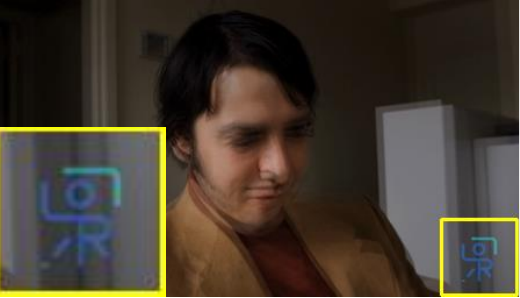}}
\subfigure[MEMC-Net$^*$.]{\label{fig7:subfig:b}
\includegraphics[width=0.1575\linewidth]{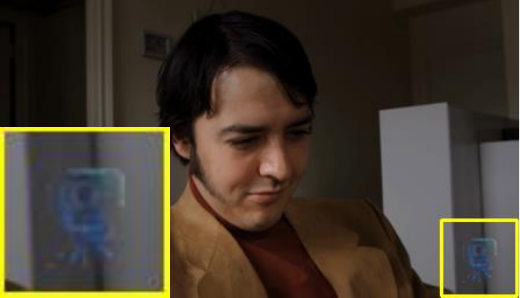}}
\subfigure[CyclicGen.]{\label{fig7:subfig:c}
\includegraphics[width=0.1575\linewidth]{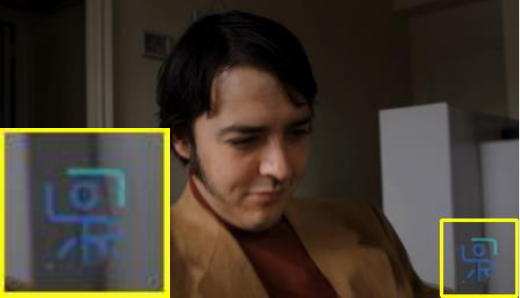}}
\subfigure[ToFlow.]{\label{fig7:subfig:d}
\includegraphics[width=0.1575\linewidth]{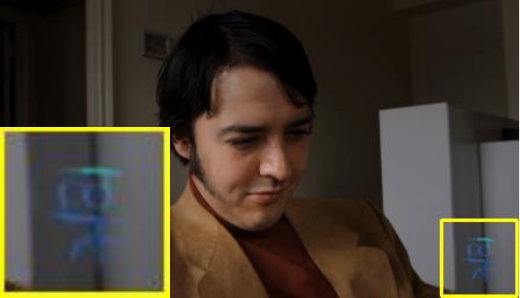}}
\subfigure[DAIN.]{\label{fig7:subfig:e}
\includegraphics[width=0.1575\linewidth]{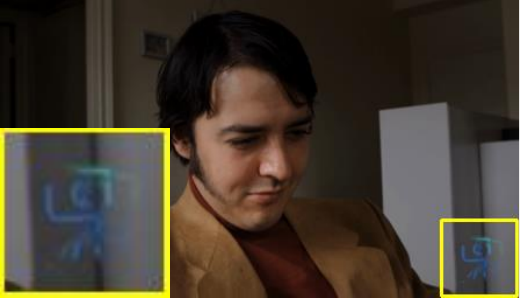}}
\subfigure[STAR-T$_{\mathrm{HR}}$.]{\label{fig7:subfig:f}
\includegraphics[width=0.1575\linewidth]{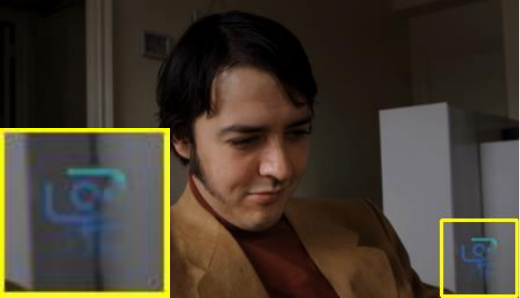}}

\subfigure[CAIN.]{\label{fig7:subfig:g}
\includegraphics[width=0.1575\linewidth]{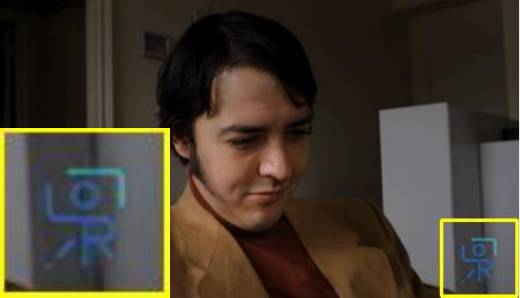}}
\subfigure[DSepConv.]{\label{fig7:subfig:h}
\includegraphics[width=0.1575\linewidth]{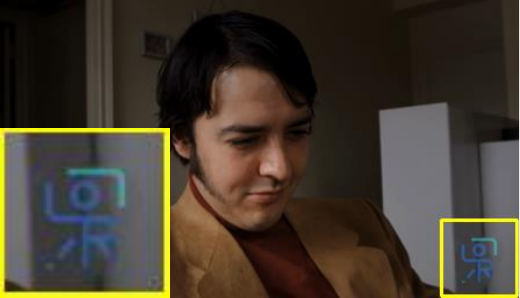}}
\subfigure[AdaCoF.]{\label{fig7:subfig:i}
\includegraphics[width=0.1575\linewidth]{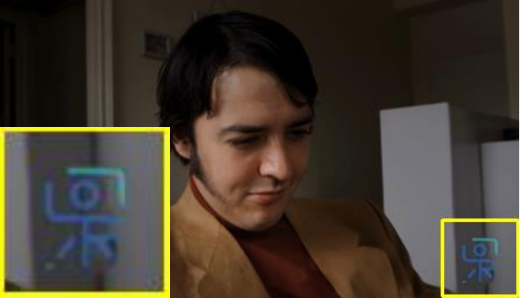}}
\subfigure[Ours-$\mathcal{L}_{C}$.]{\label{fig7:subfig:j}
\includegraphics[width=0.1575\linewidth]{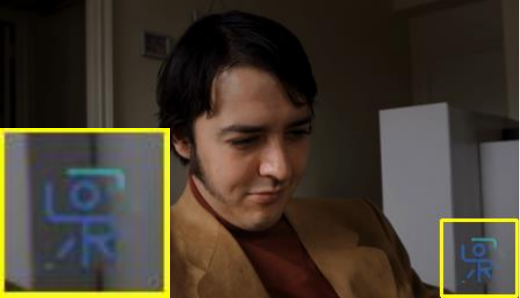}}
\subfigure[Ours-$\mathcal{L}_{F}$.]{\label{fig7:subfig:k}
\includegraphics[width=0.1575\linewidth]{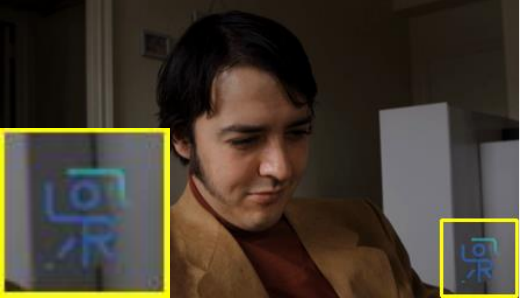}}
\subfigure[Ground Truth.]{\label{fig7:subfig:l}
\includegraphics[width=0.1575\linewidth]{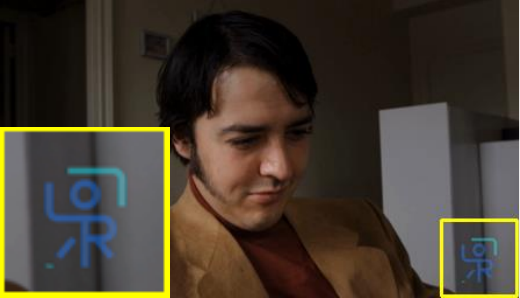}}
\vspace{-3mm}
\caption{Qualitative evaluation on a video with discontinuous motion.
}
\label{fig7}
\end{figure*}
The CyclicGen, ToFlow and DAIN produce broken results on the man's skin due to the usage of inaccurate optical flow and most of the methods lose information of the head area and appear blurry. In contrast, both our methods handle these situations better than the others.

The example in Figure \ref{fig6} is subject to explicit camera motion. We observe that the interpolation results from MEMC-Net$^*$, CyclicGen, ToFlow, DAIN and STAR-T$_{\mathrm{HR}}$ fail to reconstruct the bottle clearly because both the bottle areas from the two source frames are wrongly estimated as occlusion. On the contrary, our two results are sharp and free from blurriness, with the $\mathcal{L}_{F}$-trained model retaining more high-frequency details. Additionally, compared to the other kernel-based methods DSepConv and AdaCoF, the proposed method produces more complete result. We attribute this to the use of bias estimator, which learns residual information for better pixel reconstruction.

We further show an example where the motion is discontinuous in Figure \ref{fig7}. From the overlayed frame in \ref{fig7:subfig:a} we can observe that the motion is continuous except the sign highlighted with yellow rectangle. This discontinuity makes it hard to estimate optical flow accurately, causing ghosting artifacts for those methods strictly relying on optical flow (MEMC-Net$^*$, ToFlow, DAIN, STAR-T$_{\mathrm{HR}}$). In this example, the other methods, including ours, perform well.

\begin{table*}[ht]
\centering
\caption{Quantitative comparisons against kernel-based methods on SNU-FILM \cite{cain} dataset (abbreviated by S.F.). The numbers in \textbf{bold} represent the best performance.}
\resizebox{\linewidth}{!}{
\begin{tabular}{lccccc}
\toprule
\multirow{2}*{Methods}
\multirow{3}{*}{ } &
\multirow{2}*{\shortstack{Kernel\\(size)} }
\multirow{3}{*}{ } &
\multicolumn{1}{c}{ S.F.-\texttt{Easy} } &
\multicolumn{1}{c}{ S.F.-\texttt{Medium} } &
\multicolumn{1}{c}{ S.F.-\texttt{Hard} } &
\multicolumn{1}{c}{ S.F.-\texttt{Extreme} } \\

\cmidrule(lr){3-3}
\cmidrule(lr){4-4}
\cmidrule(lr){5-5}
\cmidrule(lr){6-6}

& & PSNR$\uparrow$/SSIM$\uparrow$/LPIPS$\downarrow$& PSNR$\uparrow$/SSIM$\uparrow$/LPIPS$\downarrow$ & PSNR$\uparrow$/SSIM$\uparrow$/LPIPS$\downarrow$ & PSNR$\uparrow$/SSIM$\uparrow$/LPIPS$\downarrow$\\
\midrule
SepConv-$\mathcal{L}_{1}$ \cite{sepconv} & learned(51) & 39.47 / \textbf{0.990} / 0.017 & 34.98 / 0.976 / 0.032 & 29.35 / 0.925 / 0.075 & 24.31 / \textbf{0.845} / 0.154\\
SepConv-$\mathcal{L}_{F}$ \cite{sepconv} & learned(51) & 39.33 / 0.989 / \textbf{0.012} & 34.79 / 0.975 / \textbf{0.024} & 29.10 / 0.921 / 0.057 & 24.10 / 0.837 / 0.124\\
DSepConv \cite{dsepconv} & learned(5) & 39.94 / \textbf{0.990 }/ 0.019 & 35.30 / 0.977 / 0.035 & 29.56 / 0.925 / 0.074 & 24.34 / 0.840 / 0.149\\
AdaCoF \cite{adacof} & learned(5) & 39.43 / \textbf{0.990} / 0.020 & 34.90 / 0.975 / 0.037 & 29.41 / 0.924 / 0.076 & 24.29 / 0.844 / 0.149\\
AdaCoF+ \cite{adacof} & learned(11) & 39.53/ \textbf{0.990} / 0.020 & 34.99 / 0.976 / 0.036 & 29.50 / 0.925 / 0.074 & \textbf{24.45} / \textbf{0.845} / 0.146\\
\midrule
Ours-$\mathcal{L}_{C}$ & learned(5) & \textbf{40.01} / \textbf{0.990} / 0.019 & \textbf{35.37} / \textbf{0.978} / 0.034 & \textbf{29.59} / \textbf{0.926} / 0.074 & 24.39 /0.843 / 0.145\\
Ours-$\mathcal{L}_{F}$ & learned(5) & 39.50 / \textbf{0.990} / 0.013 & 35.02 / 0.976 / \textbf{0.024} & 29.33 / 0.921 / \textbf{0.055} & 24.12 / 0.834 / \textbf{0.121}\\
\bottomrule
\end{tabular}}
\label{tab:6}
\end{table*}

\begin{figure}[t]
\centering
\includegraphics[width=\linewidth]{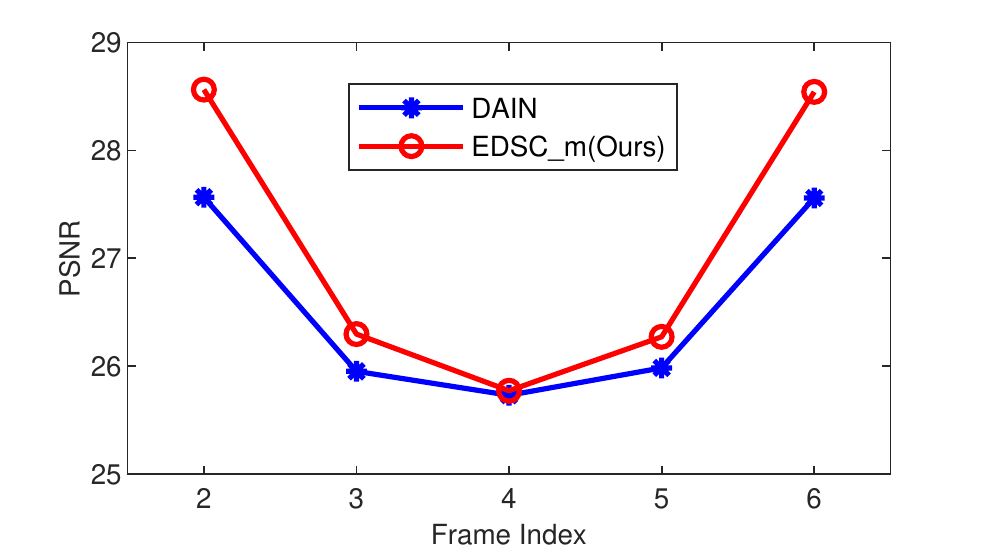}
\caption{PSNR at each time step when generating $\times6$ slow motion frames on the Vimeo90K-Septuplet test set \cite{Toflow}.
}
\label{fig14}
\end{figure}

\begin{figure}[t]
\centering
\subfigure{
\begin{minipage}[t]{0.3\linewidth}
\centerline{\includegraphics[width=1.05\linewidth]{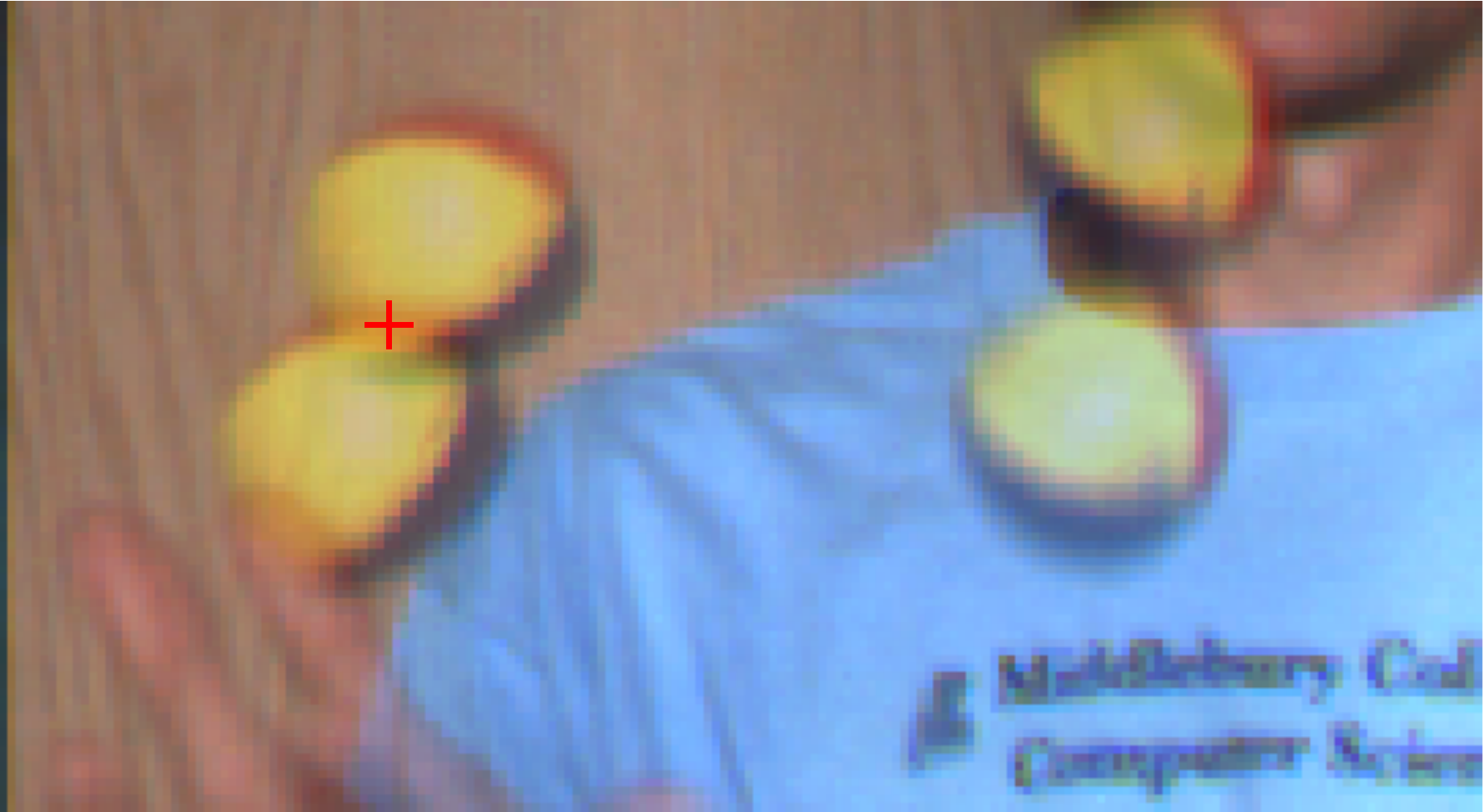}}
\par
\vspace{0.5mm}
\centerline{
\includegraphics[width=0.53in]{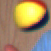}
\includegraphics[width=0.53in]{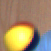}
}
\centerline{(a) Inputs}
\end{minipage}
}
\subfigure{
\begin{minipage}[t]{0.3\linewidth}
\centerline{\includegraphics[width=1.05\linewidth]{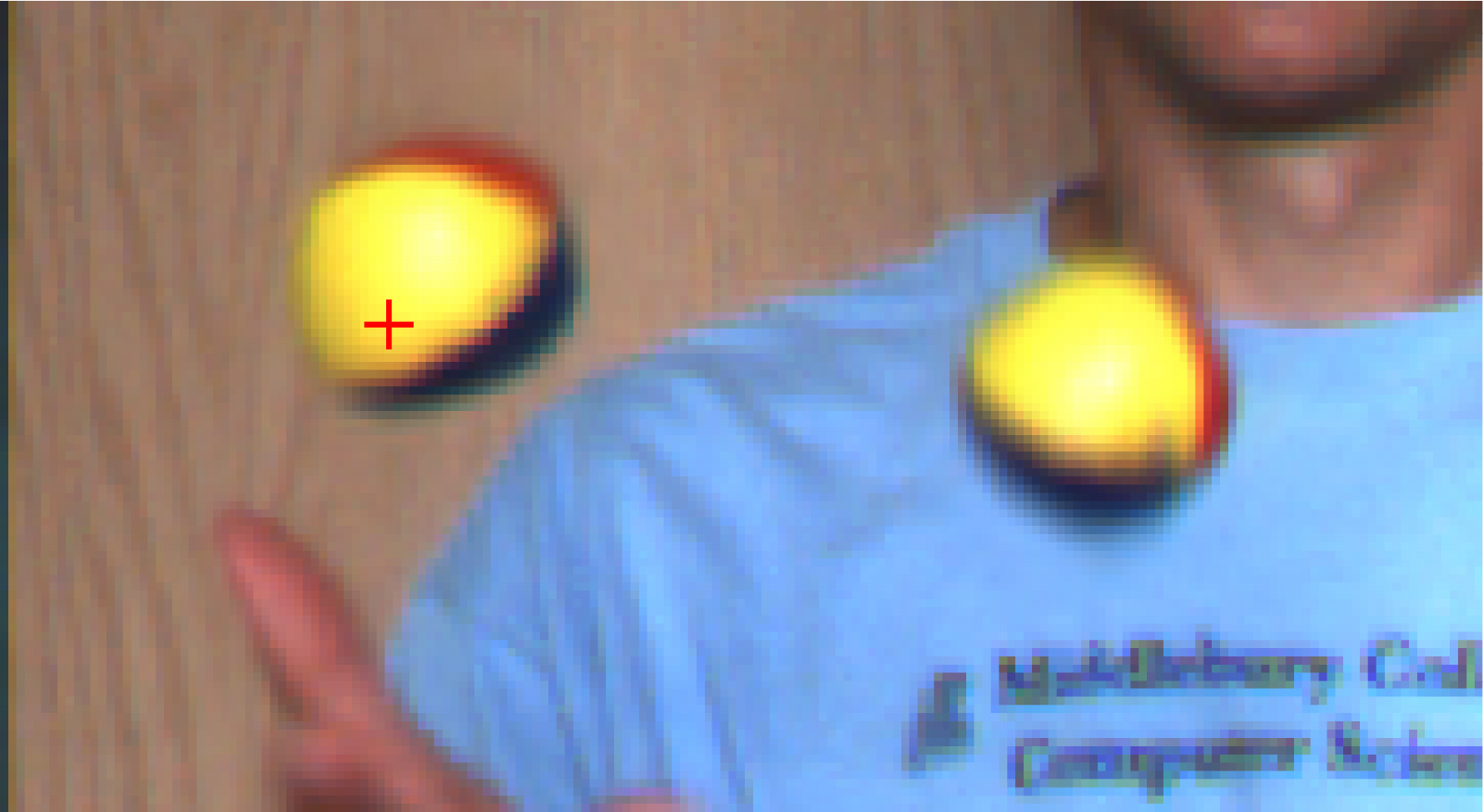}}
\par
\vspace{0.5mm}
\centerline{
\includegraphics[width=0.53in]{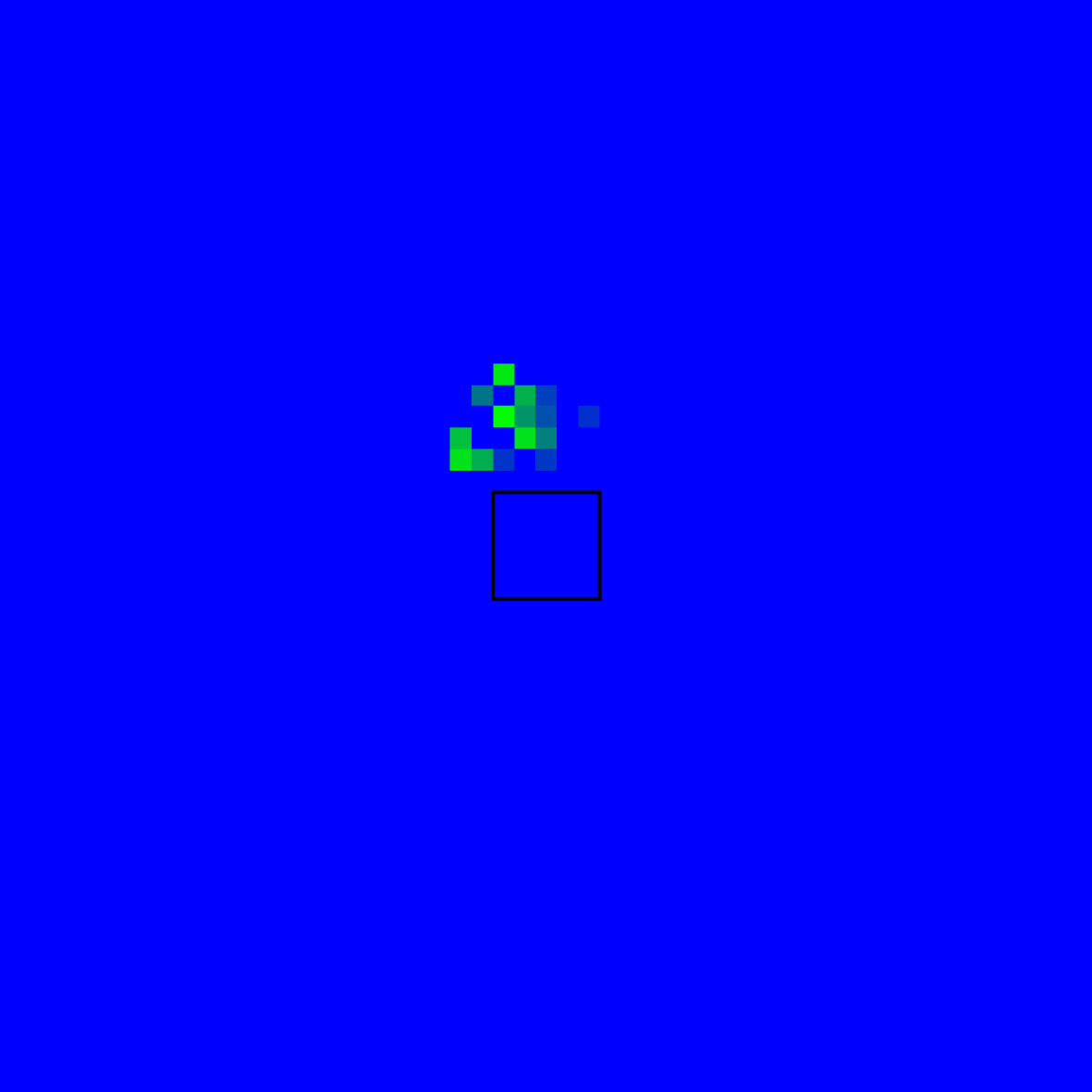}
\includegraphics[width=0.53in]{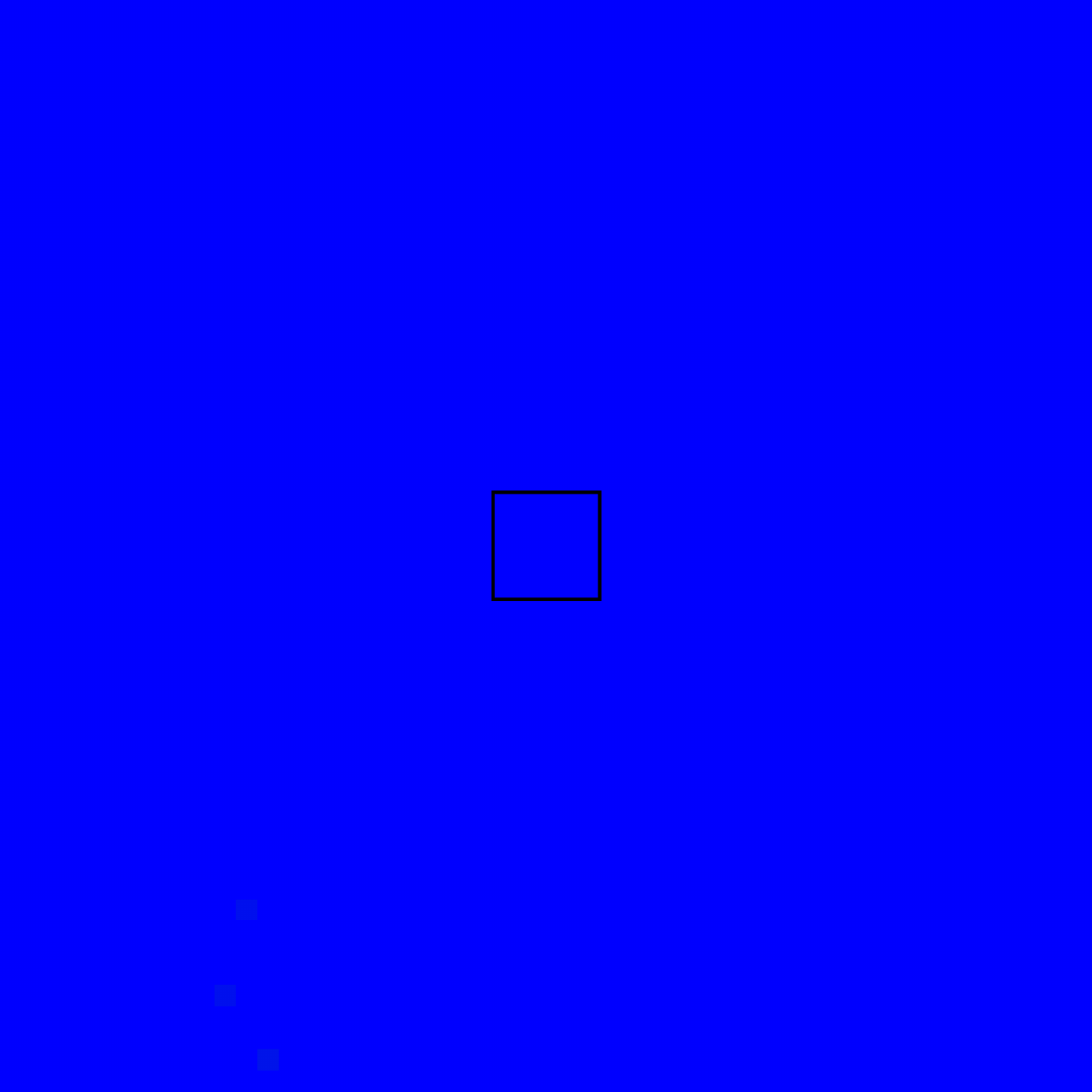}
}
\centerline{(b) t=0.1}
\end{minipage}
}
\subfigure{
\begin{minipage}[t]{0.3\linewidth}
\centerline{\includegraphics[width=1.05\linewidth]{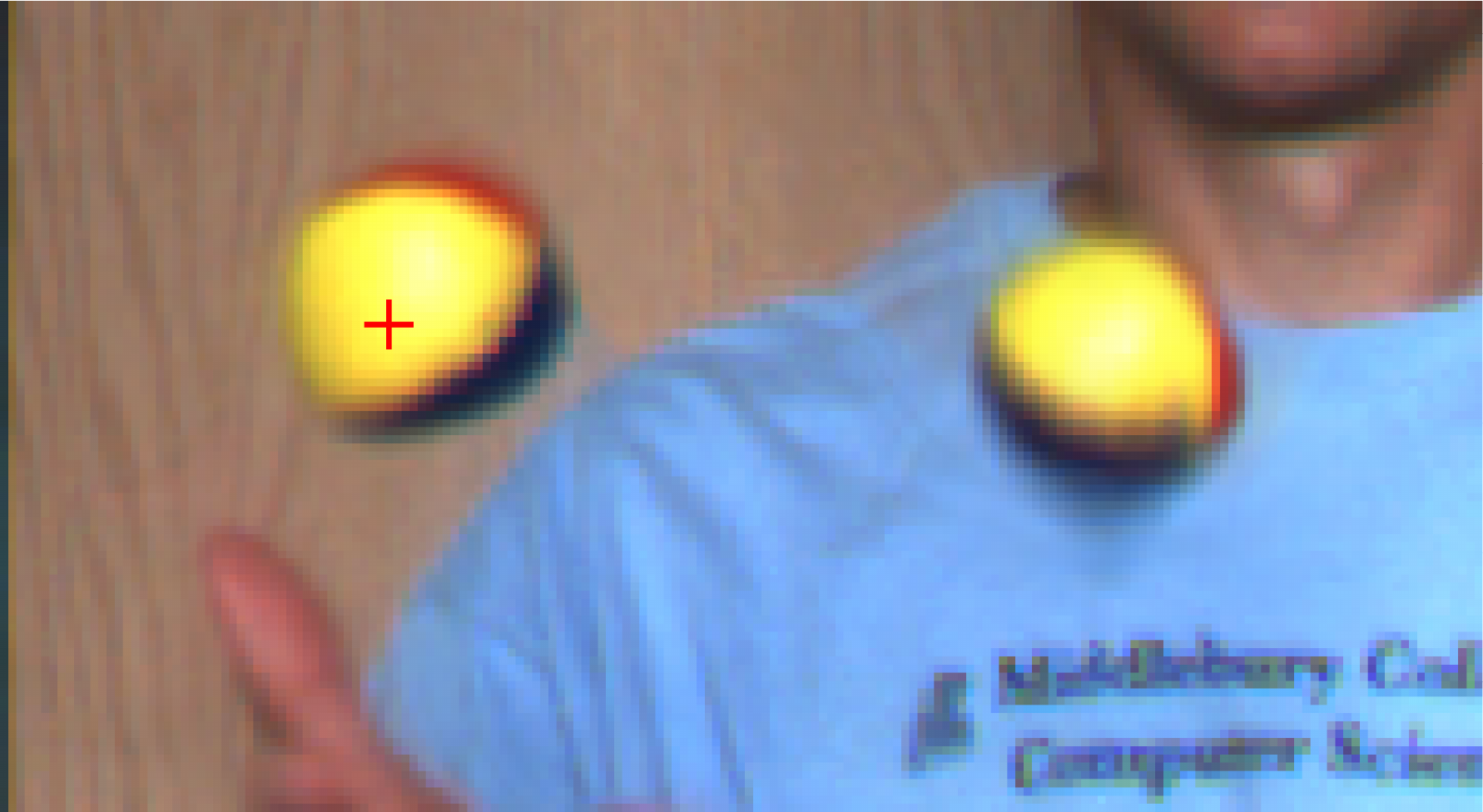}}
\par
\vspace{0.5mm}
\centerline{
\includegraphics[width=0.53in]{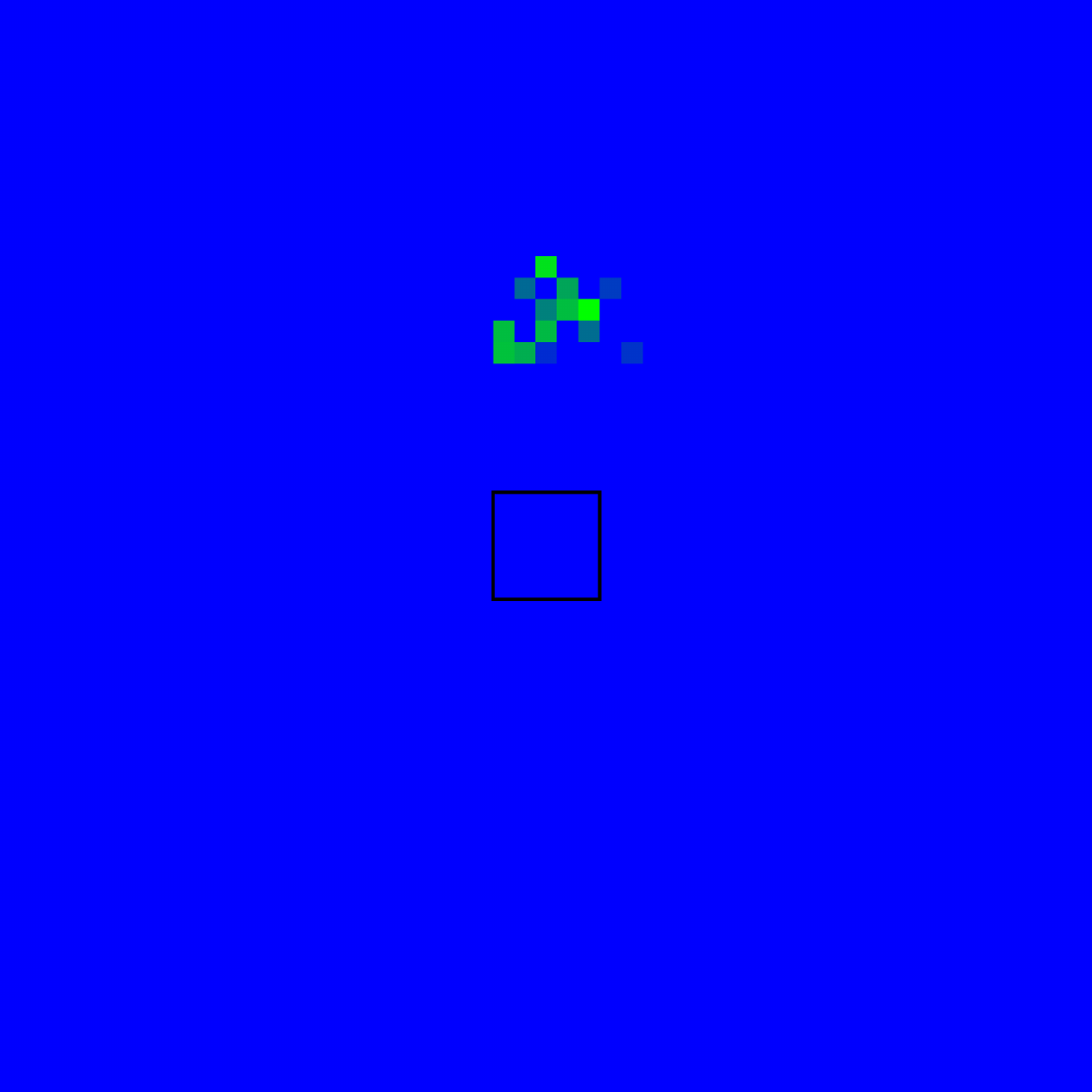}
\includegraphics[width=0.53in]{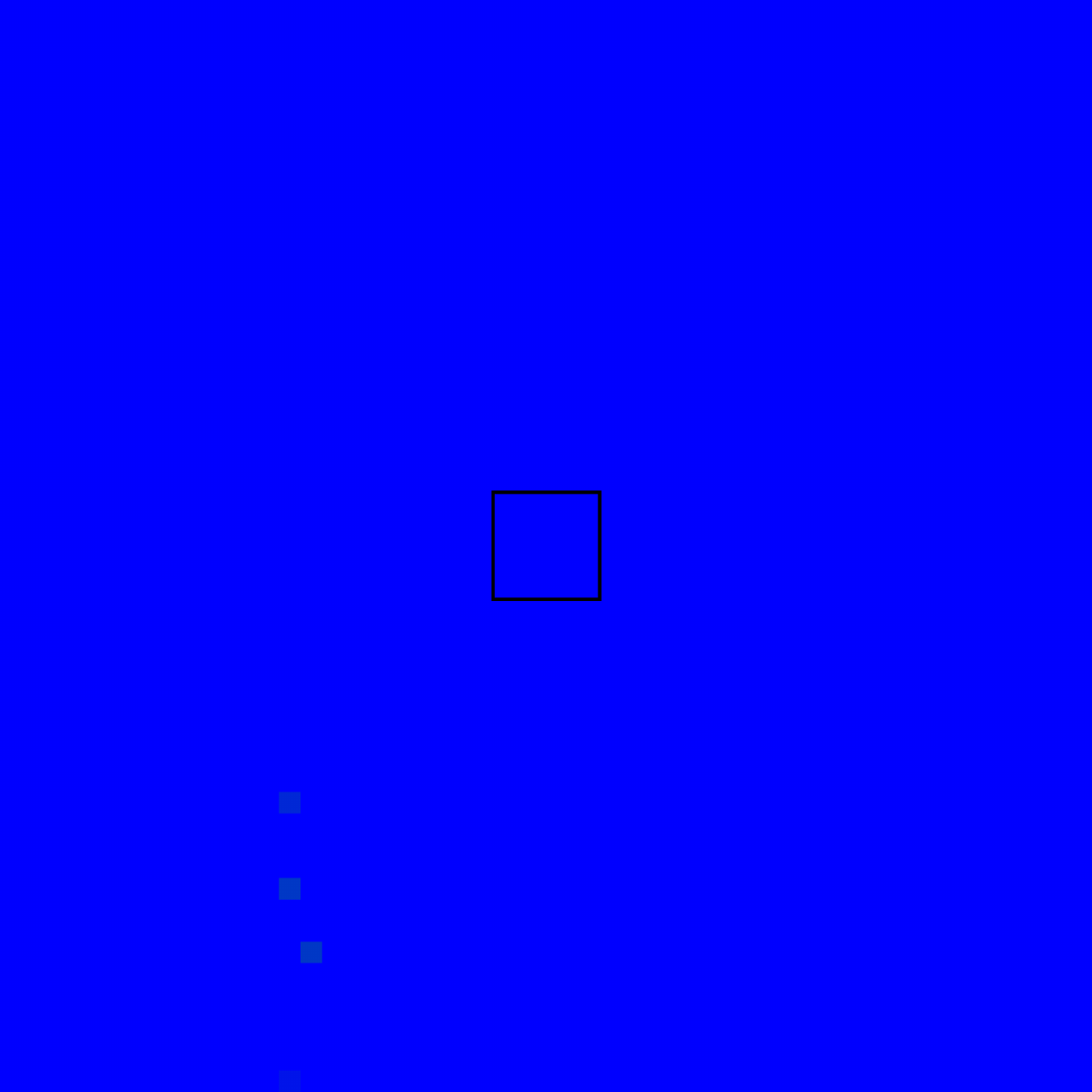}
}
\centerline{(c) t=0.3}
\end{minipage}
}

\subfigure{
\begin{minipage}[t]{0.3\linewidth}
\centerline{\includegraphics[width=1.05\linewidth]{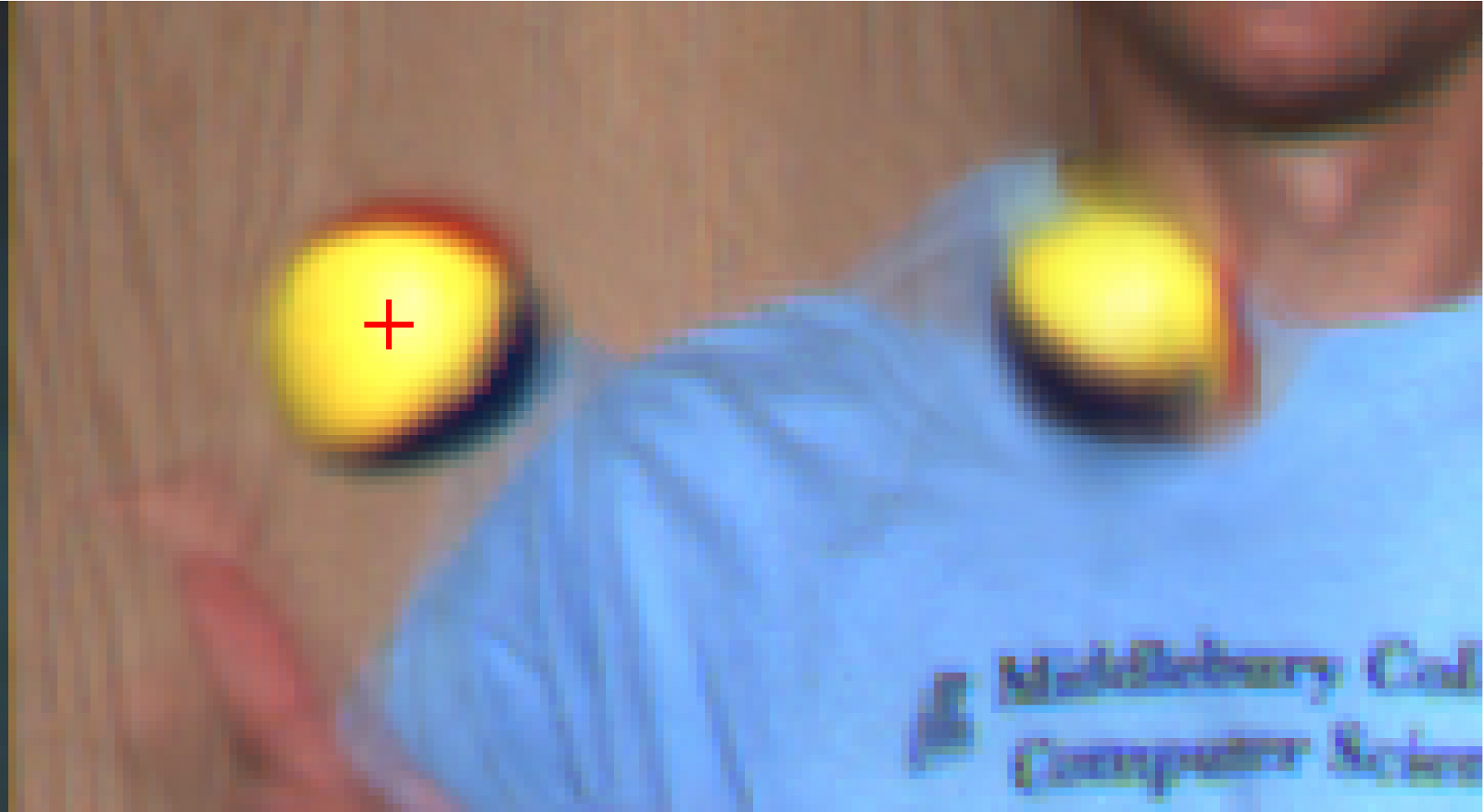}}
\par
\vspace{0.5mm}
\centerline{
\includegraphics[width=0.53in]{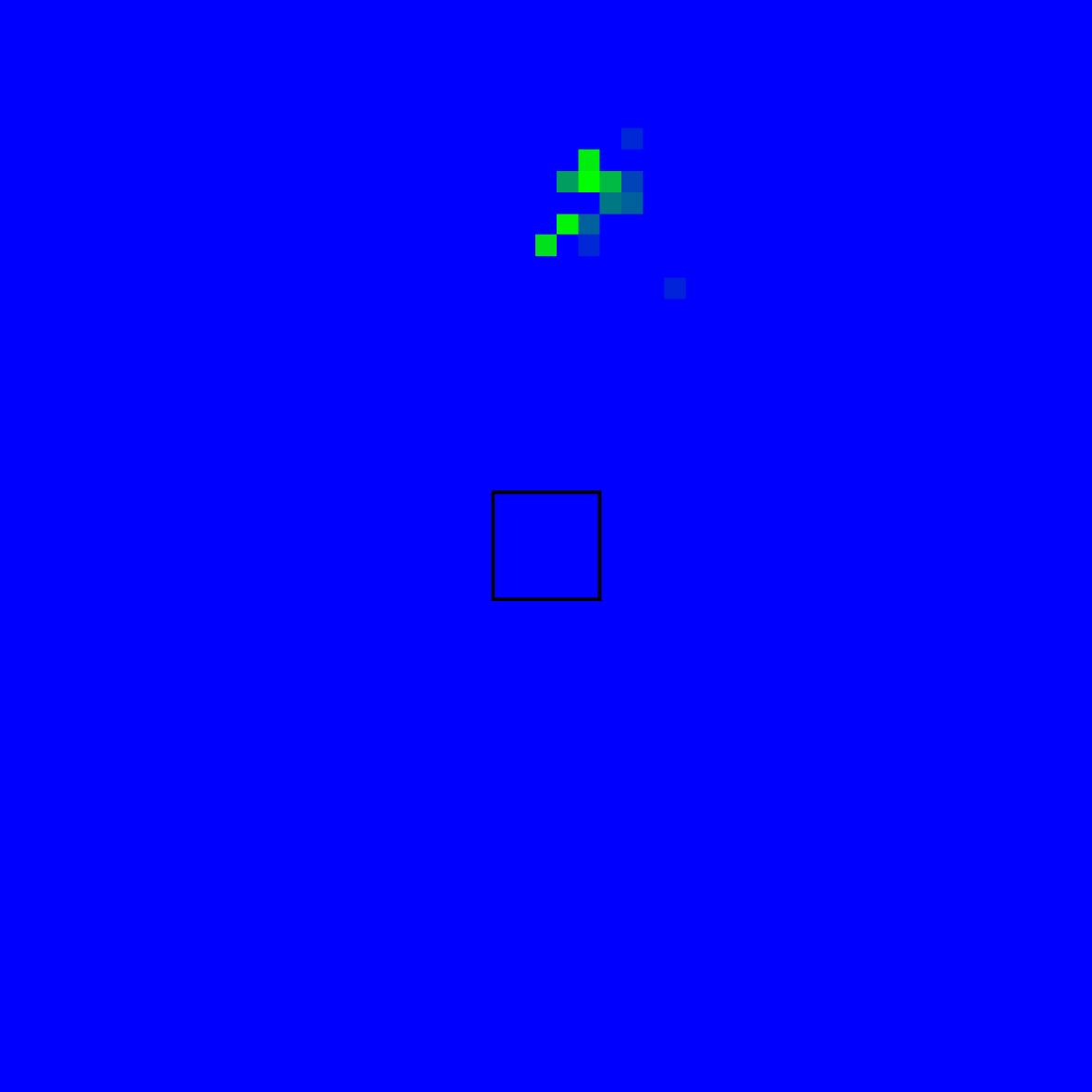}
\includegraphics[width=0.53in]{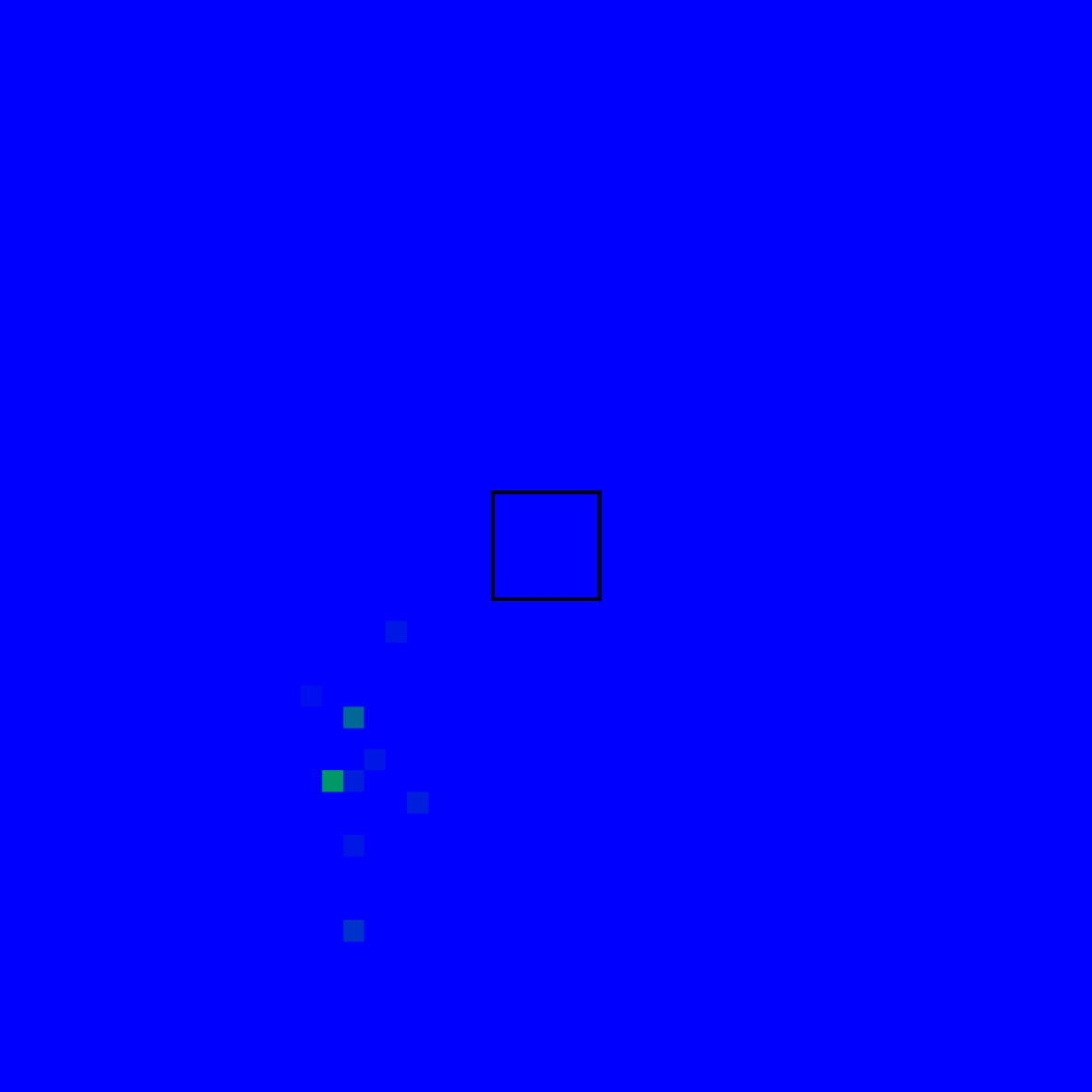}
}
\centerline{(d) t=0.5}
\end{minipage}
}
\subfigure{
\begin{minipage}[t]{0.3\linewidth}
\centerline{\includegraphics[width=1.05\linewidth]{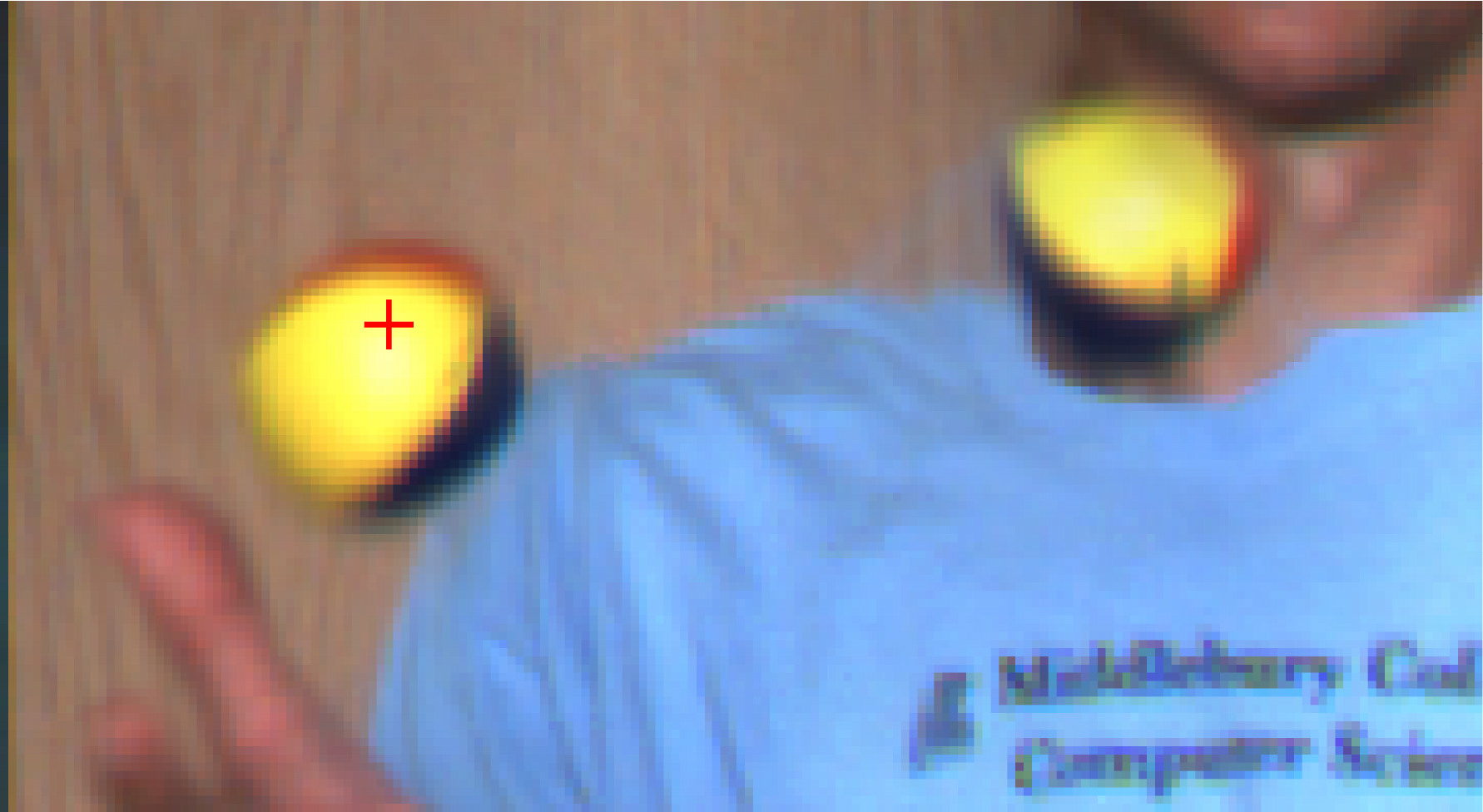}}
\par
\vspace{0.5mm}
\centerline{
\includegraphics[width=0.53in]{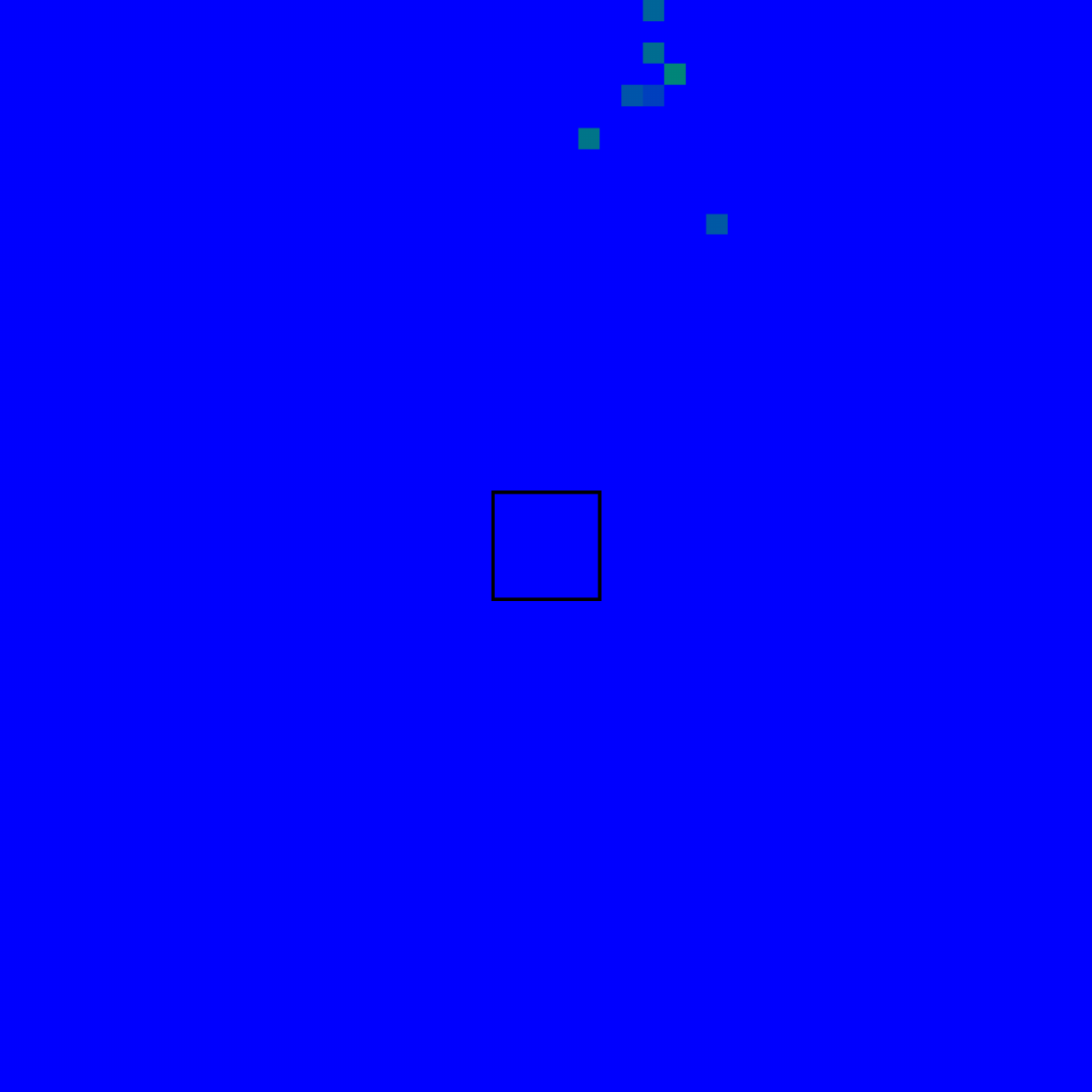}
\includegraphics[width=0.53in]{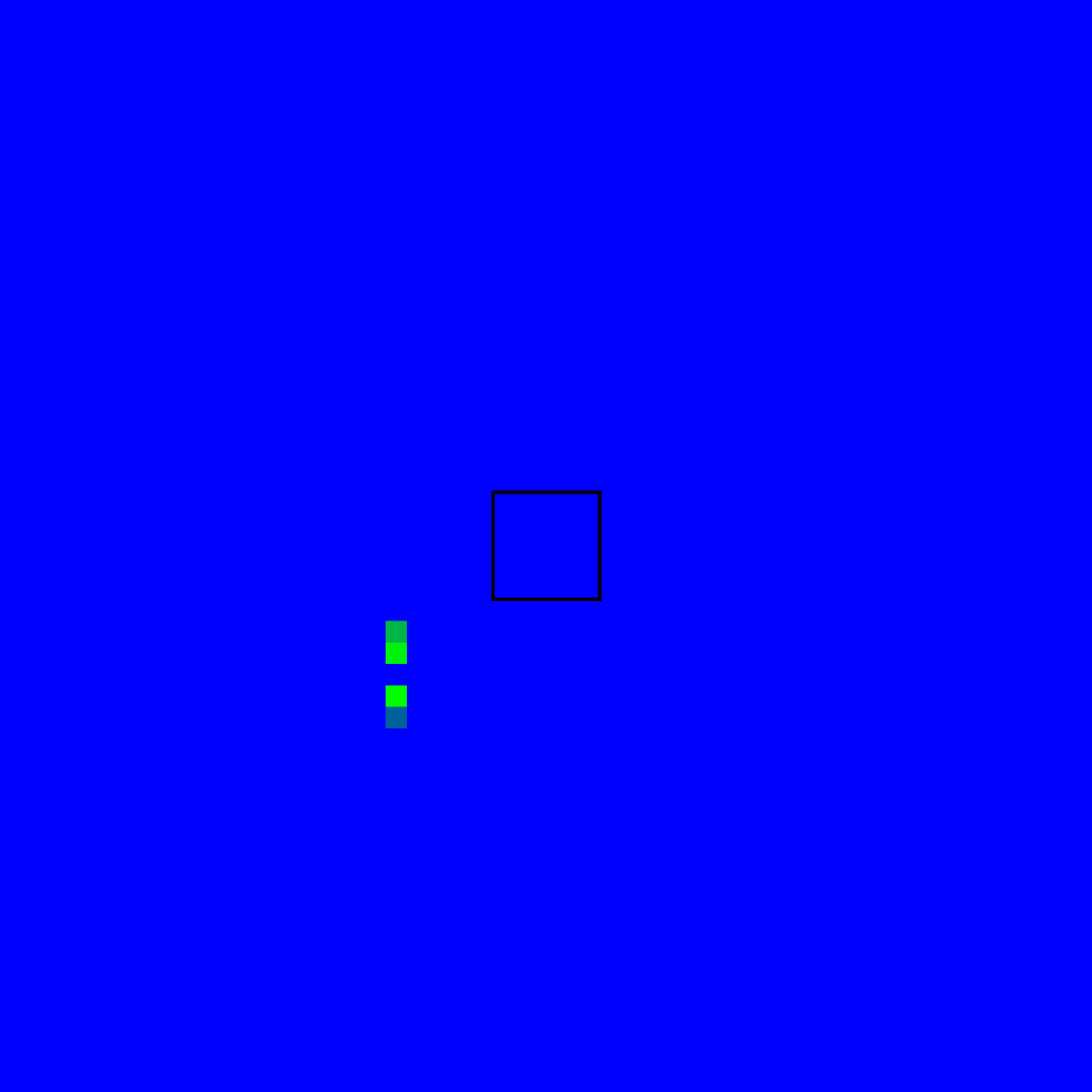}
}
\centerline{(e) t=0.7}
\end{minipage}
}
\subfigure{
\begin{minipage}[t]{0.3\linewidth}
\centerline{\includegraphics[width=1.05\linewidth]{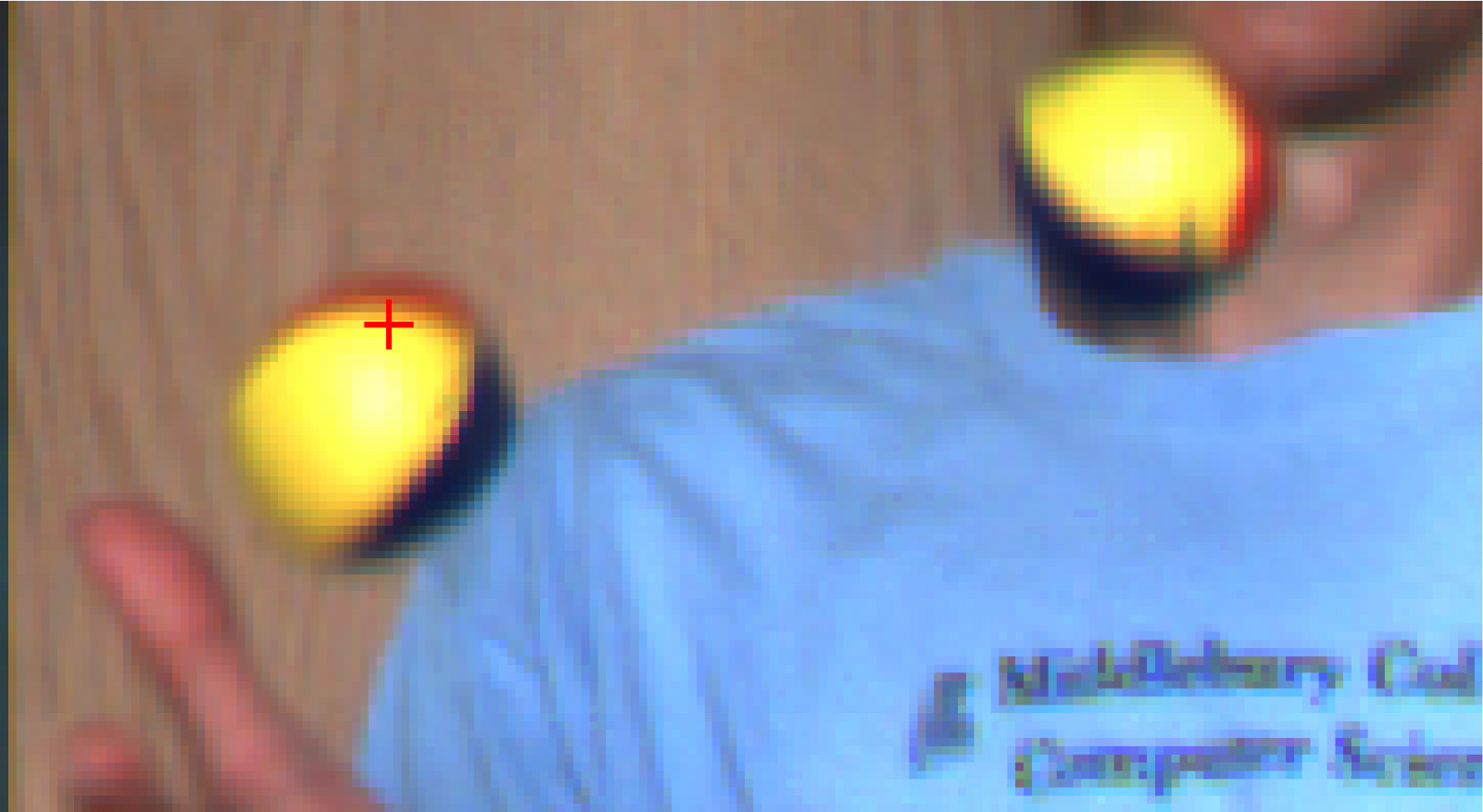}}
\par
\vspace{0.5mm}
\centerline{
\includegraphics[width=0.53in]{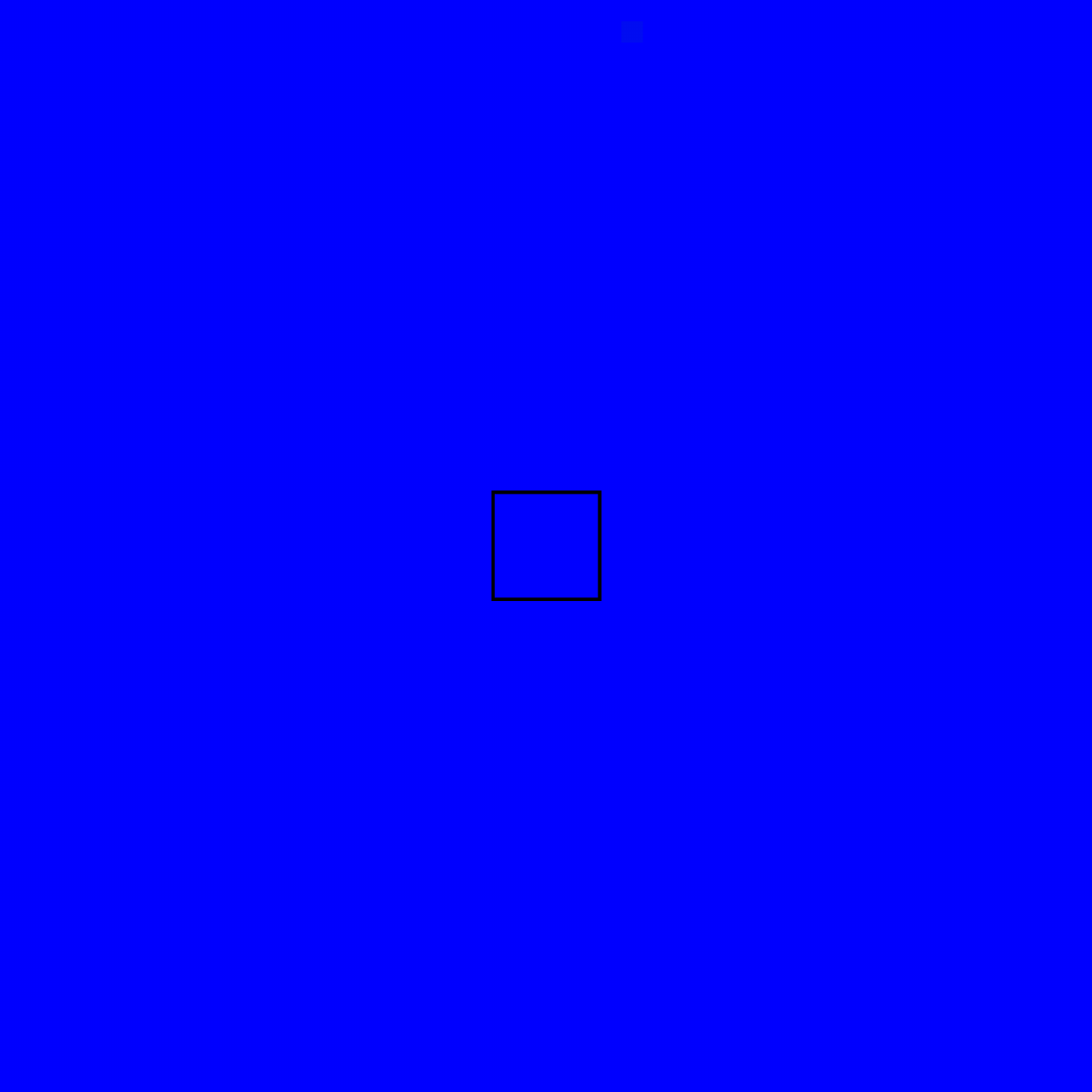}
\includegraphics[width=0.53in]{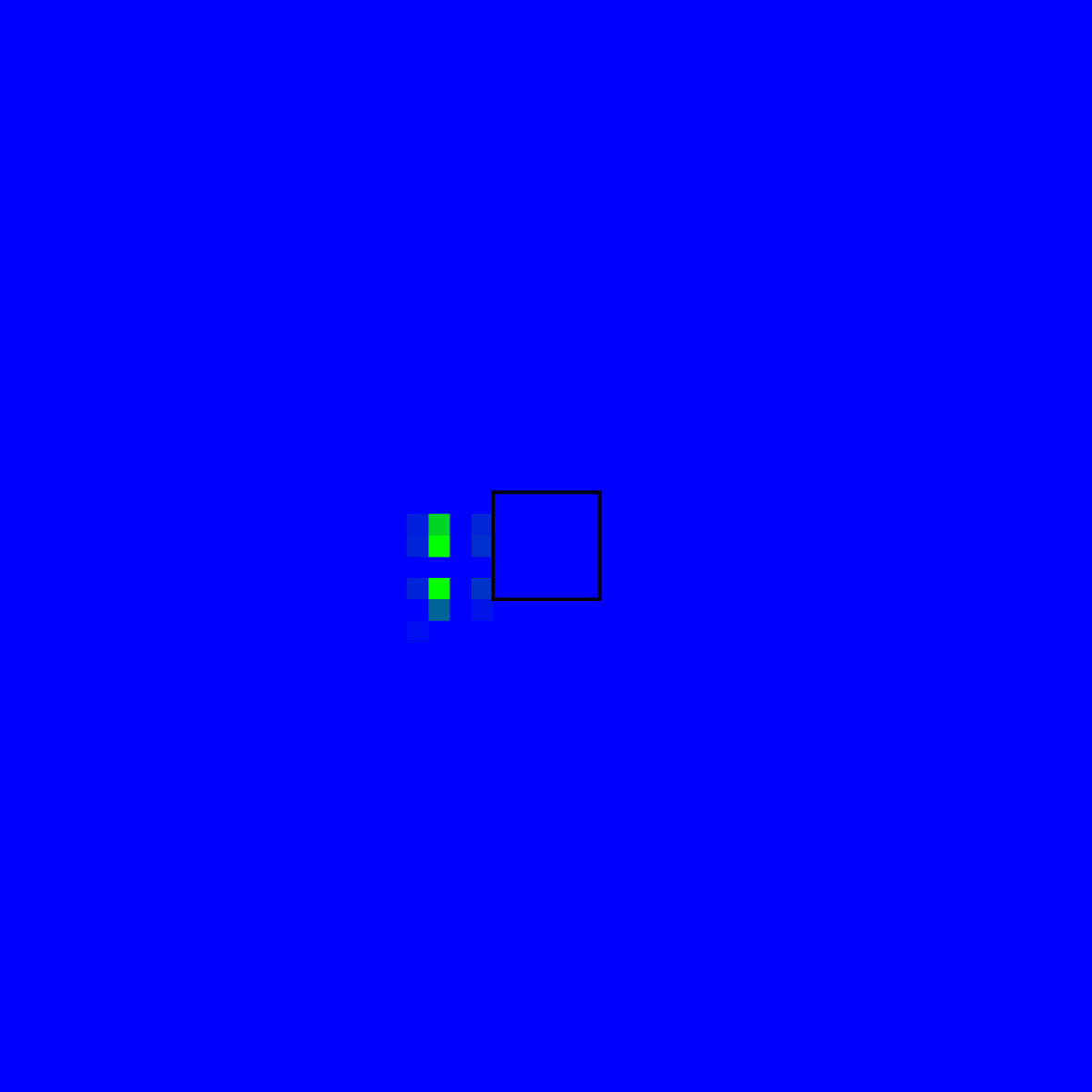}
}
\centerline{(f) t=0.9}
\end{minipage}
}

\centering
\caption{Arbitrary time interpolation generated from our proposed EDSC\_m. The overlayed and synthesized frames are shown in the first and third rows. We also show the reference patches in (a) and effective sampling locations of a synthesized pixel which centers at the red \textbf{+} in the other sub figures. The small black rectangle represents the location of local kernel and greener regions indicate higher absolute values. Please zoom in the figures for a better view.}
\label{fig13}
\end{figure}

\subsubsection{Arbitrary-position frame interpolation}
\label{arbitrary}
We perform a quantitative evaluation on the Vimeo90K-Septuplet test set \cite{Toflow}. Specifically, we interpolate frame 2 through 6 from frame 1 and frame 7 on all its 7,824 sequences to generate $\times6$ slow motion frames. We also compare our method to DAIN \cite{dain}, which can interpolate arbitrary in-between frames.
The PSNR scores at each frame index are shown in Figure \ref{fig14}, it can be clearly seen that our method outperforms DAIN for each individual in-between time step. In spite of the usage of adaptive convolutional kernels, the sampling locations for each synthesized pixel of DAIN heavily depend on optical flow, thus little inaccuracy may result in less plausible results. On the contrary, we learn which pixels to reference without a strict guidance (optical flow).

We found our solution efficient in terms of multi-frame interpolation. Although adding additional temporal channels as input seems to be redundant, it needs only little increment of computational cost: 0.072G (0.52\%) in terms of FLOPs and 0.006M (0.07\%) in terms of parameters.
In addition, it just increases 0.001 seconds of execution time to interpolate a 1280$\times$720 frame using an Nvidia Titan X GPU.

In Figure \ref{fig13}, we show a set of interpolation results at $t=0.1,0.3,0.5,0.7$ and 0.9. We also visualize the effective
sampling locations of a pixel (indicated by the red \textbf{+}), which locates at the same position of the synthesized frames. First, despite some time steps (\emph{e.g.}, $t=0.1,0.3,0.7,0.9$) are not involved during the training process, our method can generate plausible results. Second, our method is aware of the inequality of information between the two input frames when producing intermediate frames with different temporal positions. To be detailed, when $t<0.5$, our model mainly takes information from the first frame, whereas for $t>0.5$, the non-zero elements are mainly in the second frame. This is in line with the assumption that the former frame is more reliable in synthesis for earlier time steps and so is the latter for later time steps. Third, our method is aware of the motion between the two input frames. For instance, the non-zero elements are spatially farther away from the center in the first kernels when $t$ getting bigger, while those move in opposite direction in the second kernels. This phenomenon shows that the learned offsets vary from different $t$, indicating the effectiveness of the usage of analogous coord-conv trick to deal with temporal consistency.

\subsection{Model Analysis}
\subsubsection{Effect of dealing with different motion degrees}
We investigate the ability of different algorithms to handle different motion degrees. Typically, methods with a component of optical flow estimation can capture large motion as long as it is accurately computed. That is why it is so popular to make use of the off-the-shelf optical flow estimators and further perform fine-tuning. However, there is no optical flow utilized in kernel based methods. The capacity to deal with large motion hinges on kernel estimates. Therefore, for a fair comparison we evaluate the performance with respect to the amount of motion among kernel based methods on a more comprehensive dataset SNU-FILM \cite{cain}. As shown in Table \ref{tab:6}, our $\mathcal{L}_{C}$-trained model achieves the best performance on the \texttt{Easy}, \texttt{Medium} and \texttt{Hard} sets in terms of PSNR and SSIM, while is marginally worse than AdaCoF+ which learns kernels with a larger size $11\times11$.
\begin{figure}[!th]
\centering
\subfigure{
\begin{minipage}[t]{0.21\linewidth}
\centerline{\includegraphics[width=1.05\linewidth]{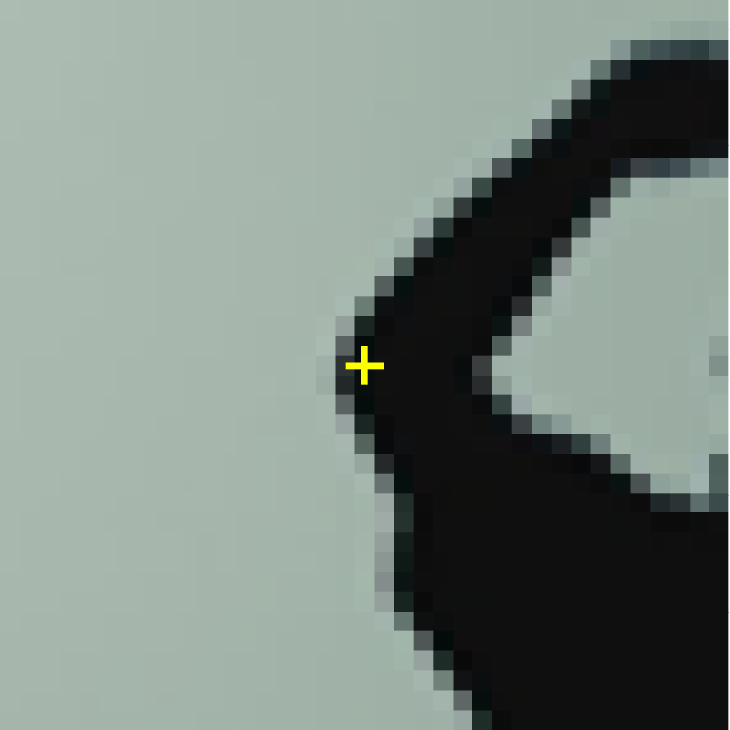}}
\centerline{\small{Patch 1}}
\vspace{1mm}
\centerline{\includegraphics[width=1.05\linewidth]{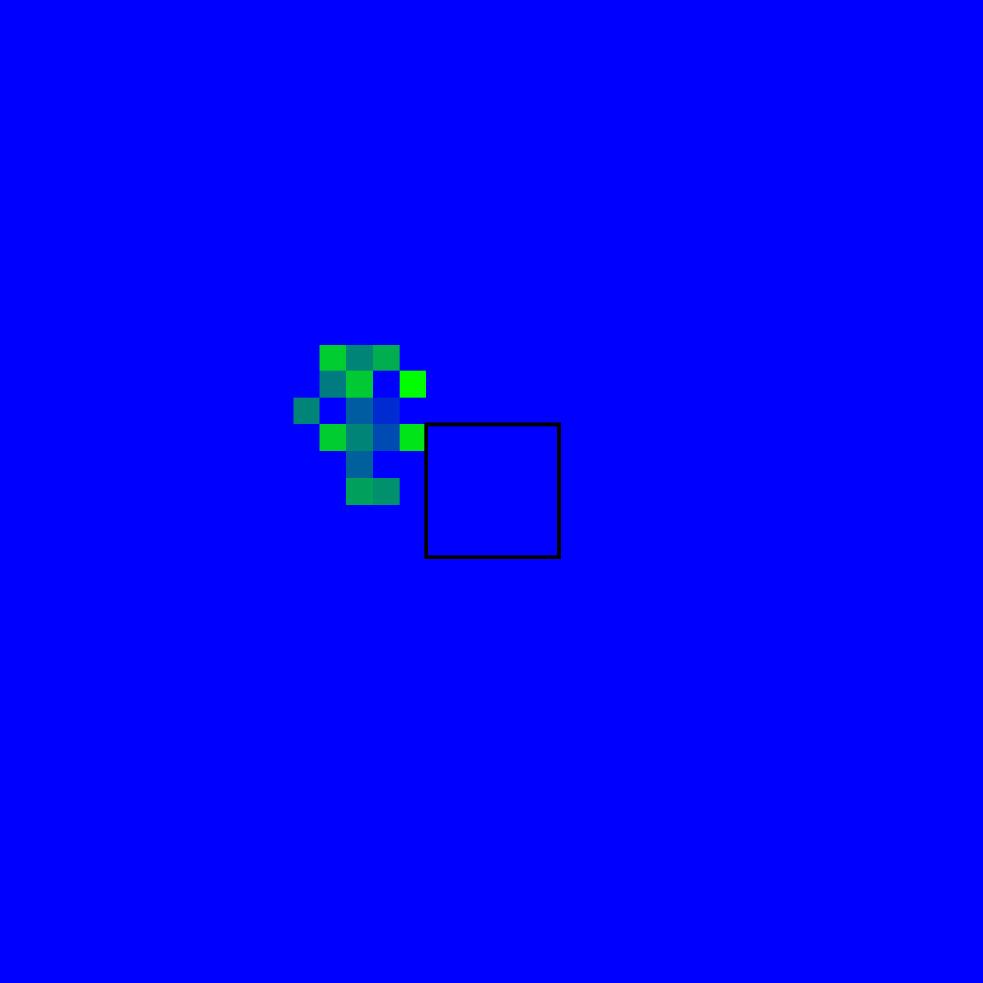}}
\centerline{\small{Kernel 1}}
\end{minipage}
}
\subfigure{
\begin{minipage}[t]{0.21\linewidth}
\centerline{\includegraphics[width=1.05\linewidth]{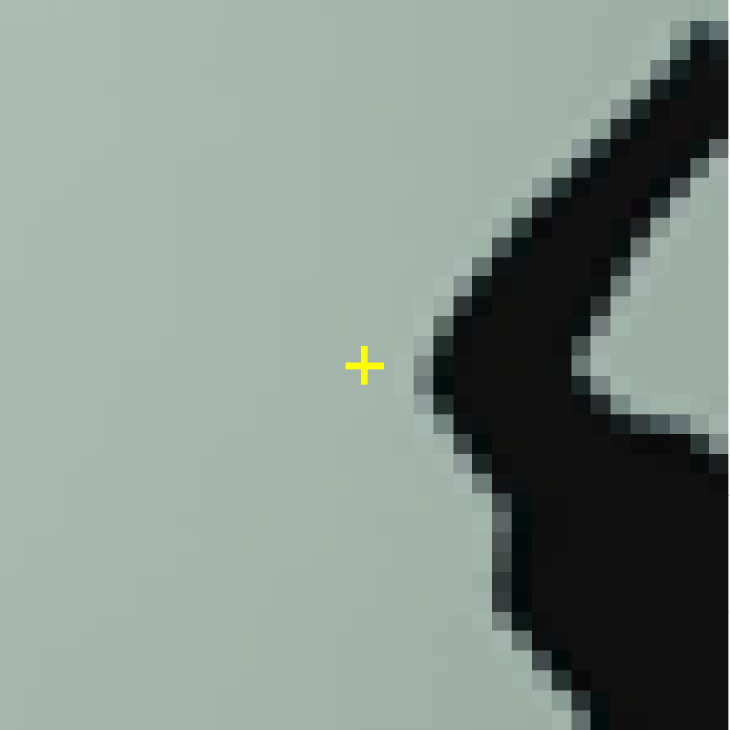}}
\centerline{\small{Patch 2}}
\vspace{1mm}
\centerline{\includegraphics[width=1.05\linewidth]{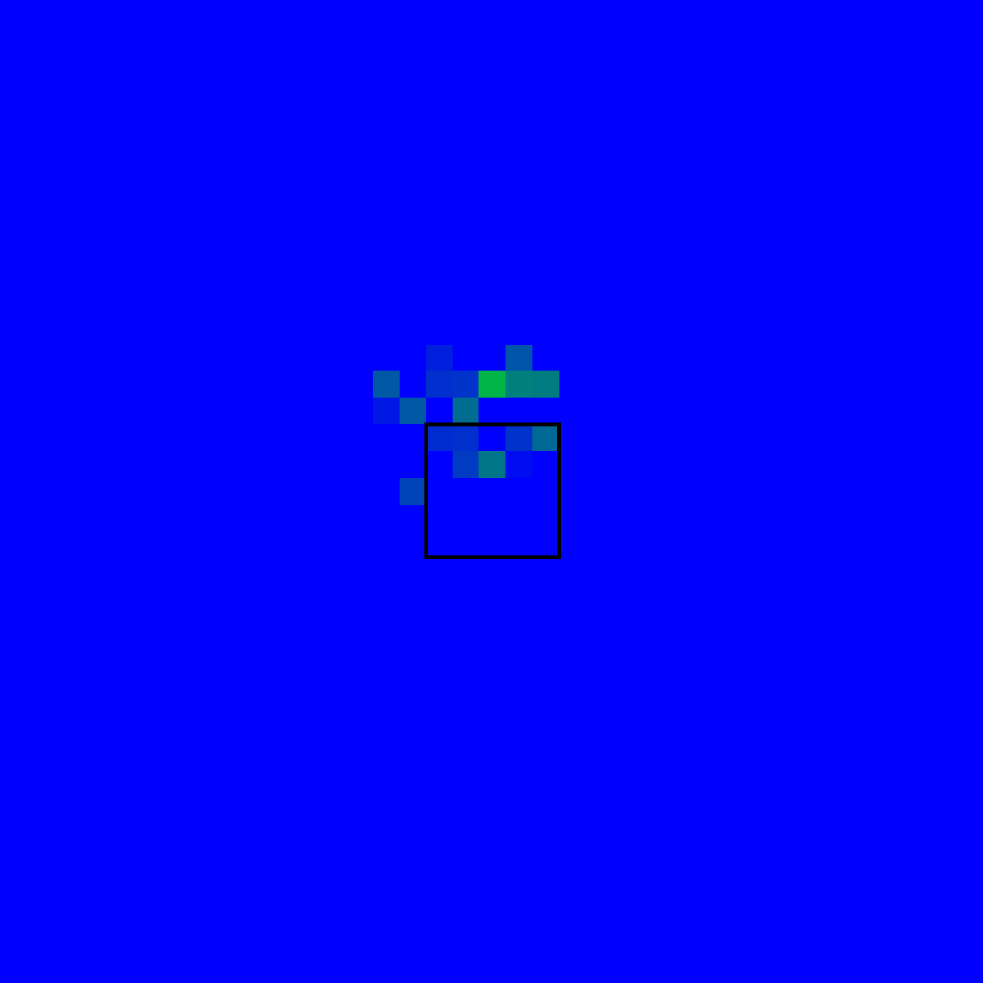}}
\centerline{\small{Kernel 2}}
\end{minipage}%
}
\subfigure{
\begin{minipage}[t]{0.46\linewidth}
\centerline{\includegraphics[width=0.99\linewidth]{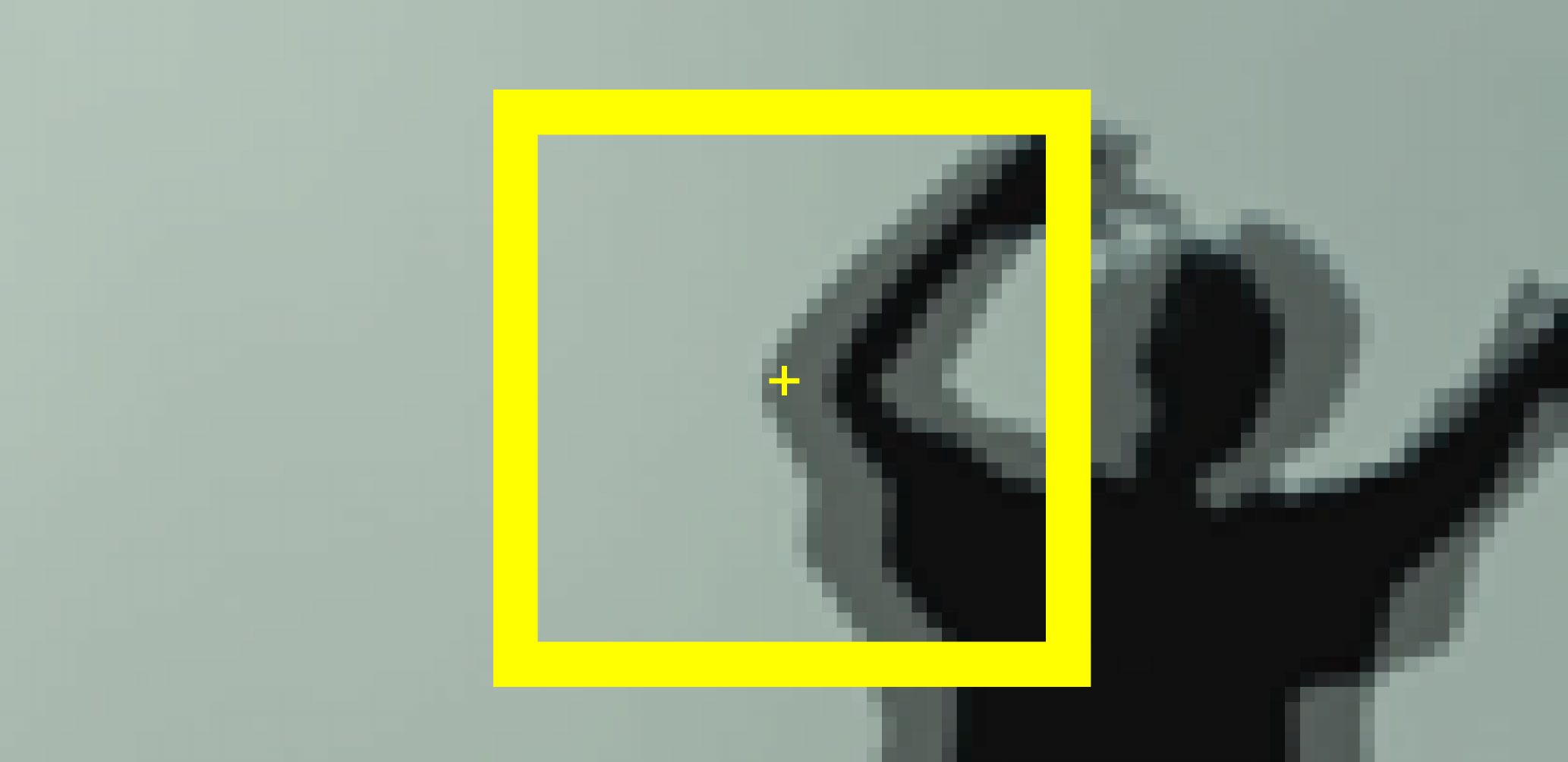}}
\centerline{\small{Overlayed}}
\centerline{\includegraphics[width=0.98\linewidth]{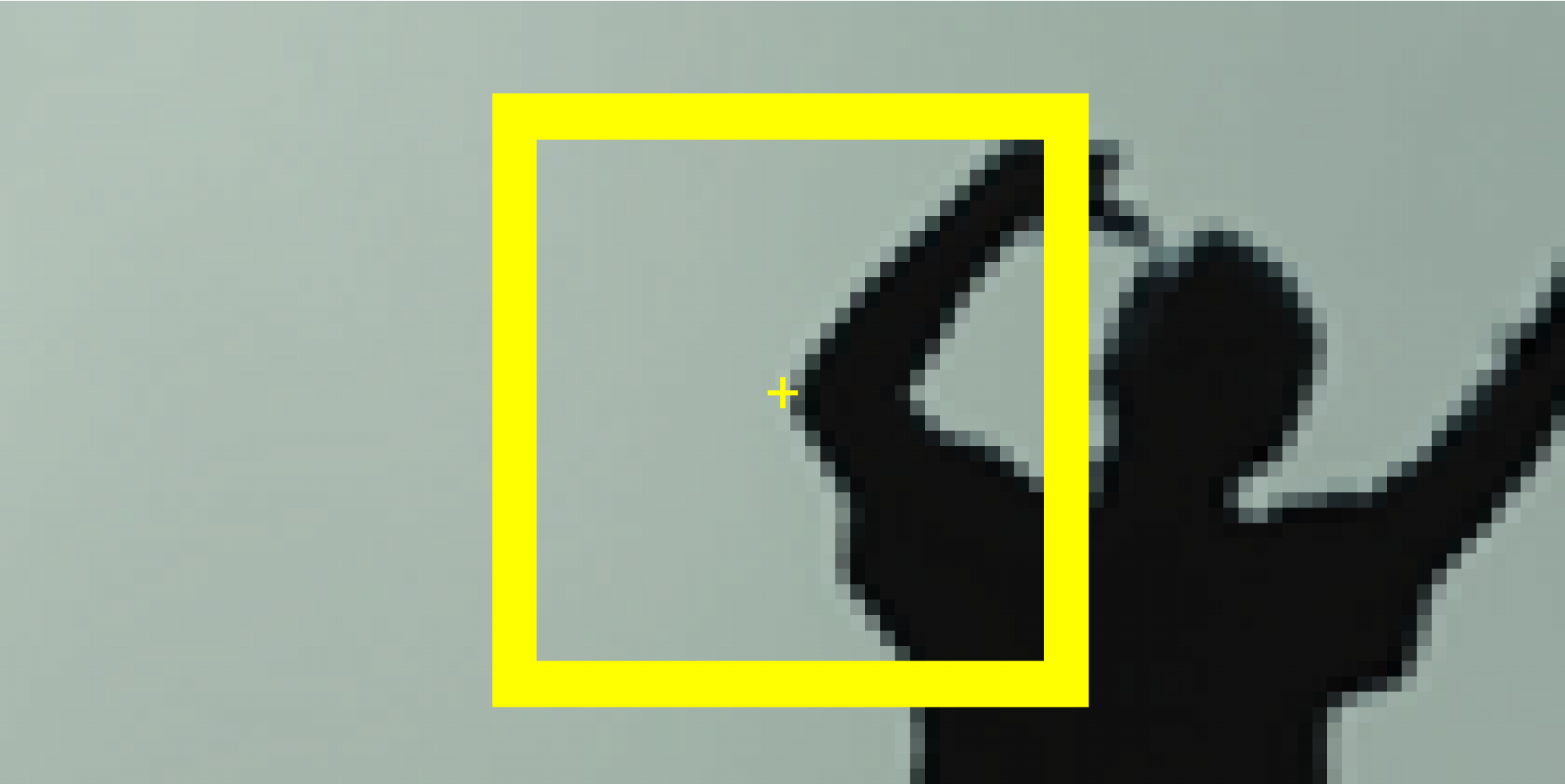}}
\centerline{\small{Synthesized}}
\end{minipage}
}
\centering
\caption{By convolving the patches with corresponding kernels we can get the final synthesized frame. We show the effective sampling locations of occlusion area centered at yellow \textbf{+} in the synthesized frame. The second row provides the magnified views of non-zero kernel values, in which the black rectangle represents the local kernel and greener regions indicate higher absolute values. By learning offsets, our method can obtain information outside the regular local kernel.}
\label{fig8}
\end{figure}

\begin{figure}[!th]
\centering

\subfigure{
\begin{minipage}[t]{0.21\linewidth}
\centerline{\includegraphics[width=1.05\linewidth]{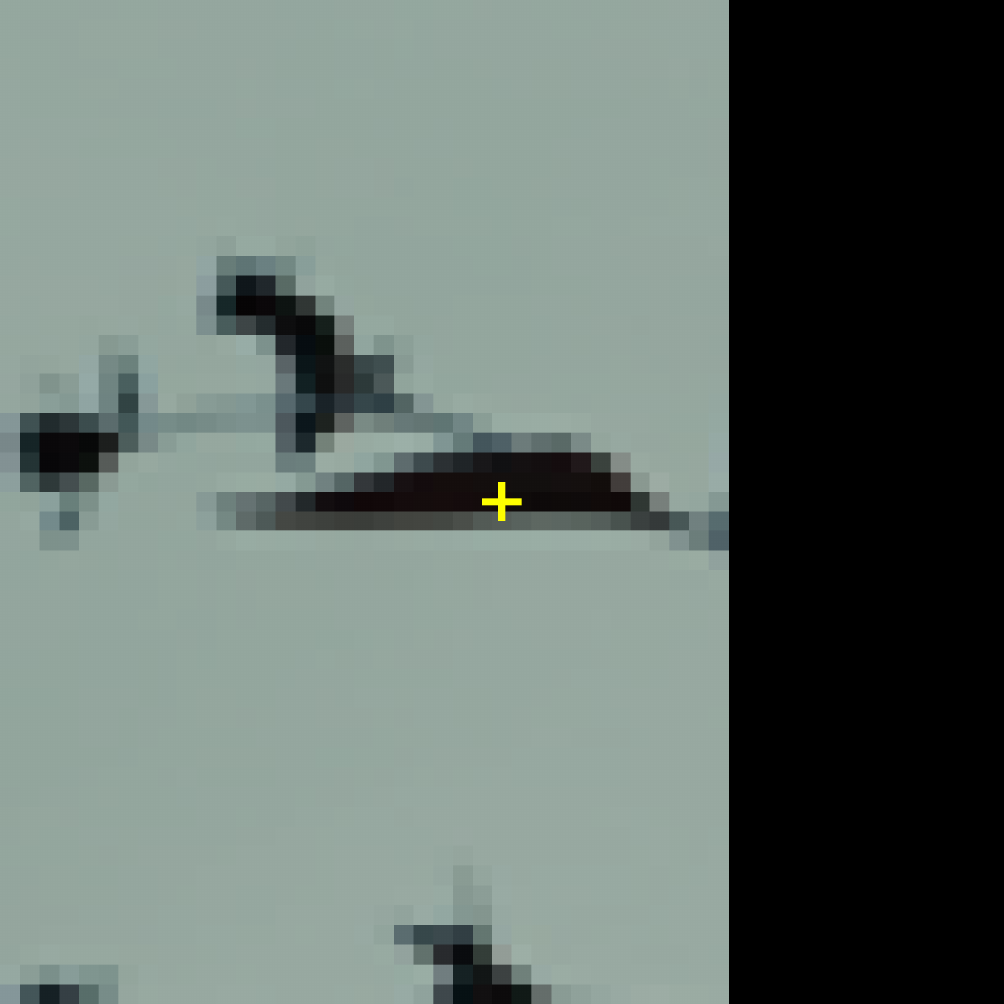}}
\centerline{\small{Patch 1}}
\vspace{0.9mm}
\centerline{\includegraphics[width=1.05\linewidth]{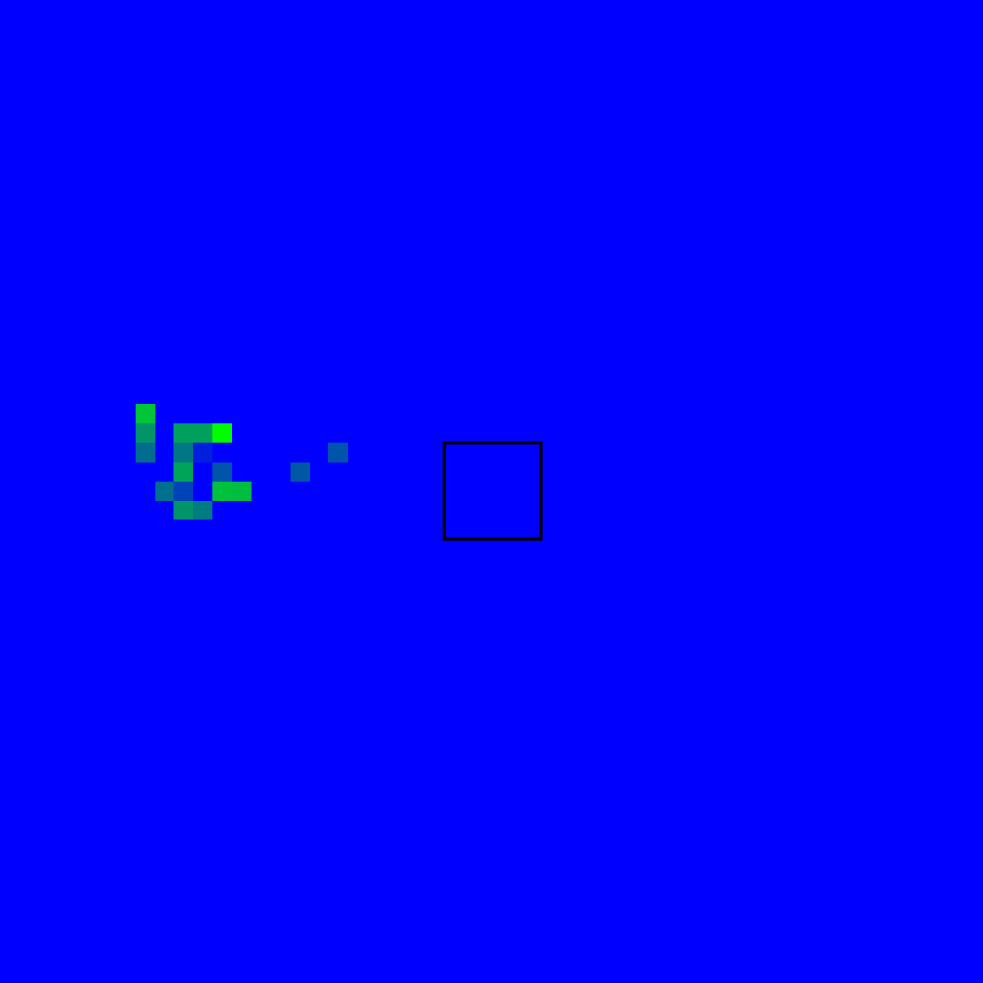}}
\centerline{\small{Kernel 1}}
\end{minipage}
}
\subfigure{
\begin{minipage}[t]{0.21\linewidth}
\centerline{\includegraphics[width=1.05\linewidth]{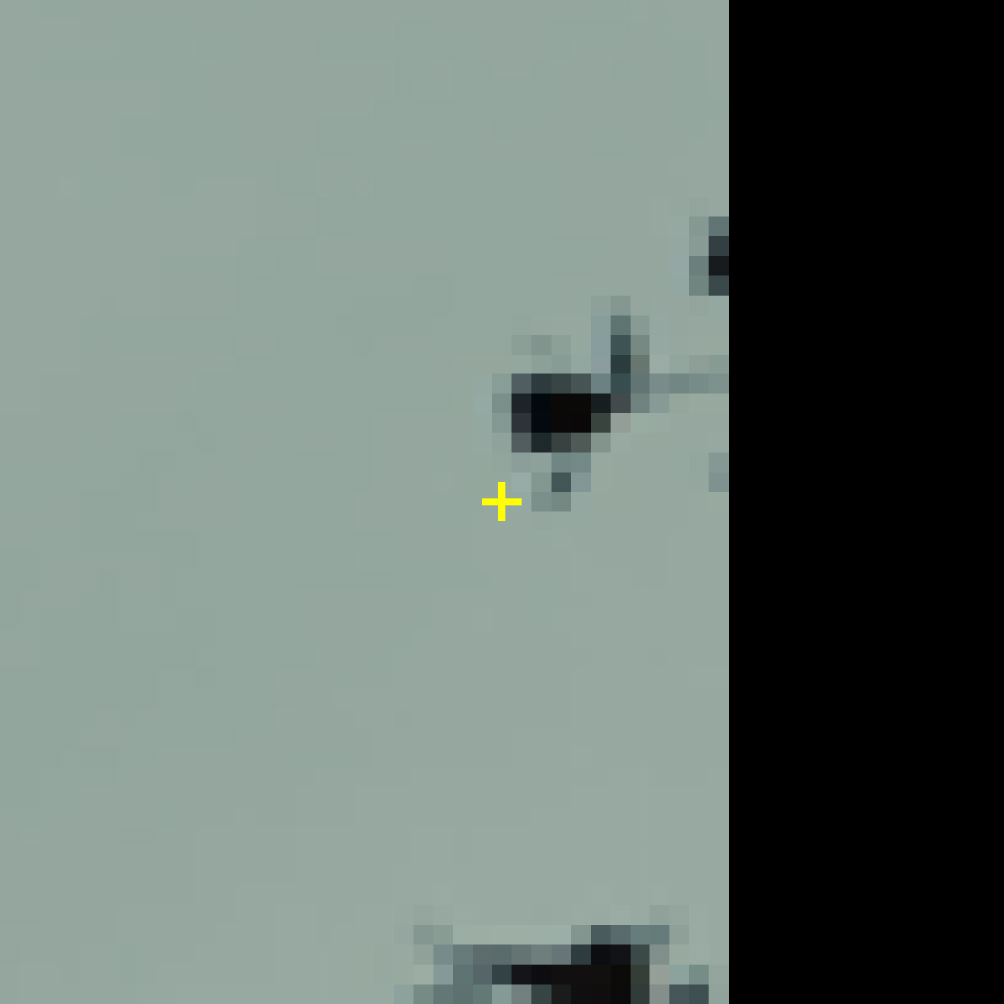}}
\centerline{\small{Patch 2}}
\vspace{0.9mm}
\centerline{\includegraphics[width=1.05\linewidth]{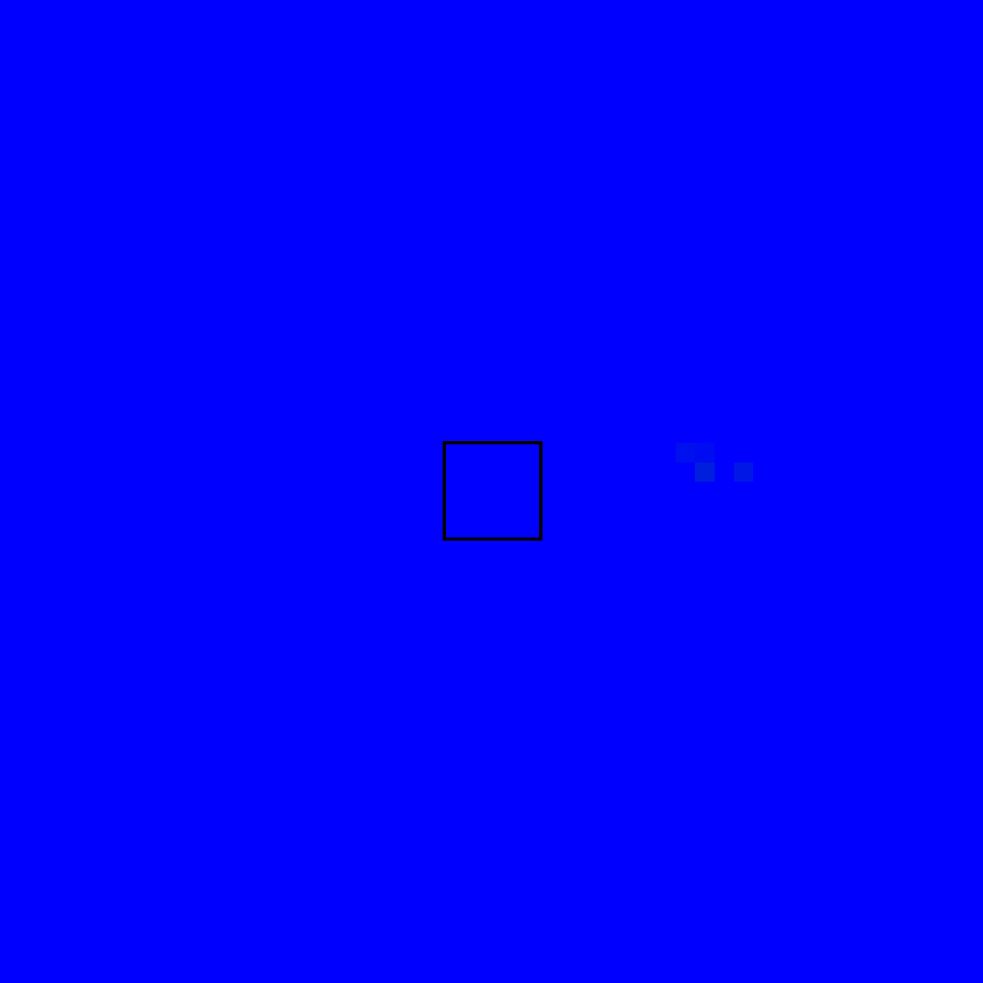}}
\centerline{\small{Kernel 2}}
\end{minipage}%
}
\subfigure{
\begin{minipage}[t]{0.46\linewidth}
\centerline{\includegraphics[width=0.96\linewidth]{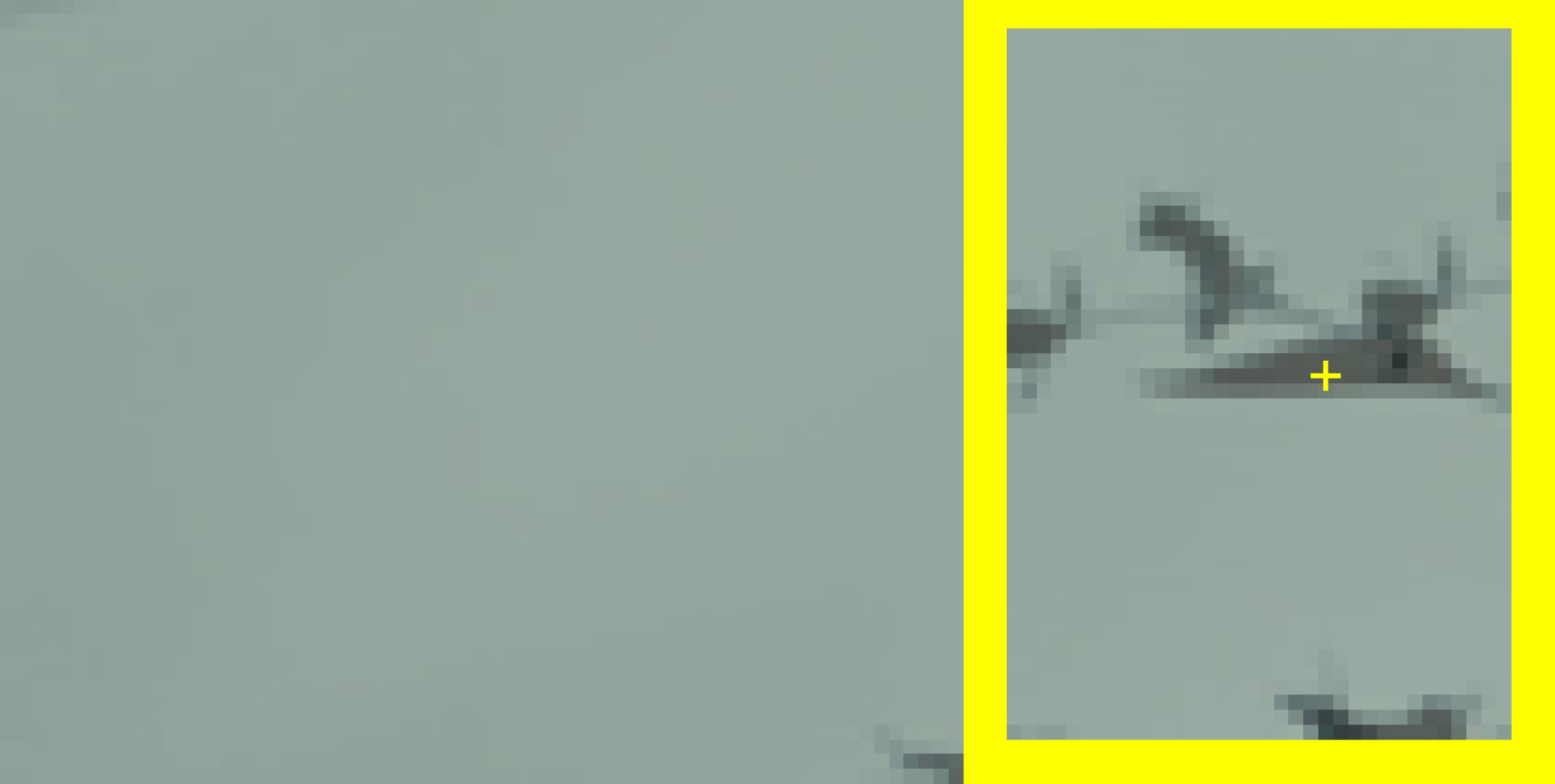}}
\centerline{\small{Overlayed}}
\centerline{\includegraphics[width=0.96\linewidth]{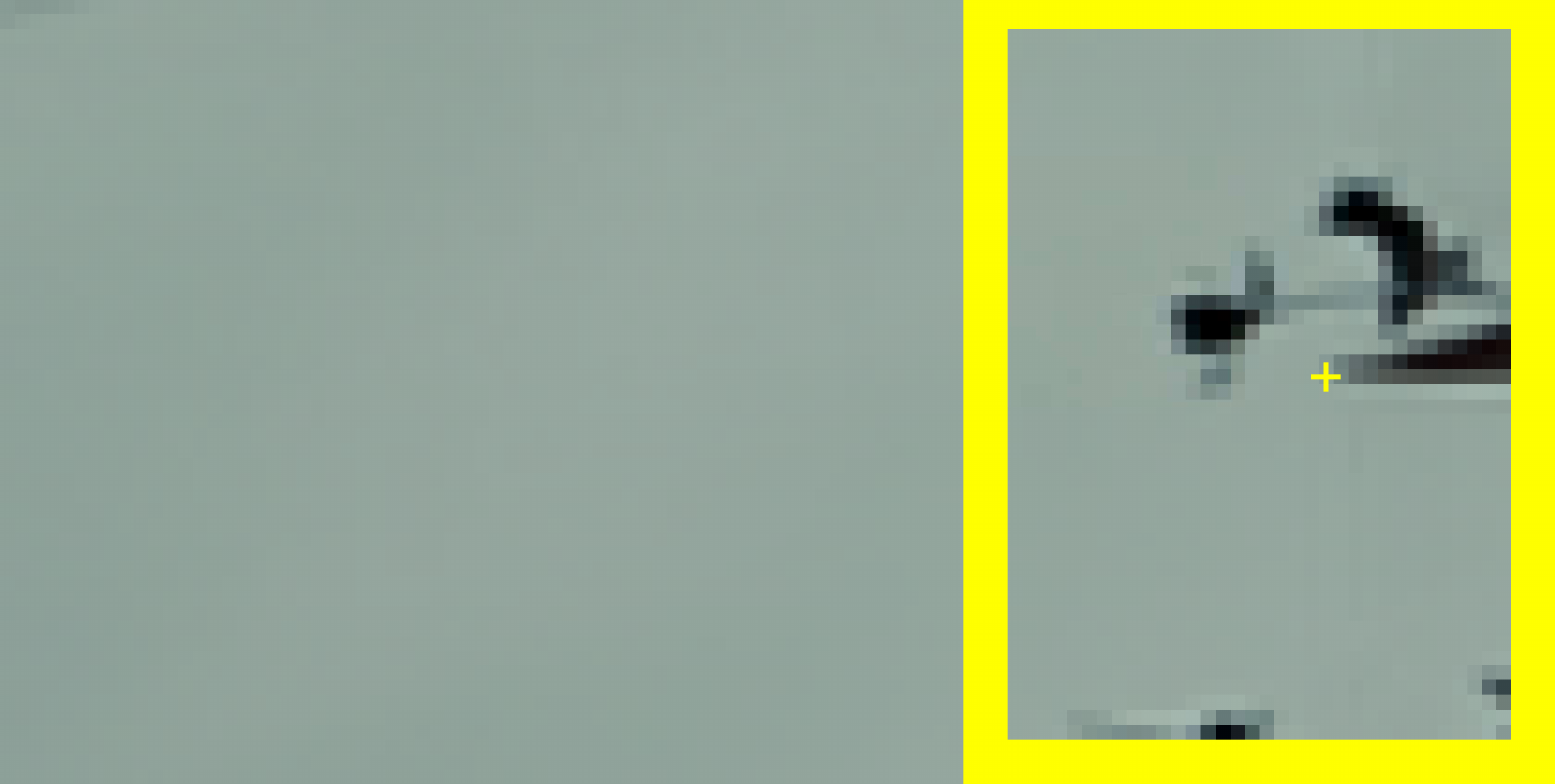}}
\centerline{\small{Synthesized}}
\end{minipage}
}
\centering
\caption{Effective sampling locations of occlusion area centered at yellow \textbf{+} in the synthesized frame. The reference patches are padded to align the kernels. Caused by large motion, the occlusion is handled by taking pixels mainly from one of the reference patches.}
\label{fig9}
\end{figure}
Besides, our $\mathcal{L}_{F}$-trained model performs favorably against the others on most of the subsets in terms of LPIPS. The main reason
\begin{table}[t]
\centering
\caption{Quantitative comparisons against methods using warping operation guided by given optical flow. We report average interpolation errors of occlusion regions IE(O) and boundary regions IE(B) on the UCF101\cite{UCF101, DVF}, Vimeo90K\cite{Toflow} and Middlebury-\texttt{Other} \cite{middleburry} datasets.}
{
\resizebox{\linewidth}{!}{
\begin{tabular}{lccc}
\toprule
\multirow{2}*{Methods} &
\multicolumn{1}{c}{ UCF101 }&
\multicolumn{1}{c}{ Vimeo90K }&
\multicolumn{1}{c}{  M.B.-\texttt{Other} }\\

\cmidrule(lr){2-2}
\cmidrule(lr){3-3}
\cmidrule(lr){4-4}
& IE(O)$\downarrow$/IE(B)$\downarrow$ & IE(O)$\downarrow$/IE(B)$\downarrow$ & IE(O)$\downarrow$/IE(B)$\downarrow$\\

\midrule
ToFlow & 6.89 / 2.08 & 6.00 / 2.59 & 5.05 / 2.14\\
MEMC-Net$^*$ & 6.72 / 1.89 & 5.42 / 2.55 & 4.33 / 2.01\\
DAIN & 6.71 / 1.85 & 5.22 / 2.41 & 4.24 / 1.97\\
\midrule
Ours-$\mathcal{L}_{C}$ & \textbf{6.62} / \textbf{1.83} & \textbf{5.20} / \textbf{2.09} & \textbf{4.04} / \textbf{1.72}\\
\bottomrule
\end{tabular}}}
\label{tab:7}
\end{table}
\begin{figure}[t]
\centering
\subfigure{
\begin{minipage}[t]{0.11\linewidth}
\centerline{\includegraphics[width=1.2\linewidth]{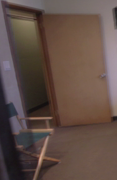}}
\vspace{1.0mm}
\centerline{\includegraphics[width=1.2\linewidth]{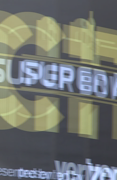}}
\centerline{\footnotesize{Overlayed}}
\end{minipage}
}
\subfigure{
\begin{minipage}[t]{0.11\linewidth}
\centerline{\includegraphics[width=1.2\linewidth]{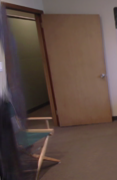}}
\vspace{1.0mm}
\centerline{\includegraphics[width=1.2\linewidth]{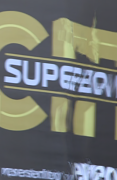}}
\centerline{\footnotesize{ToFlow}}
\end{minipage}
}
\subfigure{
\begin{minipage}[t]{0.11\linewidth}
\centerline{\includegraphics[width=1.2\linewidth]{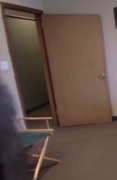}}
\vspace{1.0mm}
\centerline{\includegraphics[width=1.2\linewidth]{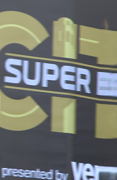}}
\centerline{\footnotesize{M.N.}}
\end{minipage}
}
\subfigure{
\begin{minipage}[t]{0.11\linewidth}
\centerline{\includegraphics[width=1.2\linewidth]{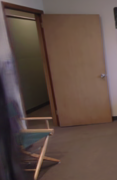}}
\vspace{1.0mm}
\centerline{\includegraphics[width=1.2\linewidth]{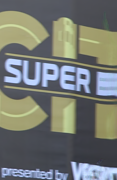}}
\centerline{\footnotesize{DAIN}}
\end{minipage}
}
\subfigure{
\begin{minipage}[t]{0.11\linewidth}
\centerline{\includegraphics[width=1.2\linewidth]{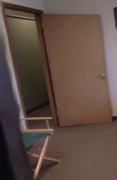}}
\vspace{1.0mm}
\centerline{\includegraphics[width=1.2\linewidth]{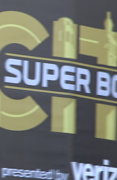}}
\centerline{\footnotesize{Ours-$\mathcal{L}_{C}$}}
\end{minipage}
}
\subfigure{
\begin{minipage}[t]{0.11\linewidth}
\centerline{\includegraphics[width=1.2\linewidth]{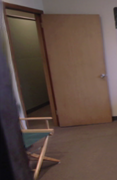}}
\vspace{1.0mm}
\centerline{\includegraphics[width=1.2\linewidth]{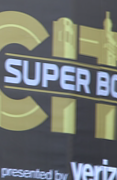}}
\centerline{\footnotesize{Ours-$\mathcal{L}_{F}$}}
\end{minipage}
}
\subfigure{
\begin{minipage}[t]{0.11\linewidth}
\centerline{\includegraphics[width=1.2\linewidth]{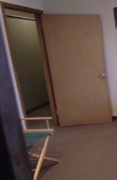}}
\vspace{1.0mm}
\centerline{\includegraphics[width=1.2\linewidth]{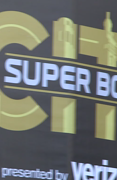}}
\centerline{\footnotesize{G.T.}}
\end{minipage}
}
\centering
\caption{We crop the interpolation frames from boundary regions with significant occlusion (the left part in the top row, the right part in the bottom row). M.N. is short for MEMC-Net$^*$ and G.T. is the abbreviation for ground truth.}
\label{fig10}
\end{figure}of these improvements is that our model better captures the content of source images by joint learning kernels, offsets, masks and biases.
Notice that despite $51\times51$ pixels are involved in SepConv for each pixel's synthesis, the available information is constrained in a local neighborhood and thousands of unrelated pixels make it prone to inaccuracies.

\subsubsection{Effect of occlusion handling}

Here, we use our $\mathcal{L}_{F}$-trained model to explain how our method handles occlusion and show two kinds of representative examples in Figures \ref{fig8} and \ref{fig9}, respectively.

\begin{figure*}[!th]
\centering
\subfigure{
\begin{minipage}[t]{0.15\linewidth}
\centerline{\includegraphics[width=1.05\linewidth]{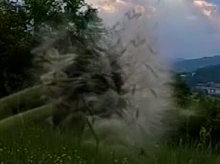}}
\centerline{Overlayed}
\end{minipage}
}
\subfigure{
\begin{minipage}[t]{0.15\linewidth}
\centerline{\includegraphics[width=1.05\linewidth]{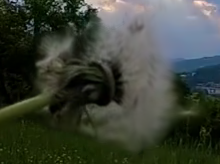}}
\centerline{$\alpha=0.00$ ($\mathcal{L}_{C}$)}
\end{minipage}
}
\subfigure{
\begin{minipage}[t]{0.15\linewidth}
\centerline{\includegraphics[width=1.05\linewidth]{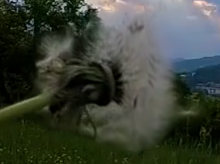}}
\centerline{$\alpha=0.25$}
\end{minipage}
}
\subfigure{
\begin{minipage}[t]{0.15\linewidth}
\centerline{\includegraphics[width=1.05\linewidth]{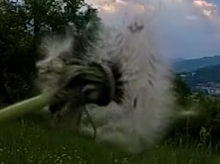}}
\centerline{$\alpha=0.50$}
\end{minipage}
}
\subfigure{
\begin{minipage}[t]{0.15\linewidth}
\centerline{\includegraphics[width=1.05\linewidth]{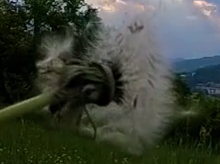}}
\centerline{$\alpha=0.75$}
\end{minipage}
}
\subfigure{
\begin{minipage}[t]{0.15\linewidth}
\centerline{\includegraphics[width=1.05\linewidth]{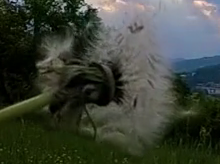}}
\centerline{$\alpha=1.00$ ($\mathcal{L}_{F}$)}
\end{minipage}
}
\centering
\caption{Qualitative evaluation on the effects of $\mathcal{L}_{C}$ and $\mathcal{L}_{F}$.
One can perform smooth control to produce different imagery effects by tweaking $\alpha$. }
\label{fig11}
\end{figure*}
It is noteworthy that each pair of 1D kernels is convolved to produce its equivalent 2D kernel for a better understanding. We also multiply the mask values by the kernel weights to emphasize the effective sampling locations and refer to the result as ``Kernel" for simplicity. Bias values are omitted because they are not fit for pixel-level visualization.

In Figure \ref{fig8}, we show a pixel from the background which is occluded by the elbow moving right. Despite that this pixel can be only seen in Patch 2, our method produces kernels that choose pixels with similar appearance from \emph{both the patches}.

Figure \ref{fig9} shows a pixel that moves outside the second frame, which always locates at the boundary areas. In this case, the pixel is only visible in Patch 1, and the generated kernels choose to sample corresponding pixels mainly from \emph{one of the patches} (Patch 1).

We further compare our approach with methods which utilize warping or adaptive warping operations \cite{MEMCNet,Toflow, dain} based on off-the-shelf optical flow estimators \cite{spynet,PWCNet} to see their abilities to handle occlusion. Since there is no labeled occlusion regions from the test datasets, we use as a measure of occlusion the brightness constancy $\bm{\mathrm{d}}$ that can be obtained by backward warping $\mathop{\omega}\limits ^{\leftarrow}$ operation \cite{middleburry, softsplat}.
Specifically, the IE of occluded regions is defined:
\begin{equation}
\begin{aligned}
\mathrm{IE_O}=\Big[\frac{1}{N_O}&\sum_{(x,y)}\big(\bm{\mathrm{\hat{I}}}(x,y) - \bm{\mathrm{I}}^{\mathrm{GT}}(x,y)\big)^2\Big]^{\frac{1}{2}},
\bm{\mathrm{d}}(x,y) \ge\mathrm{mean}(\bm{\mathrm{d}})\\
&\bm{\mathrm{d}}=\Big(\big\| \bm{\mathrm{I}}_1 - \mathop{\omega}\limits ^{\leftarrow}(\bm{\mathrm{I}}_2, \bm{\mathrm{F}}_{1\to2}) \big\|_1^2\Big)^{\frac{1}{2}},
\end{aligned}
\label{eq12}
\end{equation}
where $N_O$ is the number of occluded pixels whose brightness constancy are bigger than the mean value of $\bm{\mathrm{d}}$ and $\bm{\mathrm{F}}_{1\to2}$ represents the optical flow calculated by PWC-Net\cite{PWCNet}.
Additionally, we report the average IE from boundary 10 pixels wide of the synthesized frames, a special region where obvious occlusion often occurs due to camera motion.
\begin{equation}
\begin{aligned}
&\mathrm{IE_B}=\Big[\frac{1}{N_B}\sum_{(x,y)}\big(\bm{\mathrm{\hat{I}}}(x,y) - \bm{\mathrm{I}}^{\mathrm{GT}}(x,y)\big)^2\Big]^{\frac{1}{2}},\\
&x\le 10 \quad\mathrm{or}\quad x\ge \mathrm{width}(\bm{\mathrm{\hat{I}}})-10 \quad\mathrm{or}\\
&y\le 10 \quad\mathrm{or}\quad y\ge \mathrm{height}(\bm{\mathrm{\hat{I}}})-10,
\end{aligned}
\label{eq13}
\end{equation}
where $N_B$ is the number of boundary pixels.
As shown in Table \ref{tab:7}, all the three methods perform worse than ours in terms of both occluded IE and boundary IE. In particular, we achieve considerable gains in boundary IE. This is because the warped frames guided by optical flow are prone to be inaccurate especially in occluded regions, making it more difficult for later post-processing to improve the quality. In Figure \ref{fig10}, when significant occlusion occurs in the boundary, our approach is able to produce better results with less blur.

\begin{table}[t]
\centering
\caption{Quantitative comparisons against the effect of interpolation coefficient.}
{
\resizebox{\linewidth}{!}{
\begin{tabular}{lccc}
\toprule
\multirow{2}*{$\alpha$} &
\multicolumn{1}{c}{ UCF101 }&
\multicolumn{1}{c}{ Vimeo90K }&
\multicolumn{1}{c}{  M.B.-\texttt{Other} }\\

\cmidrule(lr){2-2}
\cmidrule(lr){3-3}
\cmidrule(lr){4-4}
& PSNR$\uparrow$/LPIPS$\downarrow$ & PSNR$\uparrow$/LPIPS$\downarrow$ & IE$\downarrow$/LPIPS$\downarrow$\\
\midrule
0.00 ($\mathcal{L}_{C}$) & \textbf{35.13} / 0.029 & \textbf{34.84} / 0.026 & 2.02 / 0.020\\
\midrule
0.25 & 35.12 / 0.026 & 34.82 / 0.021 & \textbf{2.00} / 0.015\\
0.50 & 35.08 / 0.024 & 34.75 / 0.018 & 2.02 / 0.012\\
0.75 & 35.02 / \textbf{0.023} & 34.66 / 0.017 & 2.05 / 0.011\\
\midrule
1.00 ($\mathcal{L}_{F}$) & 34.78 / \textbf{0.023} & 34.49 / \textbf{0.016} & 2.15 / \textbf{0.010}\\
\bottomrule
\end{tabular}}}
\label{tab:8}
\end{table}

\subsubsection{Effect of loss functions}

We use two versions of loss functions to train our model by minimizing color and perceptual difference, respectively. Moreover, we can achieve a continuous transition between the effects of two loss functions by using Deep Network Interpolation (DNI) methodology \cite{dni}. To be more detailed, the model parameters of a new interpolated model can be derived by:
\begin{equation}
\theta_{interp} = (1-\alpha)\theta_{\mathcal{L}_{C}}+\alpha\theta_{\mathcal{L}_{F}},
\label{eq12}
\end{equation}
where $\theta$ represents network parameters and $\alpha\in[0,1]$ denotes the interpolation coefficient. As shown in Figure \ref{fig11}, $\mathcal{L}_{F}$-trained model recovers the details well whist model trained with $\mathcal{L}_{C}$ does not. By adjusting $\alpha$, the imagery effects change smoothly.

We further perform quantitative comparisons with different $\alpha$ values on the three datasets shown in Table \ref{tab:8}. Bigger $\alpha$ leads to better performance in terms of LPIPS whereas performs worse in terms of PSNR, which indicates that we can balance distortion and perceptual quality by simply changing $\alpha$ to meet different requirements of users.

\begin{table}[t]
\centering
\caption{Runtime of the proposed method (seconds).}
{
\begin{tabular}{lcccc}
\toprule
Resolution & Enc-Dec. & Estimators & D.C. & Total\\
\midrule
$448\times256$p & 0.019 & 0.013 & 0.002 & 0.034\\
$640\times480$p & 0.035 & 0.029 & 0.003 & 0.067\\
$1280\times720$p & 0.098 & 0.091 & 0.009 & 0.198\\
$1920\times1080$p & 0.211 & 0.202 & 0.020 & 0.433\\
\bottomrule
\end{tabular}}
\label{tab:9}
\end{table}

\begin{table}[t]
\centering
\caption{Runtime comparisons in seconds with existing methods on a $640\times480$ sequence.}
{
\resizebox{\linewidth}{!}{
\begin{tabular}{lp{0.9cm}p{0.9cm}|lp{0.9cm}p{0.9cm}}
\toprule
Methods & Processor & Runtime & Methods & Processor & Runtime\\
\midrule
AdaConv\cite{adaconv} & Titan X & 2.8 & CtxSyn\cite{Ctxsyn} &  Titan X & 0.07 \\
SepConv\cite{sepconv} & Titan X & 0.2 & ToFlow\cite{Toflow} &  Titan X & 0.393 \\
DSepConv\cite{dsepconv} & Titan X & 0.3 & CyclicGen\cite{cyclicgen} & --- & 0.088 \\
AdaCoF\cite{adacof} & RTX 2080 Ti & 0.03 & MS-PFT\cite{MSPFT} &  GTX 1080 & 0.44 \\
MEMC-Net$^*$\cite{MEMCNet} & Titan X & 0.12 & STAR-T$_{\mathrm{HR}}$\cite{STAR} &  Tesla V100 & 0.049 \\
DAIN\cite{dain} & Titan X & 0.13 & SoftSplat\cite{softsplat} &  Titan X & 0.1 \\
SuperSlomo\cite{superslomo} & --- & 0.5 & EDSC(ours) &  Titan X & 0.067 \\

\bottomrule
\end{tabular}}}
\label{tab:910}
\end{table}

\subsubsection{Execution speed}

Table \ref{tab:9} shows the runtime of each component of our method on a single NVIDIA Titan X GPU using sequences with different resolutions. Enc-Dec. is short for the encoder-decoder architecture and D.C. is short for the deformable convolution process which utilizes the learned components. We further compare the runtime between our method and some existing methods shown in Table \ref{tab:910}. For a fair comparison, we use the runtime of the ``Urban" sequence in the Middlebury \texttt{Evaluation} set, which is publicly available on the benchmark website. Since the runtimes were submitted by the authors themselves, we also list their processors from their papers. As we can see, our model runs faster than most of the existing methods.

\subsection{Ablation study}
In this section, we perform comprehensive ablations to analyse the major components of our method, including the settings of the encoder-decoder architecture, different sizes of the estimated kernels and the usage of mask and bias estimators.
\begin{table}[t]
\centering
\caption{Ablation experiments to quantitatively analyze the effect of different rate of HetConv in the encoder-decoder architecture.}
{
\resizebox{\linewidth}{!}{
\begin{tabular}{lccccc}
\toprule
\multirow{2}*{Rate} &
\multicolumn{1}{c}{ UCF101 }&
\multicolumn{1}{c}{ Vimeo90K }&
\multicolumn{1}{c}{ M.B. } &
\multirow{2}*{\shortstack{FLOPs\\(G)}} &
\multirow{2}*{\shortstack{Param.\\(M)}} \\

\cmidrule(lr){2-2}
\cmidrule(lr){3-3}
\cmidrule(lr){4-4}
& PSNR$\uparrow$/SSIM$\uparrow$ & PSNR$\uparrow$/SSIM$\uparrow$ & IE$\downarrow$\\
\midrule
1/1 & 35.11 / \textbf{0.969} & 34.70 / 0.974 & 2.05 & 19.4 & 21.9\\
1/2 & 35.06 / 0.968 & 34.72 / 0.974 & 2.07 & 16.3& 14.8\\
1/4 & \textbf{35.13} / 0.968 & \textbf{34.84} / \textbf{0.975} & \textbf{2.02} & 13.8 & 8.9\\
1/8 & 34.96 / 0.968 &34.61 / 0.973 & 2.13 & 12.5& 6.0\\
1/16 & 35.00 / 0.968 &34.55 / 0.973 & 2.15 & 11.9& 4.5\\
1/32 & 34.90 / 0.967 & 34.34 / 0.972 & 2.21 & 11.6& 3.8\\

\bottomrule
\end{tabular}}}
\label{tab10}
\end{table}
\subsubsection{Encoder-decoder architecture}
In typical kernel based interpolation methods, the encoder-decoder architecture occupies most of the network parameters (\emph{e.g.}, 97.7\% in SepConv \cite{sepconv} and 97.1\% in DSepConv \cite{dsepconv}).
To reduce model parameters, we replace some of the $3\times3$ filters into $1\times1$ in each convolution layer by using HetConv \cite{hetconv}, leaving only a specific rate (1/$P$ in \cite{hetconv}) of $3\times3$ kernels out of total kernels. As shown in Table \ref{tab10}, the FLOPs and the number of network parameters decrease when the rate getting smaller. The model performs the best when rate equals to 1/4, indicating that we can find a good balance between accuracy and computation, which is in line with the findings in \cite{hetconv} for the task of classification.

\subsubsection{Generated kernel size}
\begin{figure}[!t]
\centering
\subfigure[Inputs]{
\begin{minipage}[b]{0.34\linewidth}
\centerline{\includegraphics[width=1\linewidth]{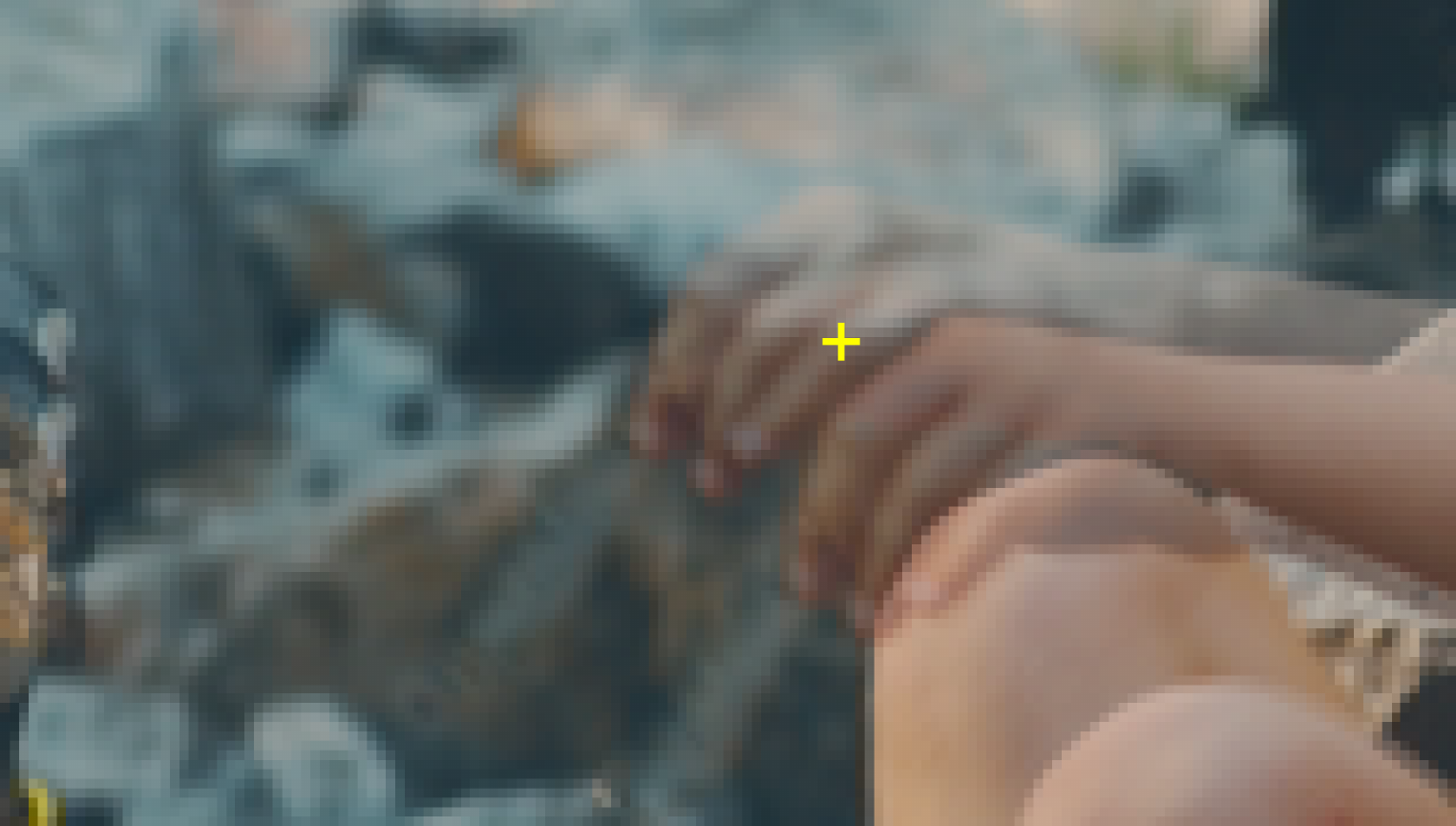}}
\end{minipage}
\hspace{-1.5mm}
\begin{minipage}[b]{0.1\linewidth}
\centerline{\includegraphics[width=0.935\linewidth]{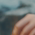}}
\vspace{0.5mm}
\centerline{\includegraphics[width=0.935\linewidth]{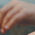}}
\end{minipage}
}
\subfigure[$n$=1]{
\begin{minipage}[b]{0.34\linewidth}
\centerline{\includegraphics[width=1\linewidth]{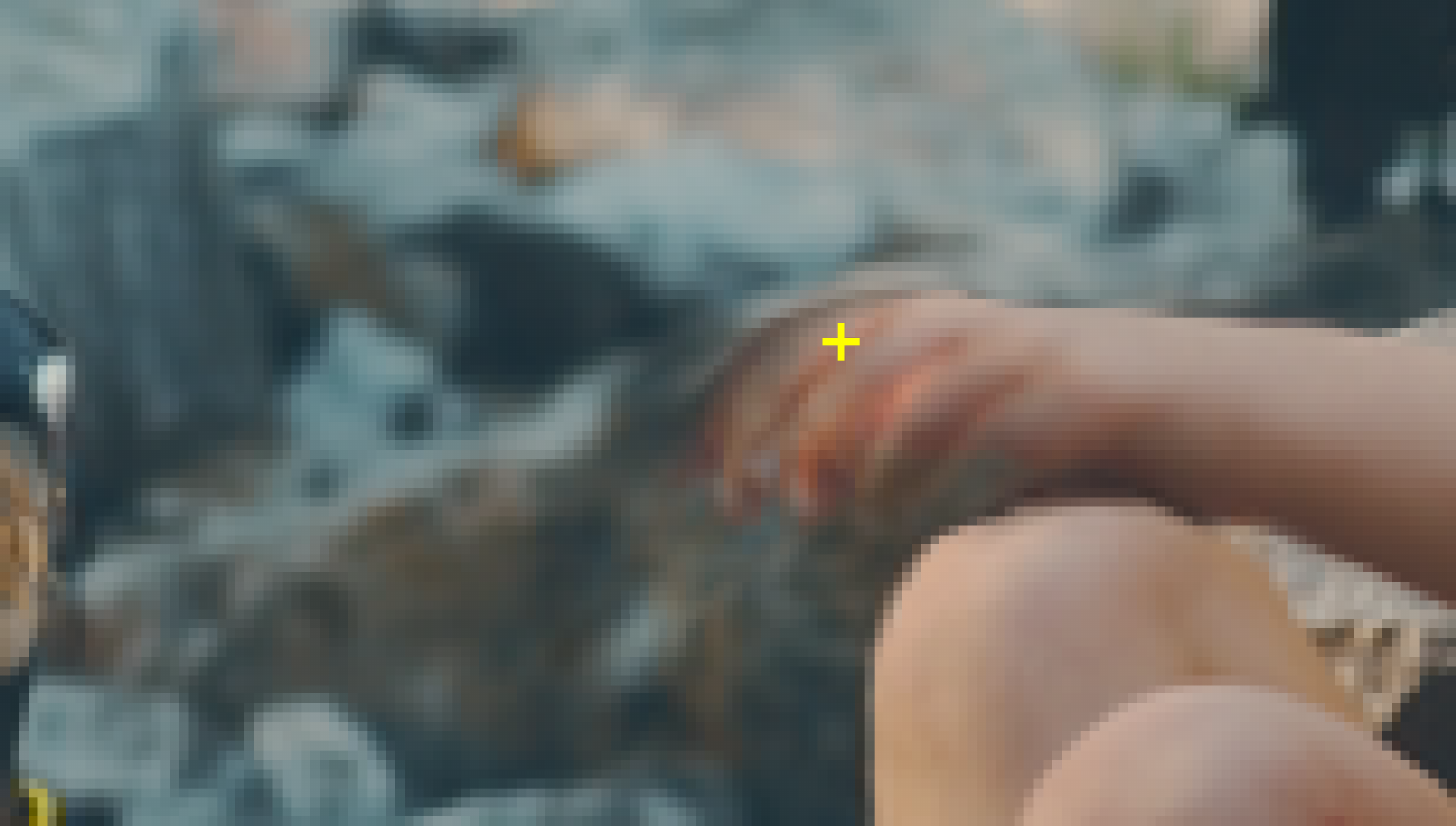}}
\end{minipage}
\hspace{-1.5mm}
\begin{minipage}[b]{0.1\linewidth}
\centerline{\includegraphics[width=0.935\linewidth]{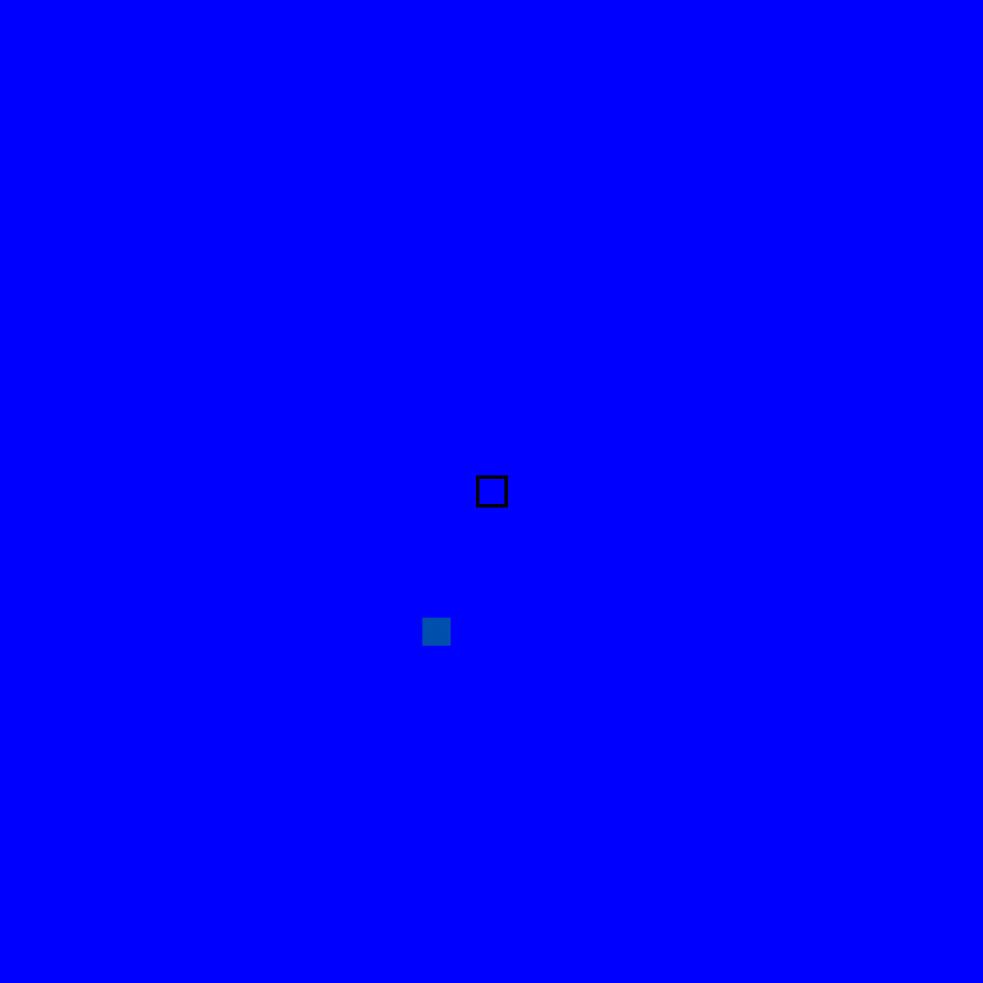}}
\vspace{0.5mm}
\centerline{\includegraphics[width=0.935\linewidth]{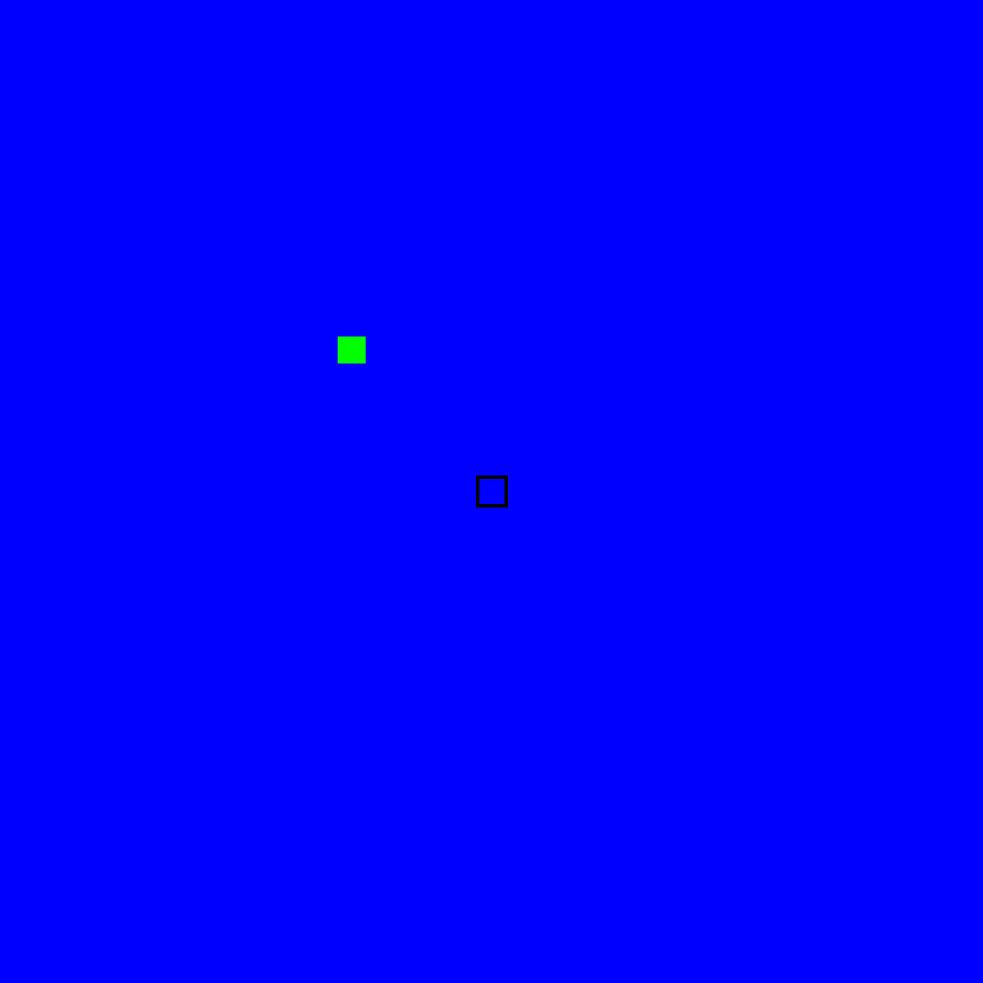}}
\end{minipage}
}

\subfigure[$n$=3]{
\begin{minipage}[b]{0.34\linewidth}
\centerline{\includegraphics[width=1\linewidth]{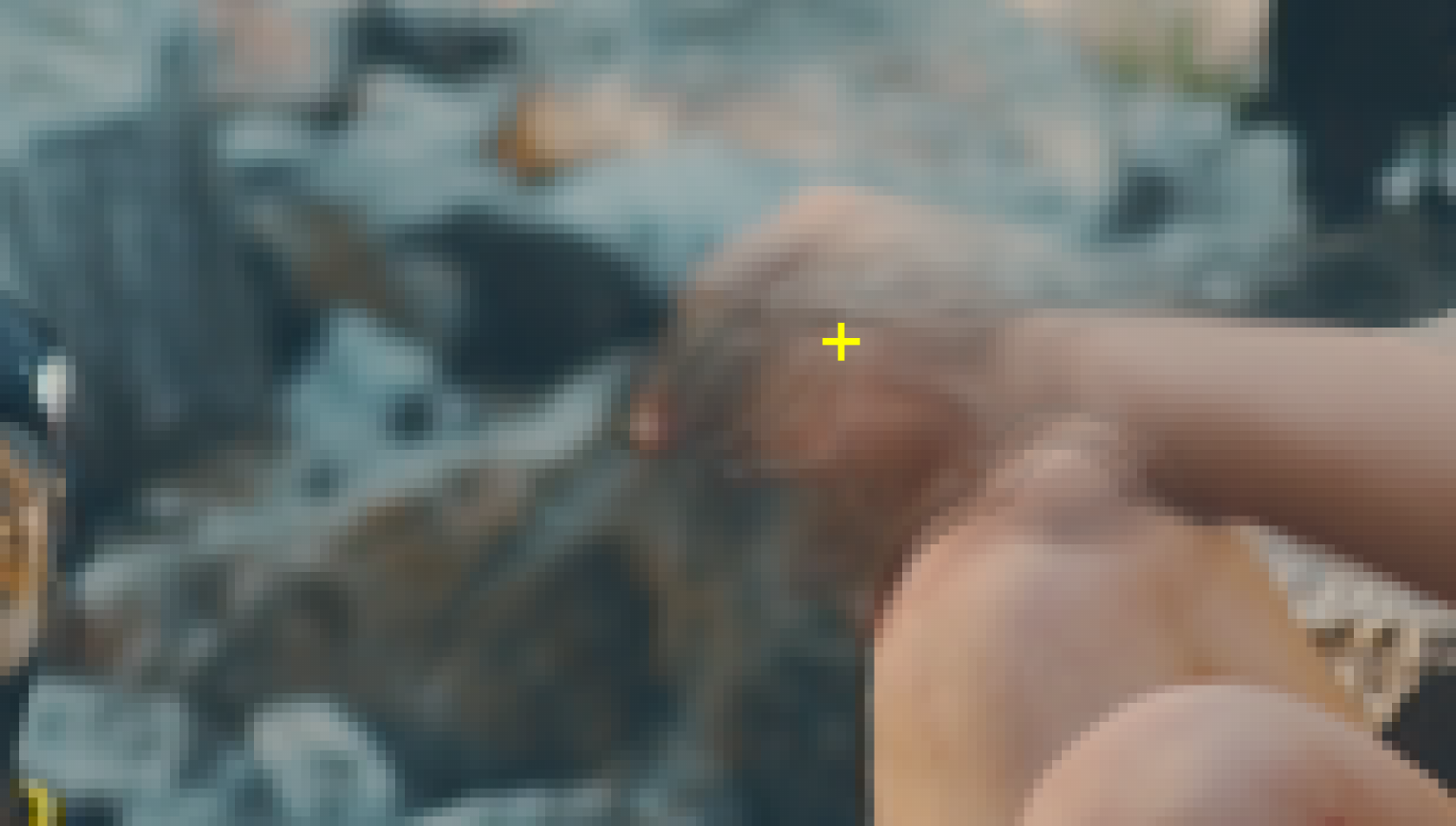}}
\end{minipage}
\hspace{-1.5mm}
\begin{minipage}[b]{0.1\linewidth}
\centerline{\includegraphics[width=0.935\linewidth]{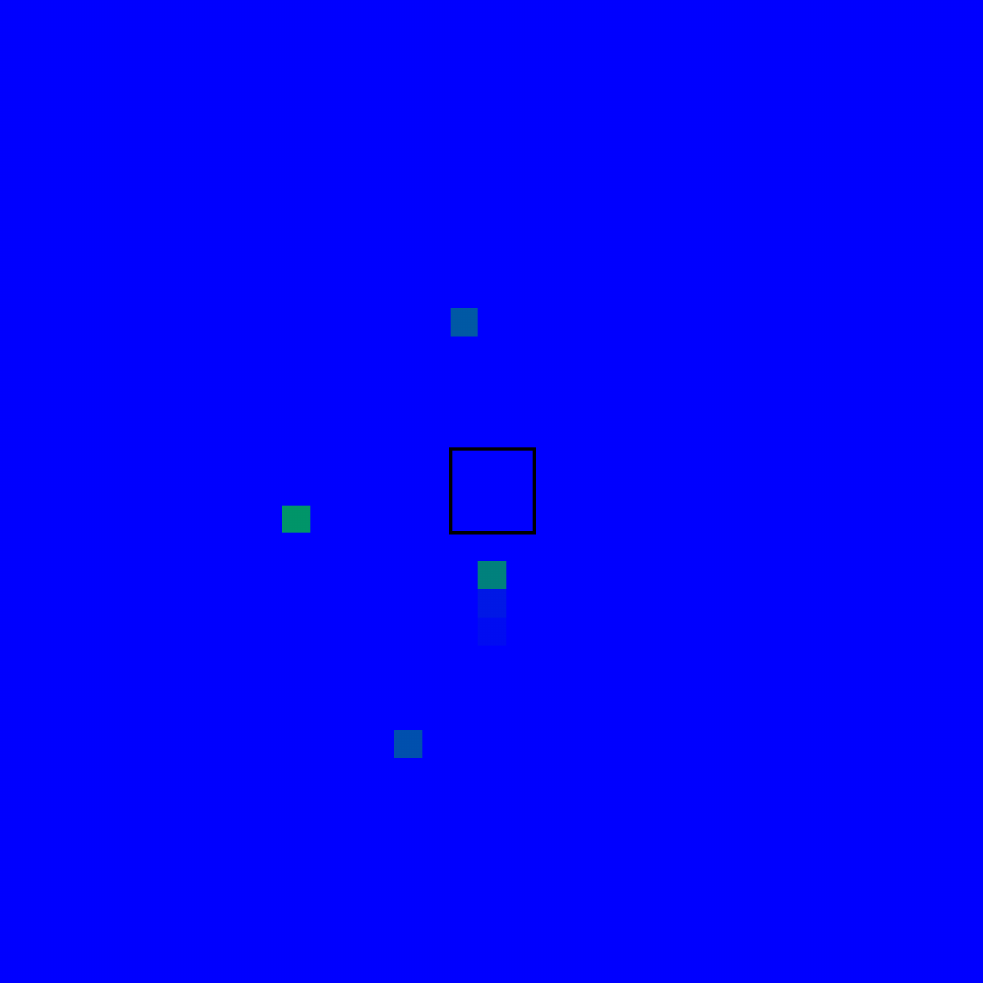}}
\vspace{0.5mm}
\centerline{\includegraphics[width=0.935\linewidth]{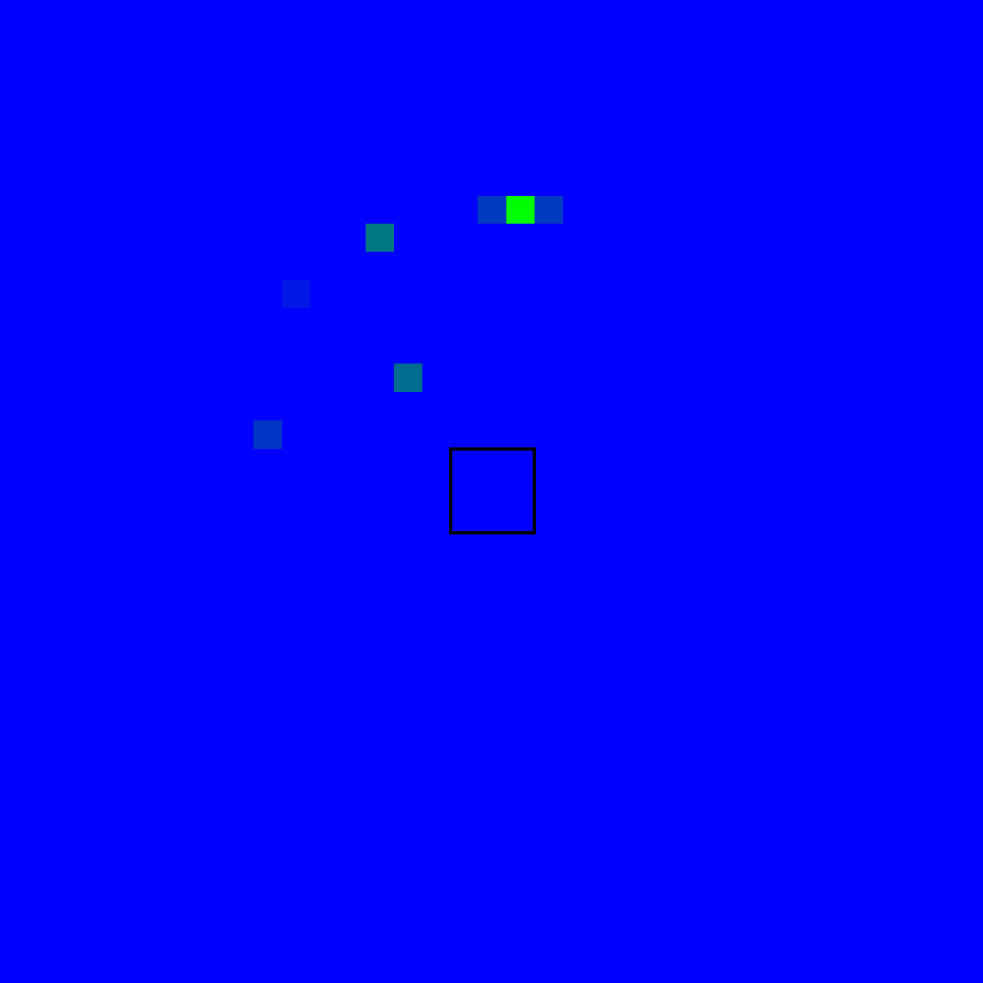}}
\end{minipage}
}
\subfigure[$n$=5]{
\begin{minipage}[b]{0.34\linewidth}
\centerline{\includegraphics[width=1\linewidth]{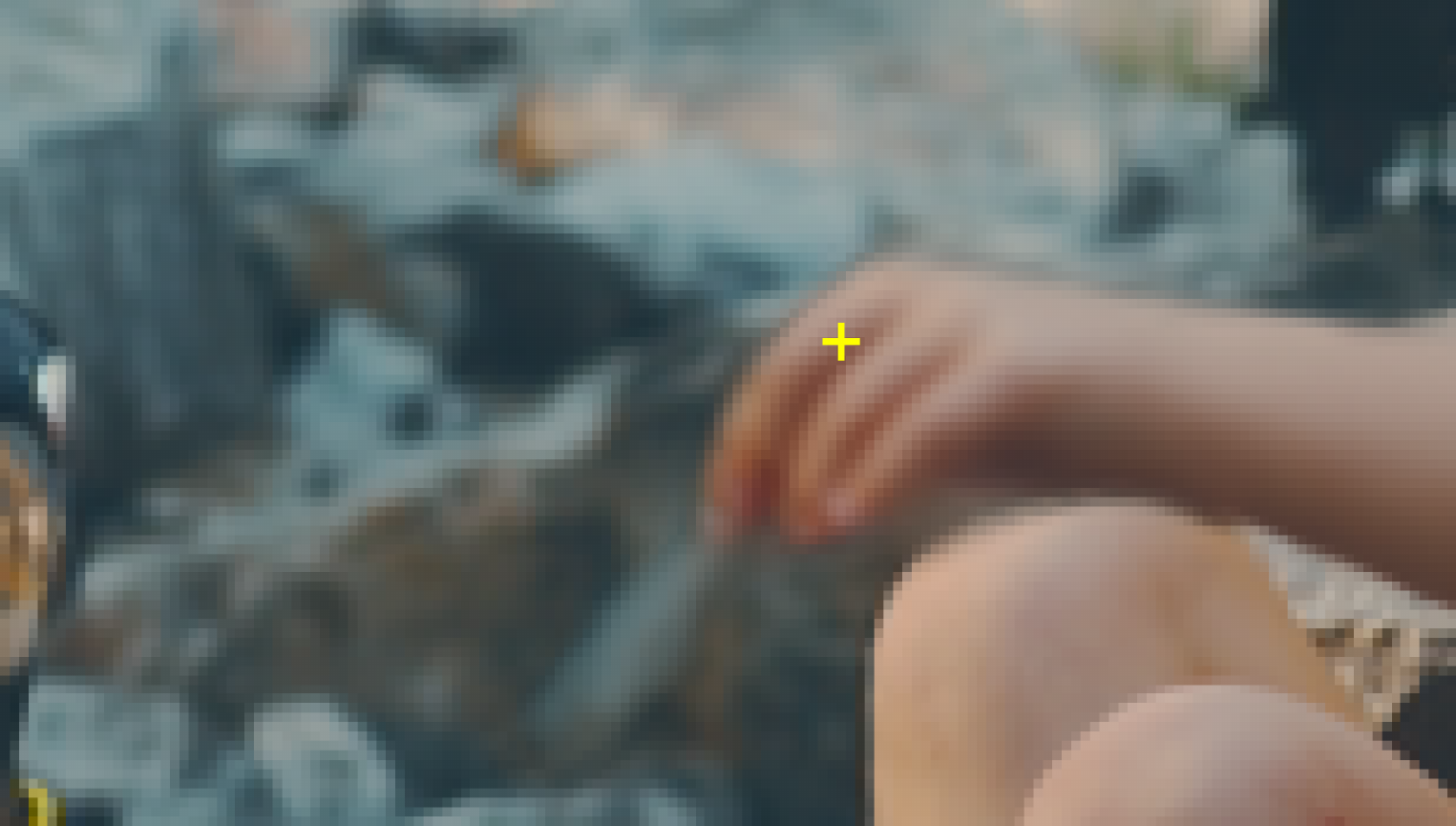}}
\end{minipage}
\hspace{-1.5mm}
\begin{minipage}[b]{0.1\linewidth}
\centerline{\includegraphics[width=0.935\linewidth]{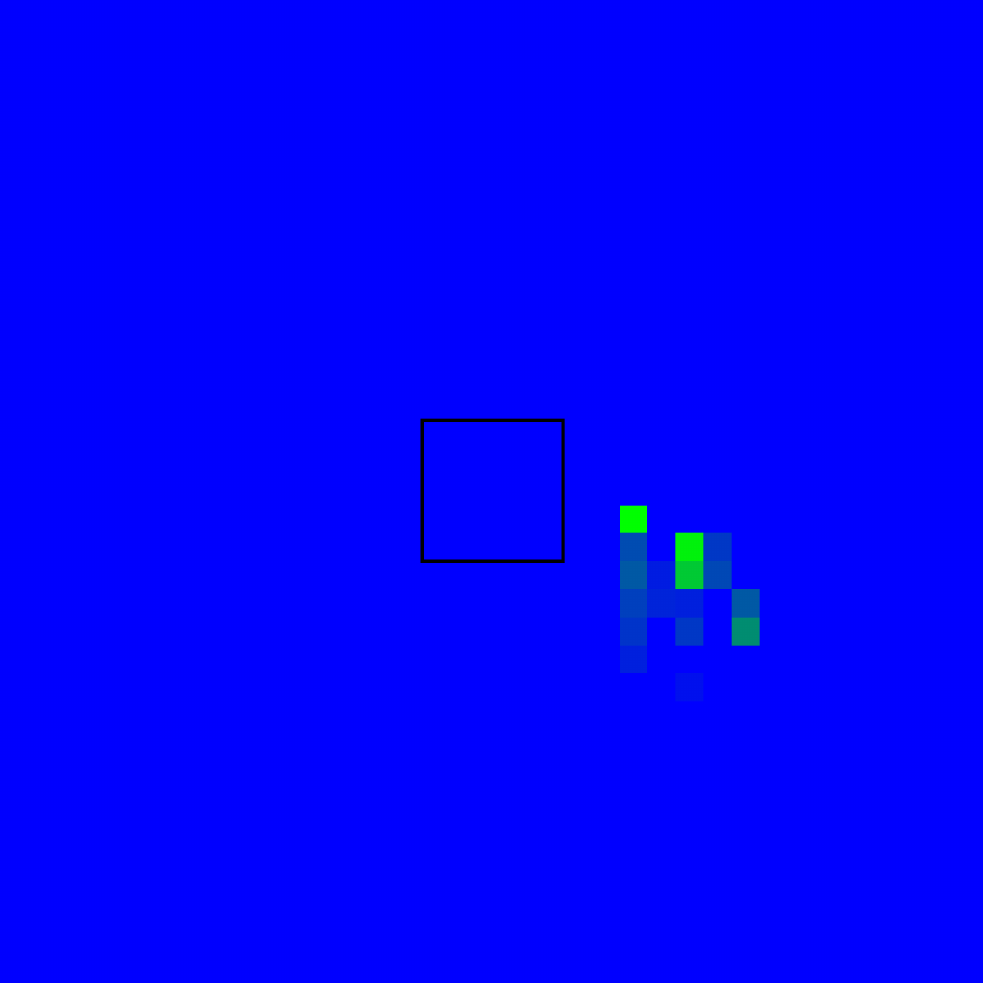}}
\vspace{0.5mm}
\centerline{\includegraphics[width=0.935\linewidth]{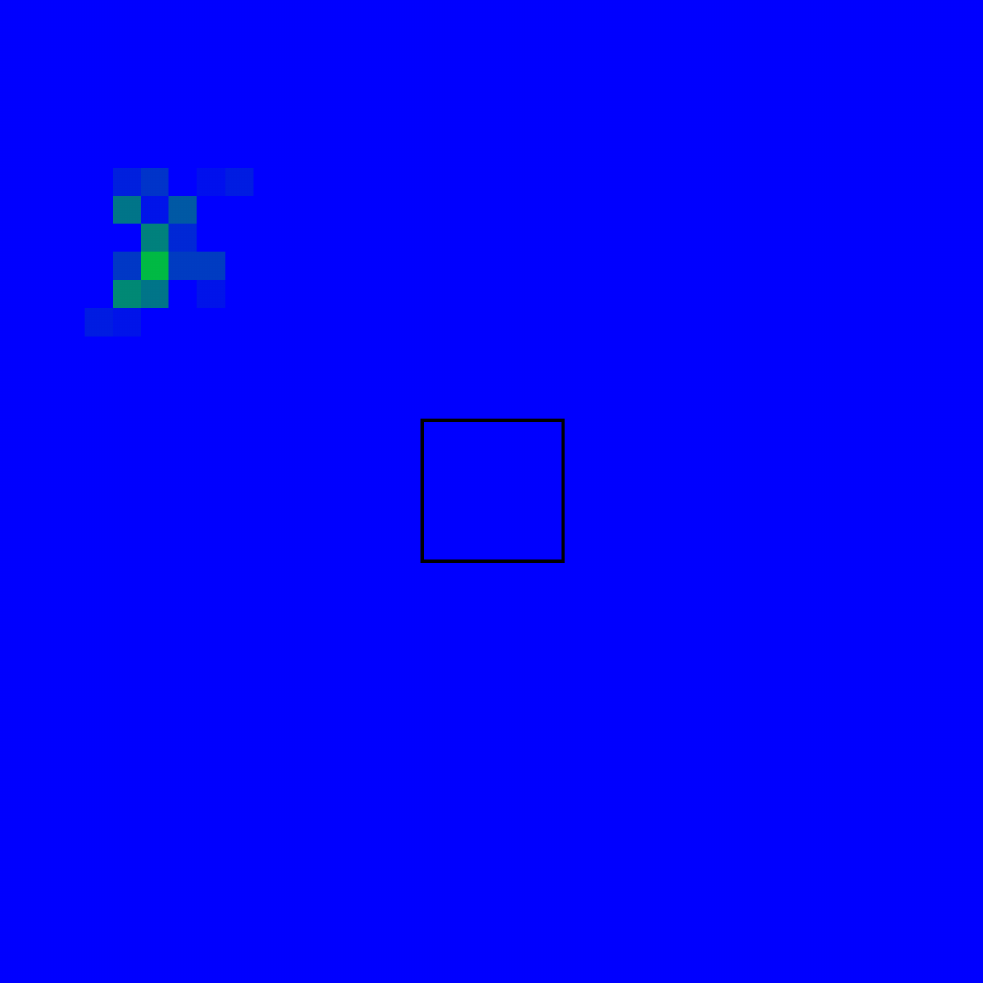}}
\end{minipage}
}
\centering
\caption{Synthesizing frames with different kernel size $n$. We show the overlayed frame and input patches centered at the yellow \textbf{+} in (a). Synthesized frames and effective sampling locations are shown in (b), (c) and (d). Please zoom in the figures for a better view.}
\label{fig15}
\end{figure}

For each pixel to be synthesized, the generated kernel size $n$ indicates how many pixels in the non-regular grid augmented with offsets could be used. Larger $n$ enables the network to reference more pixels but it inevitably has more FLOPs and runtime. As shown in Table \ref{tab11}, the performance improves but the computation (FLOPs) and runtime increase when using larger kernel sizes. Please note that we do not recommend to use network with kernel size larger than 5 (\emph{e.g.}, $n=7,9,11$) because they increase by 39.8\%, 128.1\% and 294.8\% in terms of FLOPs and by 25.4\%, 65.7\% and 119.4\% in terms of runtime, compared to $n=5$. Figure \ref{fig15} shows an example of the effect of different kernel sizes. We also choose pixels from the synthesized frames indicated by the yellow \textbf{+} and visualize the effective sampling locations. We see that the proposed model with $n=5$ can correctly use pixels from lower right of the first patch and upper left of the second, producing the best result.

\begin{table}[t]
\centering
\caption{Ablation experiments to quantitatively analyze the effect of different kernel size.}
{
\resizebox{\linewidth}{!}{
\begin{tabular}{cccccc}
\toprule
\multirow{2}*{Size} &
\multicolumn{1}{c}{ UCF101 }&
\multicolumn{1}{c}{ Vimeo90K }&
\multicolumn{1}{c}{ M.B. } &
\multirow{2}*{\shortstack{FLOPs\\(G)}} &
\multirow{2}*{\shortstack{Runtime\\(s)}} \\

\cmidrule(lr){2-2}
\cmidrule(lr){3-3}
\cmidrule(lr){4-4}
& PSNR$\uparrow$/SSIM$\uparrow$ & PSNR$\uparrow$/SSIM$\uparrow$ & IE$\downarrow$\\
\midrule
1$\times$1 & 34.83 / 0.967 & 33.47 / 0.965 & 2.65 & 11.4 & 0.063\\
3$\times$3 & 34.99 / \textbf{0.968} & 34.59 / 0.973 & 2.16 & 11.8 & 0.064\\
5$\times$5 & \textbf{35.13} / \textbf{0.968} & \textbf{34.84} / \textbf{0.975} & \textbf{2.02} & 13.8 & 0.067\\
\bottomrule
\end{tabular}}}
\label{tab11}
\end{table}

\begin{figure}[t]
\centering
\subfigure[Frame 1]{\label{fig16:subfig:a}
\includegraphics[width=0.3\linewidth]{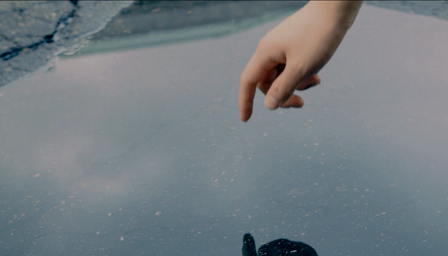}}
\subfigure[Frame 2]{\label{fig16:subfig:b}
\includegraphics[width=0.3\linewidth]{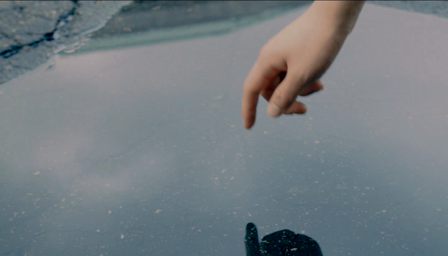}}
\subfigure[Frame 3]{\label{fig16:subfig:c}
\includegraphics[width=0.3\linewidth]{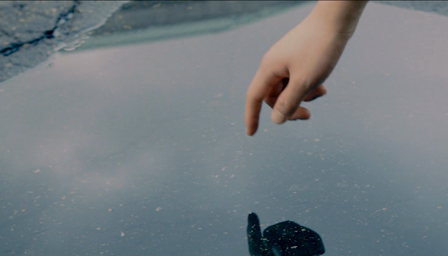}}

\subfigure[Flow$_{2\to 1}$]{\label{fig16:subfig:d}
\includegraphics[width=0.3\linewidth]{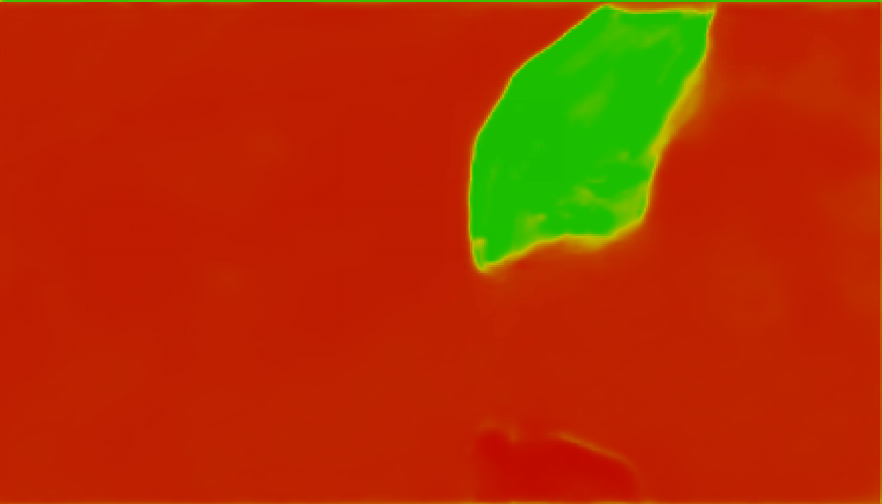}}
\subfigure[Frame 2 (Ours)]{\label{fig16:subfig:e}
\includegraphics[width=0.3\linewidth]{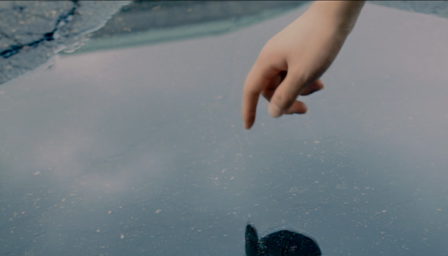}}
\subfigure[Flow$_{2\to 3}$]{\label{fig16:subfig:f}
\includegraphics[width=0.3\linewidth]{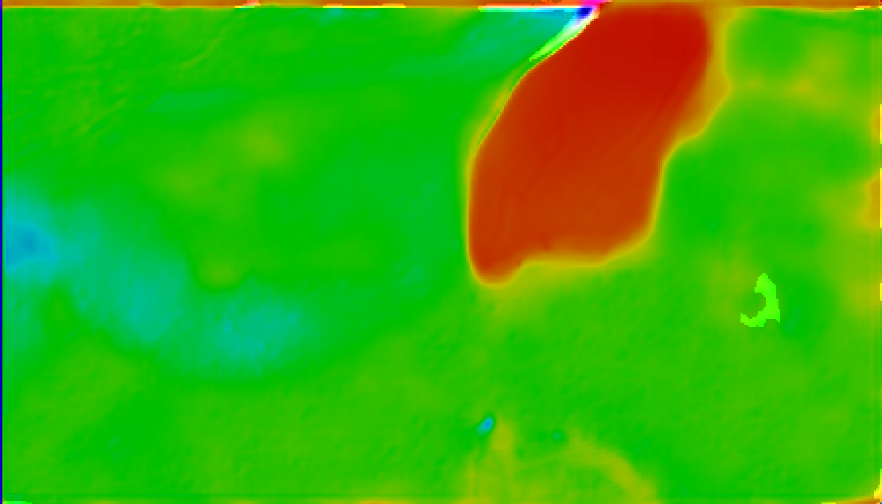}}

\caption{Visualize offsets into optical flow when $n=1$. We use Frame 1 (a) and Frame 3 (c) to generate Frame 2 (b) in (e). The backward and forward optical flows are shown in (d) and (f).}
\label{fig16}
\end{figure}

\begin{table}[t]
\centering
\caption{Ablation experiments to quantitatively analyze the effect of mask and bias estimators.}
{
\begin{tabular}{cccc}
\toprule
\multirow{2}*{Size} &
\multicolumn{1}{c}{ UCF101 }&
\multicolumn{1}{c}{ Vimeo90K }&
\multicolumn{1}{c}{ M.B. } \\

\cmidrule(lr){2-2}
\cmidrule(lr){3-3}
\cmidrule(lr){4-4}
& PSNR$\uparrow$/SSIM$\uparrow$ & PSNR$\uparrow$/SSIM$\uparrow$ & IE$\downarrow$\\
\midrule
w/o mask & 35.00 / \textbf{0.968} & 34.62 / 0.973 & 2.13 \\
w/o bias & 35.03 / \textbf{0.968} & 34.80 / \textbf{0.975} & 2.05 \\
\midrule
Full model & \textbf{35.13} / \textbf{0.968} & \textbf{34.84} / \textbf{0.975} & \textbf{2.02}\\
\bottomrule
\end{tabular}}
\label{tab12}
\end{table}
Since the flow based methods can be seen as a special case of our method when $n=1$ in section \ref{relationship}, we further visualize the offsets which are equivalent to optical flow in Figure \ref{fig16}. Without an explicit training phase for optical flow, our method learns meaningful information about motion between the frames for the task of frame interpolation.

\subsubsection{Mask estimator}
To examine the effectiveness of the mask estimator in our network, we trained a network without estimating masks. As shown in Table \ref{tab12}, the mask estimator significantly improves the performances on different datasets, especially in terms of PSNR and IE. On one hand, the learned masks help modulate the sample pixels guided by offsets, which allows the network to vary the spatial distribution and change the relative influence of the reference pixels \cite{dcn2}. On the other hand, masks reduce the burden of estimating separable convolution kernels, making the network better handle challenging cases such as occlusion.

\subsubsection{Bias estimator}
We compare the performance between model without bias estimator and the full model in Table \ref{tab12}. By introducing new bias values for each pixel's synthesis, the performance saturates in terms of SSIM while improves by 0.1 dB and 0.04 dB in terms of PSNR on UCF101 and Vimeo90K datasets, respectively. The learned bias values help to better model the linear relationship between the sampled pixels and corresponding kernels, which is in line with the flexible usage in common convolutional layers.

\section{Discussions and limitations}

By extending the approaches from \cite{sepconv, dsepconv}, our proposed EDSC\_s achieves the best performance and the EDSC\_m model is the first to able to produce an in-between frame at arbitrary time steps among all the kernel-based interpolation methods. However, our method has some limitations.
First, despite we prove theoretically the optical flow based interpolation methods to be specific instances of our method when $n=1$, the estimated bi-directional optical flows and generated frames fail to reach the same level of them. This is because the network is really simple and the offsets (which is equivalent to optical flow) are learned in an unsupervised manner, unlike those which utilize off-the-shelf flow estimation networks with a good initialization.
Second, for multiple frame interpolation, our EDSC\_m model is not so flexible as the methods in \cite{Ctxsyn,dain,softsplat} that explicitly warp pixels and features before generating the output frame. We need to train from scratch and supervise the model at different time steps $t$ while they do not.

Some recent researches enhance the performance of interpolation by making use of auxiliary information (\emph{e.g.}, more reference frames \cite{qvi, meta} and high frame rate video with low spatial resolution \cite{HIS}). Besides, a good initialization and fine-tuning of pre-trained sub-networks (such as PWC-Net \cite{PWCNet}, RBPN \cite{RBPN}, Megadepth \cite{megadepth}) can greatly help to produce high quality interpolation results. Although the well-known kernel based methods, including ours, do not utilize any of them, it would be interesting to explore its use and extend our method to generate frames with higher quality. Another direction in recent research is joint video enhancement problem \cite{bin, STAR, zoomingslowmo}. In the future, we plan to extend our approach to fix more tasks in the area of video processing.

\section{Conclusion}
In this paper, we have presented an enhanced deformable separable network for video frame interpolation. Our method improves the performance of kernel-based methods with fewer parameters by processing the information in a non-local neighborhood with learned adaptive offsets, kernels, masks and biases. And we provide the first kernel-based method that can generate as many intermediate frames as needed between two consecutive frames. Further, as demonstrated theoretically, both kernel- and flow-based methods can be regarded as special cases of our method. Comprehensive experiments show that our method performs favorably against state-of-the-art methods.



\normalsize
\bibliography{mybib}


\end{document}